\PassOptionsToPackage{dvipsnames,table}{xcolor} 
\documentclass[letterpaper,12pt]{article}
\usepackage{silence} 
\usepackage{scrextend} 

\usepackage{csm-thesis}

\usepackage[backend=biber,style=authoryear,url=false,maxcitenames=2,minbibnames=3,uniquelist=minyear,uniquename=false,giveninits,dashed=false,useprefix=true,abbreviate=false,date=year,isbn=false]{biblatex}
\DeclareNameAlias{sortname}{family-given}
\DeclareFieldFormat{doi}{%
  \ifhyperref
    {\href{https://doi.org/#1}{\nolinkurl{https://doi.org/#1}}}
    {\nolinkurl{https://doi.org/#1}}}

\DeclareFieldFormat{url}{\url{#1}}

\usepackage{xpatch}

\xpatchbibmacro{name:andothers}{ 
{\finalandcomma}%
}{%
\addspace%
}{}{}

\renewbibmacro{begentry}{\midsentence}

\DefineBibliographyStrings{english}{
  backrefpage={Cited on page},
  backrefpages={Cited on pages}
}

\DeclareSourcemap{
  \maps[datatype=bibtex, overwrite=true]{
    \map{
      \step[fieldsource=url, final]
      \step[typesource=misc, typetarget=online]
    }
  }
}
\DeclareCiteCommand{\citeauthor}
  {\boolfalse{citetracker}%
   \boolfalse{pagetracker}%
   \usebibmacro{prenote}}
  {\ifciteindex
     {\indexnames{labelname}}
     {}%
   \printtext[bibhyperref]{\printnames{labelname}}}
  {\multicitedelim}
  {\usebibmacro{postnote}}
\DeclareCiteCommand{\citeyearpar}
    {}
    {\mkbibparens{\bibhyperref{\printfield{year}}}}
    {\multicitedelim}
    {}

\renewbibmacro*{volume+number+eid}{%
  \printfield{volume}%
  \setunit*{\addnbspace}
  \printfield{number}%
  \setunit{\addcomma\space}%
  \printfield{eid}}
\DeclareFieldFormat[article]{number}{\mkbibparens{#1}}

\renewbibmacro{in:}{}

\DeclareFieldFormat[article]{volume}{\mkbibbold{#1}}

\renewbibmacro*{maintitle+title}{%
  \iffieldsequal{maintitle}{title}
    {\clearfield{maintitle}%
     \clearfield{mainsubtitle}%
     \clearfield{maintitleaddon}}
    {\iffieldundef{maintitle}
       {}
       {\usebibmacro{maintitle}%
    \newunit\newblock
    \iffieldundef{volume}
      {}
      {\printfield{volume}%
           \printfield{part}%
           \setunit{\addcolon\space}}}}%
  \usebibmacro{title}%
  \iffieldundef{edition}
    {\nopunct}%
    {\addcomma}%
  \newunit}

\DeclareFieldFormat
  [article,inproceedings,incollection,online]
  {title}{#1\isdot}

\DeclareFieldFormat[article,incollection]{pages}{#1}
\DeclareFieldFormat*{booktitle}{#1}

\usepackage{amsmath} 
\numberwithin{equation}{section} 
\usepackage{hyperref} 
\usepackage{cleveref} 
\crefname{equation}{Expression}{Expressions}
\Crefname{equation}{Equation}{Equations}
\creflabelformat{equation}{#2\textup{#1}#3} 

\usepackage{amssymb}

\newcommand*\diff{\mathop{}\!\mathrm{d}} 

\usepackage{comment}
\usepackage[utf8]{inputenc}
\usepackage{chemmacros}
\usepackage{upgreek} 
\usepackage{siunitx}
\usepackage{booktabs}
\usepackage{threeparttable}
\usepackage{adjustbox}
\usepackage{bm}
\usepackage{textcomp}
\usepackage[section]{placeins}
\usepackage{microtype}
\usepackage{epigraph}
\usepackage{relsize} 
\usepackage{mathtools} 
\usepackage{commath} 
\DeclareMathOperator{\stut}{Student-\mathnormal{t}}
\allowdisplaybreaks 
\DeclareSIUnit\mcf{Mcf}
\DeclareSIUnit\mmcf{MMcf}
\DeclareSIUnit\boe{BOE}
\DeclareSIUnit\bpd{bbl/day}
\DeclareSIUnit\bcm{bcm}
\DeclareSIUnit\dpbbl{\$/bbl}
\DeclareSIUnit\dpmmbtu{\$/MMBtu}
\DeclareSIUnit\mcfpbbl{Mcf/bbl}
\DeclareSIUnit\bbl{bbl}
\usepackage{dsfont}
\usepackage{morewrites} 
\usepackage{enumitem} 
\usepackage{nth}
\usepackage{interval}
\usepackage{rotating} 

\usepackage{multirow} 
\newcolumntype{R}{>{$}r<{$}} 
\usepackage{makecell} 

\usepackage[ruled,linesnumbered,algochapter]{algorithm2e}

	\title{Application of Machine Learning to Gas Flaring}

\degreetitle{Doctor of Philosophy}
\discipline{Petroleum Engineering}
\department{Petroleum Engineering}

\author{Rong Lu}
\advisor{Dr.~Jennifer L. Miskimins}
\dpthead{Dr.~Jennifer L. Miskimins}{Professor and Head}

\begin{document}



\frontmatter


\maketitle
\newpage


\makecopyright{2020}
\newpage


\makesubmittal
\newpage


\begin{abstract}
Currently in the petroleum industry, operators often flare the produced gas instead of commodifying it. The flaring magnitudes are large in some states, which constitute problems with energy waste and CO\textsubscript{2} emissions. In North Dakota, operators are required to estimate and report the volume flared. The questions are, how good is the quality of this reporting, and what insights can be drawn from it?

Apart from the company-reported statistics, which are available from the North Dakota Industrial Commission (NDIC), flared volumes can be estimated via satellite remote sensing, serving as an unbiased benchmark. Since interpretation of the Landsat 8 imagery is hindered by artifacts due to glow, the estimated volumes based on the Visible Infrared Imaging Radiometer Suite (VIIRS) are used. Reverse geocoding is performed for comparing and contrasting the NDIC and VIIRS data at different levels, such as county and oilfield.

With all the data gathered and preprocessed, Bayesian learning implemented by Markov chain Monte Carlo methods is performed to address three problems: county level model development, flaring time series analytics, and distribution estimation. First, there is heterogeneity among the different counties, in the associations between the NDIC and VIIRS volumes. In light of such, models are developed for each county by exploiting hierarchical models. Second, the flaring time series, albeit noisy, contains information regarding trends and patterns, which provide some insights into operator approaches. Gaussian processes are found to be effective in many different pattern recognition scenarios. Third, distributional insights are obtained through unsupervised learning. The negative binomial and Gaussian mixture models are found to effectively describe the oilfield flare count and flared volume distributions, respectively. Finally, a nearest-neighbor-based approach for operator level monitoring and analytics is introduced.
\end{abstract}

\newpage


\tableofcontents
\newpage


\listoffiguresandtables
\newpage





\begin{NoHyper} 
\listofsymbols*

\noindent
Vectors and matrices are in bold type. A subscript asterisk, such as in $y_*$, indicates reference to a test set quantity or a prediction.

\listofsymbols*{General Nomenclature}
\listofsymbols*{Probability and Statistics}
\listofsymbols*{Probability Distributions}
\listofsymbols*{Gaussian Processes}
\listofsymbols*{Vectors and Matrices}
\listofsymbols*{Standard Functions}
\listofsymbols*{Units}
\newpage


\addsymbol[General Nomenclature]{Data set: $\mathcal{D}=\{(\mathbf{x}_i, y_i) \mid i=1, \ldots , n\}$}{$\mathcal{D}$}
\addsymbol[General Nomenclature]{Prediction for a test input $\mathbf{x}_*$}{$y_*$}
\addsymbol[General Nomenclature]{Proportional to; e.g., $p(x \mid y) \propto p(x,y)$ means that $p(x\mid y)$ is \\\hspace*{1.5em}equal to $p(x,y)$ times a factor which is independent of $x$}{$\propto$}
\addsymbol[General Nomenclature]{$a$ is defined as $b$}{$a \coloneqq b$}
\addsymbol[General Nomenclature]{$b$ is defined as $a$}{$a \eqqcolon b$}
\addsymbol[General Nomenclature]{$\ell_2$ norm}{$\norm{\cdot}$}
\addsymbol[General Nomenclature]{$n$-dimensional vector space of real numbers}{$\mathds{R}^{n}$}
\addsymbol[General Nomenclature]{Natural numbers with zero}{$\mathds{N}_{0}$}
\addsymbol[General Nomenclature]{Real numbers}{$\mathds{R}$}
\addsymbol[General Nomenclature]{Pi (italic) representing a variable}{$\pi$}
\addsymbol[General Nomenclature]{Pi (upright) denoting the transcendental constant ($3.14159\dotsm$)}{$\uppi$}
\addsymbol[General Nomenclature]{Universal quantifier: for all $x$}{$\forall x$}

\addsymbol[Probability and Statistics]{Conditional probability density function}{$p(\cdot \mid \cdot)$}
\addsymbol[Probability and Statistics]{Probability density function}{$p(\cdot)$}
\addsymbol[Probability and Statistics]{Probability mass function}{$P(\cdot)$}
\addsymbol[Probability and Statistics]{Random variable $X$ is distributed according to $p$}{$X \sim p$}
\addsymbol[Probability and Statistics]{Expectation; expectation of $g(x)$ when $x \sim p$}{$\mathds{E}$ or $\mathds{E}_{p}[g(x)]$}
\addsymbol[Probability and Statistics]{Variance; variance of $g(x)$ when $x \sim p$}{$\mathds{V}$ or $\mathds{V}_{p}[g(x)]$}

\addsymbol[Probability Distributions]{Univariate Gaussian distribution with parameters $\mu, \, \sigma$}{$\mathcal{N}(\mu, \, \sigma)$}
\addsymbol[Probability Distributions]{Multivariate Gaussian distribution with parameters $\bm{\mu}, \, \bm{\Sigma}$}{$\operatorname{MVNormal}(\bm{\mu}, \, \bm{\Sigma})$}
\addsymbol[Probability Distributions]{Poisson distribution with parameter $\lambda$}{$\operatorname{Poisson}(\lambda)$}
\addsymbol[Probability Distributions]{Exponential distribution with parameter $\lambda$}{$\operatorname{Exponential}(\lambda)$}
\addsymbol[Probability Distributions]{Binomial distribution with parameters $n, \, p$}{$\operatorname{Binomial}(n, \, p)$}
\addsymbol[Probability Distributions]{Gamma distribution with parameters $\alpha, \, \beta$}{$\operatorname{Gamma}(\alpha, \, \beta)$}
\addsymbol[Probability Distributions]{Half-Normal distribution with parameter $\sigma$}{$\operatorname{Half-Normal}(\sigma)$}
\addsymbol[Probability Distributions]{Half-Cauchy distribution with parameter $\gamma$}{$\operatorname{Half-Cauchy}(\gamma)$}
\addsymbol[Probability Distributions]{LKJ distribution with parameter $\eta$}{$\operatorname{LKJcorr}(\eta)$}
\addsymbol[Probability Distributions]{Distribution over Cholesky decomposed covariance \\\hspace*{1.5em}matrices with parameters $\eta, \, \bm{\sigma}$}{$\operatorname{LKJCholeskyCov}(\eta, \, \bm{\sigma})$}
\addsymbol[Probability Distributions]{Student's $t$-distribution with parameters $\nu, \, \mu, \, \sigma$}{$\stut (\nu, \, \mu, \, \sigma)$}
\addsymbol[Probability Distributions]{Dirichlet distribution with parameter $\bm{\alpha}$}{$\operatorname{Dirichlet}(\bm{\alpha})$}
\addsymbol[Probability Distributions]{Categorical distribution with parameter $\mathbf{p}$}{$\operatorname{Categorical}(\mathbf{p})$}
\addsymbol[Probability Distributions]{Negative binomial distribution with parameters $\mu, \phi$}{$\operatorname{NegBinomial}(\mu, \phi)$}
\addsymbol[Probability Distributions]{Continuous uniform distribution with parameters $a, b$}{$\operatorname{Uniform}(a, b)$}

\addsymbol[Gaussian Processes]{Gaussian process: $f \sim \mathcal{GP}(m(\mathbf{x}), k(\mathbf{x}, \mathbf{x}'))$, the function $f$ is \\\hspace*{1.5em}distributed as a Gaussian process}{$\mathcal{GP}$}
\addsymbol[Gaussian Processes]{Mean function evaluated at $\mathbf{x}$}{$m(\mathbf{x})$}
\addsymbol[Gaussian Processes]{Covariance function evaluated at $\mathbf{x}$ and $\mathbf{x}'$}{$k(\mathbf{x}, \mathbf{x}')$}
\addsymbol[Gaussian Processes]{Vector of latent function values, $\mathbf{f} = (f(\mathbf{x}_1), \dots, f(\mathbf{x}_n))^\top$}{$\mathbf{f}$}

\addsymbol[Vectors and Matrices]{Transpose of matrix $\mathbf{L}$}{$\mathbf{L}^\top$}
\addsymbol[Vectors and Matrices]{Vector of parameters}{$\bm\theta$}
\addsymbol[Vectors and Matrices]{Cholesky decomposition: $\mathbf{L}$ is a lower triangular matrix \\\hspace*{1.5em}such that $\mathbf{L} \cdot \mathbf{L}^\top=\mathbf{K}$}{$\operatorname{Cholesky}(\mathbf{K})$}
\addsymbol[Vectors and Matrices]{Vector of all $0$'s of length $n$}{$\mathbf{0}_n$}
\addsymbol[Vectors and Matrices]{Vector of all $1$'s of length $n$}{$\mathds{1}_n$}
\addsymbol[Vectors and Matrices]{Identity matrix of size $n \times n$}{$\mathbf{I}_n$}

\addsymbol[Standard Functions]{Natural exponential function}{$\exp(\cdot)$}
\addsymbol[Standard Functions]{Natural logarithm function}{$\log(\cdot)$}
\addsymbol[Standard Functions]{Inverse-logit function}{$\operatorname{logit}^{-1}(\cdot)$}

\addsymbol[Units]{meter}{\si{\metre}}
\addsymbol[Units]{kilometer}{\si{\kilo\metre}}
\addsymbol[Units]{day}{\si{\day}}
\addsymbol[Units]{megawatt}{\si{\mega\watt}}
\addsymbol[Units]{kelvin}{\si{\kelvin}}
\addsymbol[Units]{thousand cubic feet}{\si{\mcf}}
\addsymbol[Units]{million cubic feet}{\si{\mmcf}}
\addsymbol[Units]{billion cubic meter}{\si{\bcm}}
\addsymbol[Units]{barrels of oil equivalent}{\si{\boe}}
\addsymbol[Units]{barrels per day}{\si{\bpd}}
\addsymbol[Units]{dollars per barrel}{\si{\dpbbl}}
\addsymbol[Units]{dollars per million Btu}{\si{\dpmmbtu}}
\addsymbol[Units]{thousand cubic feet per barrel}{\si{\mcfpbbl}}

\listofabbreviations*
\newpage

\addabbreviation{North Dakota Industrial Commission}{NDIC}
\addabbreviation{Landsat 8}{L8}
\addabbreviation{Deep learning}{DL}
\addabbreviation{Visible Infrared Imaging Radiometer Suite}{VIIRS}
\addabbreviation{VIIRS Nightfire}{VNF}
\addabbreviation{National Oceanic and Atmospheric Administration}{NOAA}
\addabbreviation{National Aeronautics and Space Administration}{NASA}
\addabbreviation{Markov chain Monte Carlo}{MCMC}
\addabbreviation{Hamiltonian Monte Carlo}{HMC}
\addabbreviation{Independent and identically distributed}{i.i.d.}
\addabbreviation{Radiant heat}{RH}
\addabbreviation{Short-wave infrared}{SWIR}
\addabbreviation{Exempli gratia (Latin: for example)}{e.g.}
\addabbreviation{Id est (Latin: that is)}{i.e.}
\addabbreviation{Gaussian process}{GP}
\addabbreviation{Gaussian mixture model}{GMM}
\addabbreviation{Autoregressive integrated moving average}{ARIMA}
\addabbreviation{Long short-term memory}{LSTM}
\addabbreviation{World Geodetic System}{WGS}
\addabbreviation{North American Datum}{NAD}
\addabbreviation{Maximum likelihood estimation}{MLE}
\addabbreviation{Maximum a posteriori}{MAP}
\addabbreviation{Hierarchical Density-Based Spatial Clustering \\\hspace*{1.5em}of Applications with Noise}{HDBSCAN}
\addabbreviation{Density-Based Spatial Clustering of Applications \\\hspace*{1.5em}with Noise}{DBSCAN}
\addabbreviation{Gas oil ratio}{GOR}
\addabbreviation{Credible interval}{CI}
\addabbreviation{Prediction interval}{PI}
\addabbreviation{Right-hand side}{RHS}
\addabbreviation{Highest density interval}{HDI}
\addabbreviation{Interquartile range}{IQR}
\addabbreviation{Energy Information Administration}{EIA}
\addabbreviation{Zero-inflated Poisson}{ZIP}
\addabbreviation{Zero-inflated negative binomial}{ZINB}
\addabbreviation{Kernel density estimation}{KDE}
\end{NoHyper}


\begin{acknowledgments}
In the very first place, I want to express my deepest appreciation to my advisor, Dr.~Jennifer L. Miskimins. Dr.~Miskimins has been my MS/PhD advisor and mentor since 2011. Since I returned to Mines to start my PhD in 2017, Dr.~Miskimins has been providing me with the best guidance, the greatest support, and the most opportunities that I could imagine. During the first semester, I worked as a lab assistant in the High Bay; in a later semester, I worked as a teaching assistant in her well stimulation course; ever since I started to become interested in machine learning, she has provided me with a huge number of opportunities to connect with different groups of people, for brainstorming and pursuing my research interest. To a certain extent, I feel like I finally become a ``qualified'' FAST student member, thanks to all of these precious experience. What I have achieved, including this dissertation, would have never been possible without the guidance and support from Dr.~Miskimins. Her world-class technical expertise, attitudes toward work/life, and art of managing different teams at various levels are what I hope I can learn from in my career and personal life.

I am deeply grateful to my dissertation committee members: Dr.~Soutir Bandyopadhyay, Dr.~Alfred W. Eustes III, Dr.~Yilin Fan, and Prof.~Jim Crompton. My competency in my research field, as well as the shape of this dissertation are built with the help of those fruitful discussions and insightful comments from them.

I am indebted to my mentors, colleagues, and friends from the Payne Institute for Public Policy. Especially, I want to thank Dr.~Mikhail N. Zhizhin, Dr.~Christopher D. Elvidge, and Dr.~Morgan D. Bazilian. It is such an eye-opening experience for me to work with these world-class experts in remote sensing and satellite imagery. I would particularly like to thank Dr.~Zhizhin for his help, insights, and time.

I am really grateful to Dr.~Bandyopadhyay and Dr.~Luis Tenorio from the AMS Department, for their fantastic teaching, knowing me personally and motivating me to work hard. Looking back at what I have learned in machine learning which makes this dissertation possible, taking their classes are definitely the most important resources for myself (excuse me for not being a probabilist at this moment). By taking their statistical methods classes, I started to appreciate what really \textit{is} machine learning, and falling in love with mathematics, more specifically, probability theory and statistical modeling.


The TA experience at Mines makes me a better PhD student. What I have learned, technical or non-technical, made their way into this dissertation. I want to thank Prof.~Crompton, Dr.~Eustes, Dr.~Mark G. Miller, Dr.~Linda A. Battalora, and Dr.~Miskimins for providing me with those valuable TA opportunities. I am grateful to all of my students for their support and feedback.

I want to thank Dr.~Yu-Shu Wu, Dr.~Xiaolong Yin, and Dr.~Yilin Fan for their care, support, and encouragement throughout my PhD study. I would like to thank Denise Winn-Bower, Rachel McDonald, and Joe Chen for their help.

I really appreciate the feedback from the FAST member companies' representatives. A lot of the discussions and the reflections following those were incorporated into this dissertation. Especially, I want to thank Ty Woodworth for his time and help, in the process of collecting plunger lift data for me. I got very warm welcomes every time I visited their Windsor office in Northern Colorado. Ty kindly introduced me to the team he led, and I got the great opportunities to ask questions and discuss with many field experts in different areas. Those discussions helped me tremendously.

Special thanks go to the open source community. In the process of conducting this research and typesetting this dissertation, I benefited a lot from the ecosystems around Linux/GNU, \TeX{}/\LaTeX{}, and Python. Especially, I want to thank the people behind PyMC3, a probabilistic programming language that this dissertation is heavily dependent upon.

Last but not least, I would like to thank my family and friends. Thank you to my beloved wife Xiaodan, for all her love, support, and delicious dishes. I also want to thank my parents and parents-in-law for their support, encouragement, and understanding.
\end{acknowledgments}
\newpage


\begin{dedication}
I dedicate this work to my mother, Dr.~Lingying Ni, and my father, Mr.~Honggang Lu.

\includegraphics[scale=1.1]{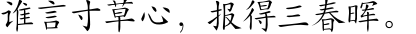}
\end{dedication}
\newpage


\bodymatter


\chapter{Introduction}

Currently in the petroleum industry, for wells which produce both crude oil and natural gas, operators often choose to flare the produced gas instead of commodifying it. The rationales behind such decisions are multifold. Variations in natural gas price can be an important factor, especially when the processing and transportation cost is higher than the value of gas~\autocite{Srivastava2019DistributionalPolicy}. The amount of gas being flared each year on a national level is huge, and an increasing trend can be observed for the top flaring countries (\ref{fig:top30_flare_nation}).

\csmlongfigure[!htbp]{top30_flare_nation}{figures/NewrankingTop30flaringcountries20142018}{\textwidth}{Top 30 countries ranked by flared gas volume in 2018.}{ United States ranks No.~4 and has a large increase from 2017 to 2018~\autocite{WorldBank2019GlobalData}.}

Due to the boom of unconventional resources (e.g., shale gas reservoirs) development in the recent decade, the United States has been among the top flaring countries in terms of total volume flared. This is backed by the data from the U.S. Energy Information Administration~(EIA) \citeyearpar{EIA2019NaturalFlared} showing North Dakota, which is underlain by the Bakken Formation, and Texas, which houses the Permian Basin and the Eagle Ford Shale, are the top two flaring states since 2013. The two states' annual flaring volume time series are shown in \ref{fig:nd_vs_tx}. Some flaring sites can be clearly identified from Google Earth's imagery (\ref{fig:goog_earth_day_img}).

\csmlongfigure[!htbp]{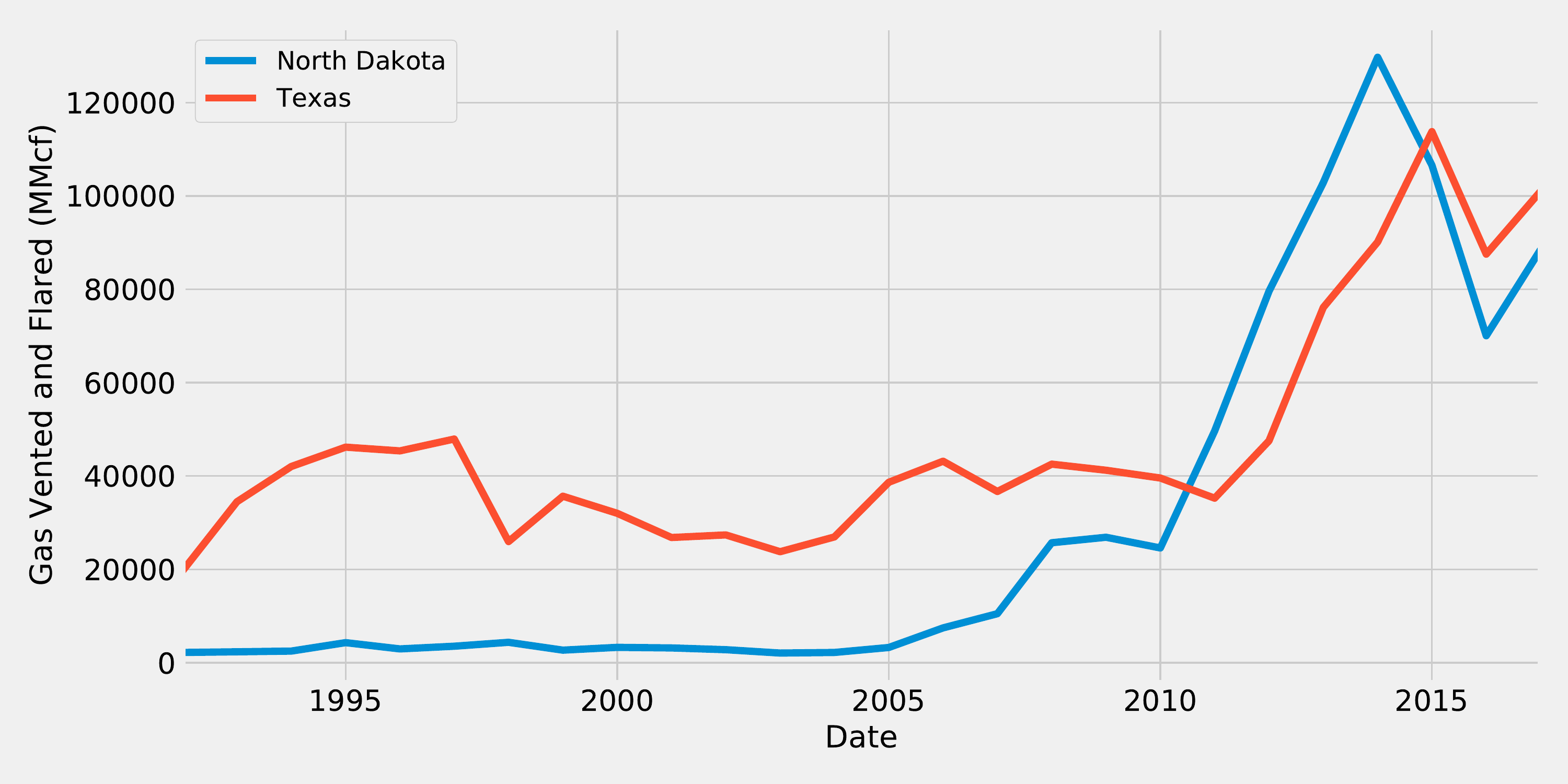}{figures/nd_vs_tx}{\textwidth}{The time series show the trend of gas flaring for the top two states in the United States~\autocite{EIA2019NaturalFlared}.}{ Texas regained the lead in 2015.}

\csmfigure[!htbp]{goog_earth_day_img}{figures/googleEarth_bakken_flare_a}{\textwidth}{This Google Earth imagery shows gas flaring being conducted on a well location in North Dakota~\autocite{GoogleEarth2019SatelliteDakota}.}

Natural gas flaring constitutes a problem of energy waste and \ch{CO2} emissions. In recent years, various organizations and government agencies have advocated reducing or eliminating routine gas flaring. For example, the \citeauthor{NorthDakotaIndustrialCommission2014North041718} (NDIC) introduced a gas flaring regulatory policy (Order 24665) in 2014, with goals of reducing flaring in different aspects (e.g., volume of gas flared). The World Bank launched the ``Zero Routine Flaring by 2030'' initiative in 2015. To monitor and benchmark flaring activity's magnitude, a precise and accurate method to obtain quantitative flaring information is desirable. However, in certain situations, this information is only available through self-reporting mechanisms. Inaccuracies might be introduced either intentionally or unintentionally. 

Satellite remote sensing is one unbiased approach for solving this problem. It can help detect active flares especially during nighttime and can be used to calibrate the estimation for flared gas volume. For this work, two different types of sensors are considered, including the Landsat 8 (L8)'s Operational Land Imager (OLI) and Thermal Infrared Sensor (TIRS), as well as the Visible Infrared Imaging Radiometer Suite (VIIRS) that is on the Suomi National Polar-orbiting Partnership (NPP) and NOAA-20 satellites. In the remainder of this dissertation, they are referred to as L8 and VIIRS, respectively. An example of detecting flaring with VIIRS low light imaging data is shown in \ref{fig:eog_viirs_poster_crop_us}.

\csmlongfigure[!htbp]{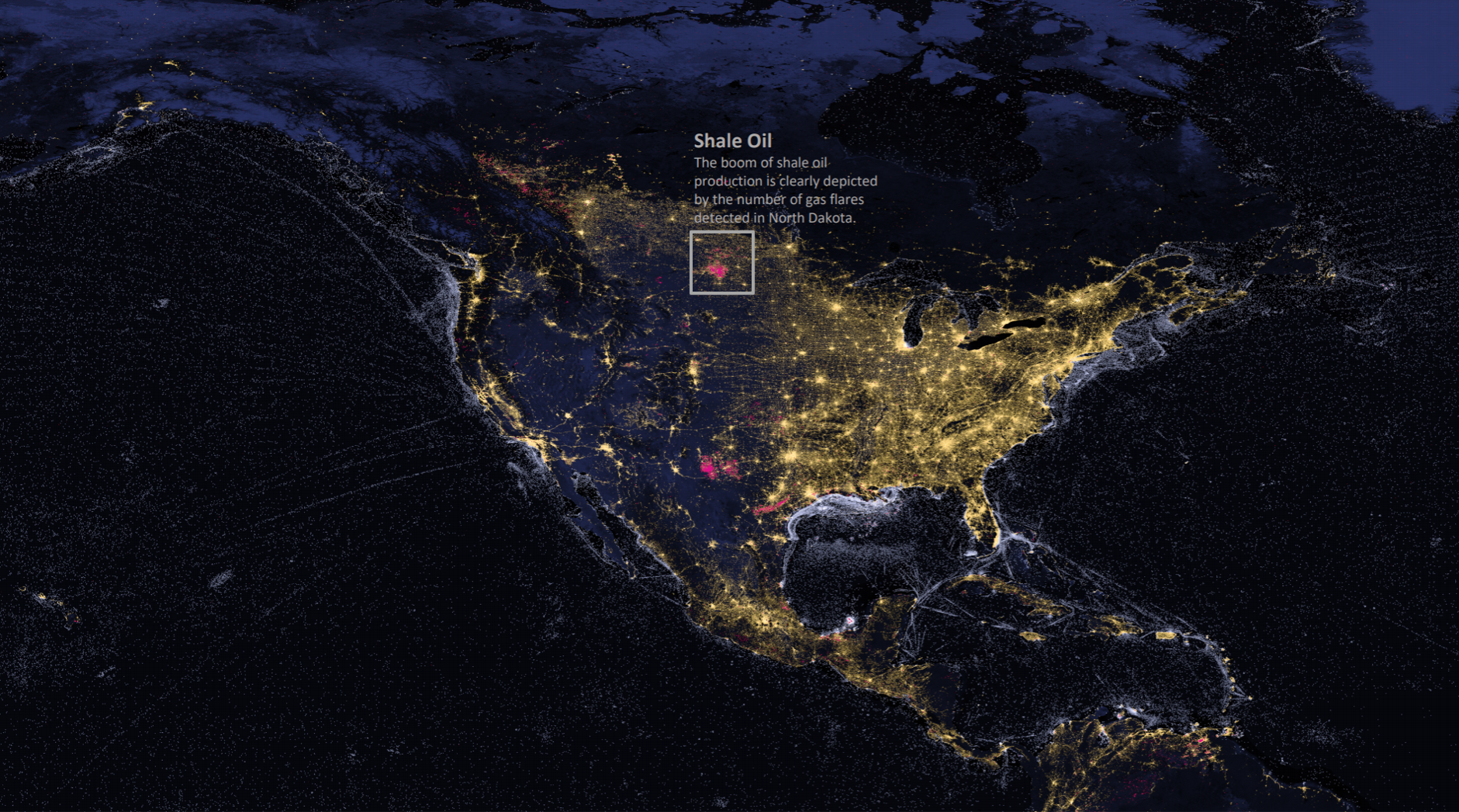}{figures/eog_viirs_poster_crop_us}{\textwidth}{Part of the original poster~\autocite{EarthObservationGroupatPayneInstitute2019EOGFile} which uses one year accumulation of VIIRS low light imaging data to showcase human activities, e.g., gas flaring, fishing, and city lights.}{ As annotated, North Dakota's flaring activities are very visible from space at night.}

\subsection{Research Goal}
\label{sec:thesis-goal}
This research is undertaken to achieve the following goals:
\begin{itemize}
    \item Evaluate the methodology for estimating flared gas volume leveraging satellite imagery; and,
    \item Find insights into operators' gas flaring behavior.
\end{itemize}

\subsection{Dissertation Objectives}\label{sec:thesis_obj}
To achieve the goals outlined in Section~\ref{sec:thesis-goal}, more specific objectives are listed below:
\begin{enumerate}
    \item Compare and contrast the flaring data from VIIRS and NDIC.
        \begin{itemize}
            \item Compare the VIIRS flared volumes to the NDIC, using the NDIC as a benchmark.
        \end{itemize}
    \item Evaluate the effectiveness of using Landsat 8 nighttime images to improve flare detection and volume estimation.
        \begin{itemize}
            \item Determine the detection limits of Landsat 8 and compare it with VIIRS' capabilities.
        \end{itemize}
        
    \item Investigate operator approaches for gas flaring. 
        \begin{itemize}
            \item Determine the correlation between gas price / oil price / oil production and flared gas volume.
            \item Evaluate if the North Dakota regulatory policy (Order 24665) achieved its goals.
            \item Develop a model that can predict flared gas volume at a state level.
        \end{itemize}
        
    \item Find any hidden structure/clusters from all the producing entities.
\end{enumerate}

\subsection{Outline and Contributions}
The main contribution of this dissertation is demonstrating that Bayesian learning implemented by Markov chain Monte Carlo methods is very effective in flaring data analytics. A series of parametric and nonparametric machine learning models are developed for various analytics goals and granularities, providing direct guidance for future modeling endeavors. To demonstrate the effectiveness and robustness, they are all tested with real data. The superiority of this approach is based on the fact that the inference stage is entirely probabilistic, in that the parametric uncertainties arising from probable models as well as the stochastic uncertainties arising from noisy observations are all properly characterized and quantified. It makes the extracted insights robust and interpretable for decision- and policy-making by, for example, a state government.

In Chapter~\ref{ch:lit_rev}, a literature review is given for the state of the art in satellite imagery processing, Bayesian inference, Markov chain Monte Carlo methods, and machine learning.

In Chapter~\ref{ch:data_pre_eda}, the data gathering processes are discussed. Results from some exploratory data analysis are presented.

In Chapter~\ref{ch:hier}, county level models are built to study the correlations between VIIRS and NDIC, and to explore the heterogeneity among the counties in North Dakota.

In Chapter~\ref{ch:gp}, flaring time series analytics is presented for the purposes of revealing trends and patterns at different levels.

In Chapter~\ref{ch:gmm}, unsupervised learning is applied on flaring data to characterize the latent structures.

In Chapter~\ref{ch:discuss}, a method of operator level monitoring and analytics is introduced, and some discussions about applying Bayesian learning are given.

In Chapter~\ref{ch:conclusions}, major conclusions drawn are presented. Recommendations based on this work are given. A number of future research areas are outlined.

\chapter{Literature Review}\label{ch:lit_rev}

In the 1990s, the World Bank started gathering nighttime satellite images, from which big cities and oilfields were both bright and needed to be sorted using extra information. The situation changed in 2012 when infrared data became available from VIIRS~\autocite{Rassenfoss2018EP2018}. One of the data products, VIIRS Nightfire (VNF) specializes in natural gas flaring observation and is even able to distinguish between biomass burning and gas flaring~\autocite{Elvidge2017SupportingViirs}. 

VNF's development was based upon VIIRS imagery. To improve the performance of flare detection and gas volume estimation, other sources of information, such as L8 imagery, can be leveraged. \ref{tab:l8_viirs_resolution} presents a comparison of L8 and VIIRS spatial and temporal resolutions~\autocite{NASA2019Landsat8, Wikipedia2019VisibleSuite}. \ref{fig:L8pixelsizes} illustrates L8's spatial resolution. In addition, L8 collects data in \num{11} different spectral bands of the electromagnetic spectrum. VIIRS has \num{22} bands. Both L8 and VIIRS are in near-polar orbits of the earth and can reveal rich features in the landscape. Therefore, L8 should be able to identify smaller gas flares compared to VIIRS' capability, although its longer satellite revisit time poses a challenge to identify less persistent flares. More details on the processing steps of VNF are discussed in Section~\ref{sec:vnf_process_steps}, the essence of which will be applied to L8.

\begin{table}[!htbp]
\centering
\caption{\label{tab:l8_viirs_resolution}Resolutions of Landsat 8 and VIIRS}
\begin{threeparttable}
	\sisetup{
        table-text-alignment = right
    }
	\begin{tabular}{lSS[table-format=2.1]}
		\toprule
		& \multicolumn{2}{c}{Resolution Type} \\ \cmidrule(lr){2-3}
		& {Spatial [\si{\metre}]} & {Temporal [\si{\day}]} \\
		\midrule
		Landsat 8 & \numrange{15}{100}\tnote{$\dagger$} & 16.0\tnote{$\ddagger$} \\
		VIIRS & \numrange{375}{750}\tnote{$\dagger$} & 0.5 \\
		\bottomrule
	\end{tabular}
    \begin{tablenotes}
        \item[$\dagger$] Depends on the band of the electromagnetic spectrum
        \item[$\ddagger$] For daytime mode
    \end{tablenotes}
\end{threeparttable}
\end{table}

\csmlongfigure[!htbp]{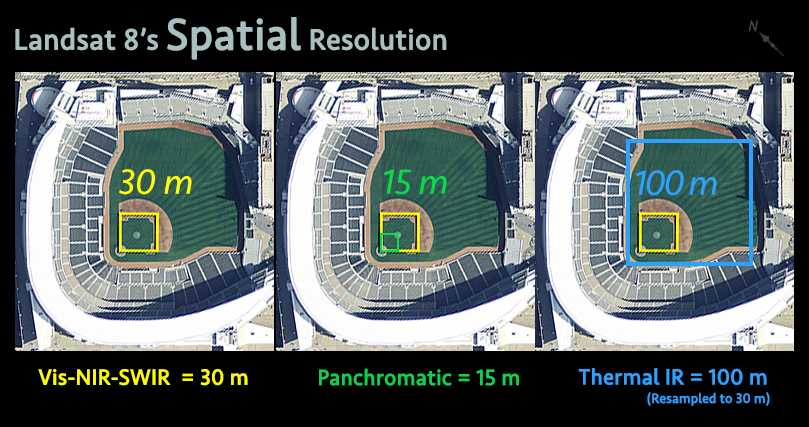}{figures/L8pixelsizes}{\textwidth}{Landsat 8's spatial resolution~\autocite{NASA2020LandsatOverview}.}{ Each Landsat pixel (\num{30} by \num{30} meter area) is roughly the size of a baseball diamond.}

Nowadays, one resource which is more than abundant is data. For a certain discipline or research field, new sources of data bring in new dimensions of information, such as satellite images are now playing a role in gas flaring analytics. How to analyze data effectively and intelligently to gain insights is a central problem. In the petroleum engineering domain, for example, data driven approaches have been proposed to analyze stimulation treatments~\autocite{Kazakov2011ApplicationSystems} and predict screenouts~\autocite{Yu2020AWells}. Machine learning is a powerful tool for this purpose. It is at the core of artificial intelligence and data science, and lies at the intersection of statistics and computer science~\autocite{Jordan2015MachineProspects}. Frameworks in computational learning theory, such as the PAC learning proposed by \textcite{pac}, help provide a theoretical backbone for some learning algorithms.

One subset of machine learning, deep learning (DL), had its debut in 2006 when \citeauthor{Hinton2006ReducingNetworks} introduced Deep Belief Networks (DBN), but it did not gain wide acceptance until 2012 when AlexNet showed the breakthrough performance on classification accuracy in the ImageNet competition~\autocite{NIPS2012_4824}. AlexNet is a DL-based model (more specifically a convolutional neural network) and achieved an error rate of \SI{15.3}{\percent}, which is more than \SI{10}{\percent} lower than the runner-up. DL dominated the competition thereafter, and DL-based models finally surpassed human performance on the classification data set in 2015~\autocite{He_2015_ICCV}. 

Although neural network-based models have gained much success in recent years, it should be noted that no one type of model can always be the best candidate for all problems. This has been formally shown by \textcite{Wolpert1996TheAlgorithms}, and is usually referred to as the ``no free lunch'' (NFL) theorem. More recently, \textcite{2017arXiv170805070O} empirically assessed 13 classification algorithms on 165 different problem sets, and the results aligned with the theorem: even the union of the top five best performing algorithms cannot dominate all of the problem sets.

In the following sections, a detailed review is given for the aspects below, which serve as the foundation and inspiration for this work:
    \begin{enumerate}
        \item Satellite image processing
        \item Bayesian inference
        \item Markov chain Monte Carlo
        \item Machine learning
        \item Analytics toolset
    \end{enumerate}

\subsection{Satellite Image Processing}
\label{sec:vnf_process_steps}
Satellite images are utilized to estimate flared gas volume. The fire detection algorithm based on Planck curve fitting and physical laws, known as VIIRS Nightfire (VNF) due to \textcite{Elvidge2013VIIRSNight}, serves as a starting point for analyzing L8 images in this research. The method consists of several major steps:
\begin{enumerate}
    \item Detection of hot pixels
    
        During nighttime, the sensors mainly record instrument noise which approximately follows a Gaussian distribution, except for the few pixels that contain an infrared emitter such as a gas flare. Therefore hot pixels can be identified by setting a cutoff on the tail of the distribution, e.g., those pixels with digital numbers exceeding the mean plus four standard deviations.
    \item Noise filtering
    
        Hot pixels that are detected in only one spectral band are treated as noise and filtered out.
    \item Atmospheric correction
    
        Losses in radiance due to scattering and absorption effects can be corrected. MODTRAN\,\textsuperscript{\textregistered} 5~\autocite{modtran}, parameterized with atmospheric water vapor and temperature profiles, is used to derive the correction coefficients for each spectral band.
    \item Planck curve fitting
    
        Planck curves are modeled for gas flares, which appear as gray bodies because they are sub-pixel sources. Therefore the output of the fitting is an estimate of the temperature and an \textit{emission scaling factor} (the emissivity term in the Planck function). The latter is used subsequently to estimate the source area.
    \item Calculation of source area
    
        The source area $S$ is calculated using
        \begin{equation}
            S = \varepsilon A\,,
        \end{equation}
        where $\varepsilon$ is the emission scaling factor and $A$ is the size of the pixel footprint.
    \item Calculation of radiant heat
    
        The radiant heat is calculated using the Stefan--Boltzmann law:
        \begin{equation}
            \mathrm{RH} = \sigma T^4 S\,,
        \end{equation}
        where $\mathrm{RH}$ is the radiant heat in \si{\mega\watt}, $\sigma$ is the Stefan--Boltzmann constant, $T$ is the temperature in \si{\kelvin}, and $S$ is the source area in \si{\square\metre}.
\end{enumerate}
Once $\mathrm{RH}$ is obtained, previous work by \textcite{Elvidge2015MethodsData} developed a calibration for estimating flared gas volume, utilizing nation-level flaring reporting provided by \textcite{Cedigaz2015AnnualCountry} and state-level reporting from Texas and North Dakota. The developed calibration can then be applied to each individual flaring site worldwide for estimation of flared gas volume, etc.

\subsection{Bayesian Inference}
\label{sec:bayes_infer}
Bayesian inference leverages conditional probability theory to establish a formal procedure for learning from data~\autocite{Betancourt2018ConditionalEngineers}. Bayesian models provide full joint probability distributions $p(\mathcal{D},\bm\theta)$ over observable data $\mathcal{D}$ and unobservable model parameters $\bm\theta$. The essence of Bayesian analysis is to obtain the posterior distribution $p(\bm\theta \mid \mathcal{D})$, which characterizes the conditional probability of parameters $\bm\theta$ given some data $\mathcal{D}$. It can be derived through Bayes' theorem:
\begin{subequations}
    \begin{align}
        p(\bm\theta \mid \mathcal{D}) &= \dfrac{p(\mathcal{D} \mid \bm\theta) \, p(\bm\theta)}{p(\mathcal{D})} \label{eq:bayes_thm} \\[0.5ex]
        &= \dfrac{p(\mathcal{D} \mid \bm\theta) \, p(\bm\theta)}{\int p(\mathcal{D} \mid \bm\theta') \, p(\bm\theta')\diff\bm\theta'} \label{eq:posterior-total} \\[0.5ex]
        &\propto p(\mathcal{D} \mid \bm\theta) \, p(\bm\theta)\,,
    \end{align}
\end{subequations}
where $p(\mathcal{D} \mid \bm\theta)$ is the likelihood (also referred to as the observation model) which denotes how likely the data is given a certain set of parameters, and $p(\bm\theta)$ is the prior which models the probability of the parameters before observing any data. The prior encodes domain expertise. Once some observations are given, it is updated into a posterior which quantifies how consistent the model configurations are with both the domain knowledge and the observed data~\autocite{Betancourt2018ConditionalEngineers}. After the posterior is obtained, most if not all inferential questions can then be answered with posterior expectation values of certain functions~\autocite{Betancourt2019ProbabilisticInference}:
\begin{equation}
    \mathds{E}_p [ g(\bm\theta) ]
        =
    \int g(\bm\theta) \, p(\bm\theta \mid \mathcal{D}) \diff\bm\theta \,, \label{eq:infer-question-posterior-expect}
\end{equation}
where $g(\bm{\theta})$ is the function encoding some inferential question (e.g., where in the model configuration space the posterior concentrates).

Predictions can be made in the form of a posterior predictive distribution:
\begin{equation}
    p(y_* \mid \mathbf{x}_*, \mathcal{D}) = \int p(y_* \mid \bm\theta, \mathbf{x}_*) \, p(\bm\theta \mid \mathcal{D}) \diff\bm\theta \,, \label{eq:posterior-pred}
\end{equation}
where $y_*$ is the predictions based on the training set $\mathcal{D}$ for a test input $\mathbf{x}_*$. Essentially this is integrating the prediction $p(y_* \mid \bm\theta, \mathbf{x}_*)$ over the posterior distribution of parameters~\autocite{gpml}. Note that by giving the final results in terms of a probability distribution, richer information and more reliable inferences are accessed compared to merely giving a point estimate through MLE or MAP (as some machine learning models do under the frequentist framework). This is achieved by incorporating into the inference process the uncertainty in the posterior parameter estimate. Other benefits include posterior predictive checks, which are conducted by checking for auto-consistency between generated data ($\mathbf{y}_*$) and observed data ($\mathbf{y}$).

\subsection{Markov Chain Monte Carlo}\label{sec:mcmc}
Many of the integration problems central to Bayesian statistics, including those in \Cref{eq:infer-question-posterior-expect,eq:posterior-pred}, are analytically intractable. A class of sampling algorithms, known as Markov chain Monte Carlo (MCMC), can be applied to approximate these~\autocite{Andrieu2003AnLearning}. Suppose for some function of interest $f(x)$, the objective is to obtain its integral, with respect to a non-standard target distribution $p(x)$ from which samples cannot be drawn directly:
\begin{equation}
    I(f) = \int f(x) \, p(x) \diff x \,.
\end{equation}
By constructing Markov chains that have $p(x)$ as the invariant distribution, MCMC samplers, while traversing the sample space $\mathcal{X}$, are able to generate samples $x^{(i)}$ that mimic samples drawn directly from the target distribution $p(x)$. In other words, this mechanism makes it possible to draw a set of samples $\{ x^{(i)} \}^{N}_{i=1}$ from $p(x)$.

Then, by the Monte Carlo principle, the integral $I(f)$ can be approximated with a sum $I_N(f)$:
\begin{equation}
    I_N(f) = \dfrac{1}{N} \sum\limits_{i=1}^N f(x^{(i)}) \xrightarrow[N\xrightarrow{}\infty]{\text{a.s.}} I(f) = \int f(x) \, p(x) \diff x \,.
\end{equation}
That is, the estimate $I_N(f)$ is unbiased and by the strong law of large numbers, it will converge almost surely (a.s.) to $I(f)$. That's why MCMC is a powerful tool in Bayesian analysis. In practice, the Metropolis-Hastings (MH) algorithm and Gibbs sampling have been popular MCMC methods~\autocite{Andrieu2003AnLearning}, but only when the parameter space is not too high-dimensional~\autocite{mcelreath2020rethink}.

Due to limited computing resources, it is impossible to run Markov chains infinitely long. In other words, inference has to be made based on finitely many draws. One approach, which is effectively leveraged in this research, is to run multiple chains in parallel and monitor various statistics for diagnosing non-convergence. Besides the effective sample size per transition of the Markov chain, the Gelman-Rubin statistic~\autocite{Gelman1992InferenceSequences}, denoted by $\hat{R}$, is used in this dissertation. The $\hat{R}$ statistic quantifies whether the ensemble of Markov chains initialized from diffuse points in parameter space finally converge to the same equilibrium phase~\autocite{Betancourt2017RobustRStan}. When $\hat{R}$ is sufficiently close to $1$ (for example $\hat{R}<1.05$), convergence is declared to be achieved. As an example, \ref{fig:mcmc_converge_demo} presents how four chains are started in different corners but approach stationarity and convergence after a certain number of iterations.

\csmlongfigure[!htbp]{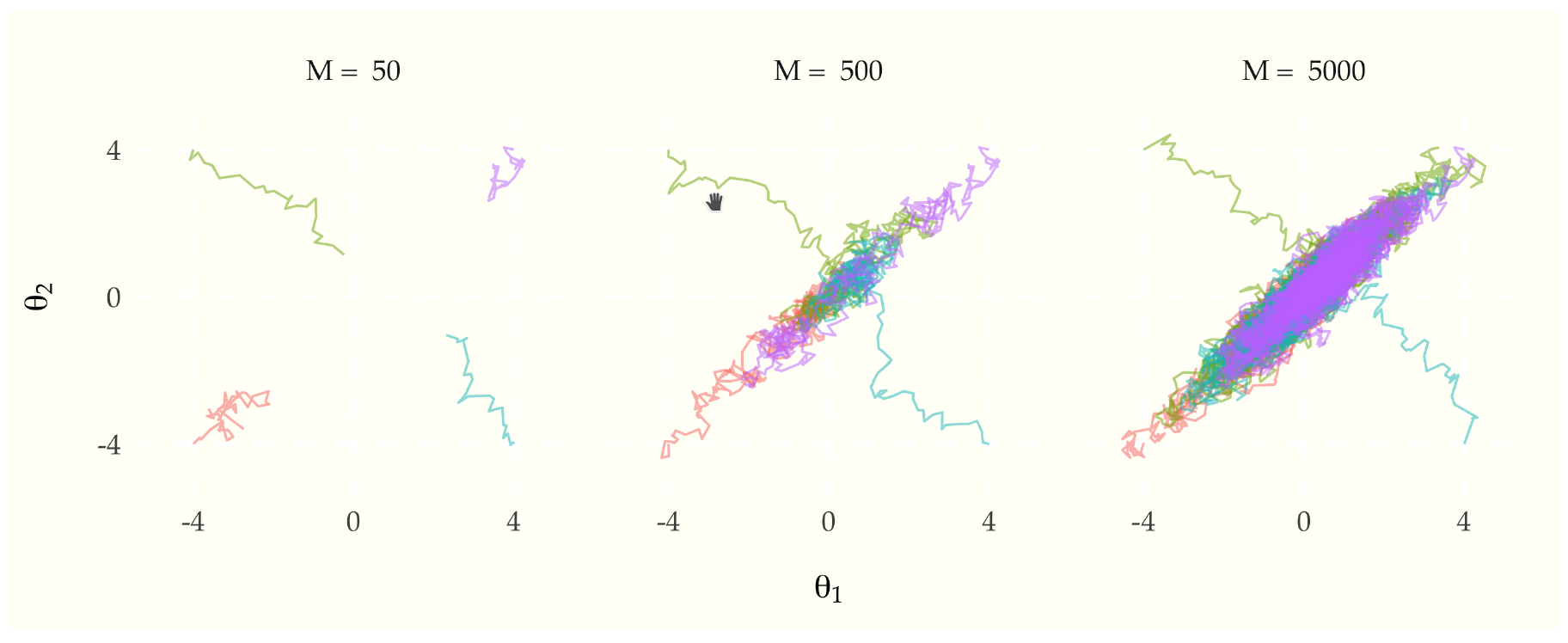}{figures/mcmc_converge_demo}{\textwidth}{The evolution of four random walk Metropolis Markov chains~\autocite{carpenter-prob-stats}, each started in a different location.}{ The target density is a bivariate normal with unit variance and correlation \num{0.9}. After $M=5000$ iterations, the four chains have mixed well and explored most of the target density.}

For many of the problems in practice, including the models in this dissertation, the parameter space is very high-dimensional and involves highly curving regions. The Metropolis-Hastings algorithm and Gibbs sampling are far from efficient in these situations. Hamiltonian Monte Carlo (HMC), originally proposed by \textcite{Duane1987HybridCarlo}, really outshines the other algorithms at this point and is the main sampling strategy adopted in this dissertation. Specifically, No-U-Turn Sampler (NUTS) introduced by \textcite{NUTS}, which is an extension to HMC, is employed for sampling from posterior distributions.

\subsection{Machine Learning}
Machine learning was defined by~\textcite{mitchellMLbook} as computers improving automatically through experience. It can also be viewed as a function estimation problem~\autocite{Vapnik2000TheTheory}, or as the process of extracting important patterns and trends from data~\autocite{hastie2009elements}.

In terms of tasks, common types of learning consist of supervised, unsupervised, semi-supervised, and reinforcement~\autocite{burkov2019hundred_page_ml_book}. Let $\mathbf{x}_i \in \mathcal{X} \subseteq \mathds{R}^d$ represent input, and $y_i \in \mathcal{Y}$ represent target, then the goals of the first two types are:
    \begin{itemize}
        \item Supervised learning aims to use the dataset, consisting of $\mathbf{X}=\{ \mathbf{x}_i \}_{i=1}^{n}$ and $\mathbf{y}=\{ y_i \}_{i=1}^{n}$, to produce a model that is able to predict an output ($y_j$) given some new/unseen input ($\mathbf{x}_j$), i.e., learning the underlying mapping $f\colon \mathcal{X} \to \mathcal{Y}$.
        \item Unsupervised learning is used to find the hidden patterns in $\mathbf{X}$; in this case there does not exist any labels ($\mathbf{y}$) or predefined targets.
    \end{itemize}

Another variation of learning is online learning, in which case training data is fed to the algorithm continuously or one example at a time~\autocite{aml_book}. In other words, streaming data is available that the algorithm has to process on the run. This is different from batch learning, where data is provided beforehand and ``frozen'' during the learning process. Online learning can be applied to the different tasks as discussed above (supervised and others). 

In terms of model characteristics, machine learning models can be categorized into parametric and nonparametric models. Parametric models are characterized by a fixed number of parameters, whereas nonparametric models have an infinite-dimensional parameter space. For example, in the latter case the parameter space can be the set of continuous functions in a regression setting~\autocite{Orbanz2010}. In this dissertation, supervised and unsupervised learning are leveraged while exploiting both parametric and nonparametric models.

From Bayesian's perspective, machine learning is essentially computing the posterior~\autocite{deFreitas2013UBCLearning}, which is then used for inference and prediction tasks. This is conducted exactly through \Cref{eq:bayes_thm}. In practice, machine learning conducted under Bayesian's framework follows a principled workflow (\ref{fig:bayes_workflow}), which is adapted for the modeling in this dissertation.

\csmfigure[!htbp]{bayes_workflow}{figures/workflow_flowchart}{\textwidth}{The flowchart adapted from~\autocite{Betancourt2020TowardsWorkflow} shows a principled Bayesian workflow.}

\subsection{Analytics Toolset}
For the past five to ten years, prosperity in contributions and progress in the open source community has been witnessed. Ecosystems around Python, R, and Julia have been prototyped, tested, and deployed in production environments in various industries. Powerful probabilistic programming languages (PPL), for example Stan~\autocite{mc_stan} and PyMC3~\autocite{Salvatier2016ProbabilisticPyMC3}, have become the workhorse for Bayesian machine learning.

The majority of this work is implemented in Python. Specifically, Bayesian learning is performed by leveraging PyMC3. Some analytic visualizations are produced employing ArviZ~\autocite{Kumar2019ArviZPython}. Geospatial operations are performed with the help of GeoPandas~\autocite{geopandas}. Satellite imagery is processed and analyzed in MATLAB, with implementations mainly following \textcite{Elvidge2013VIIRSNight}.

\chapter{Data Preprocessing and Exploratory Data Analysis}\label{ch:data_pre_eda}
In this chapter, an overview of the flaring data is given. Some other variables which might be correlated with the flaring statistics are also considered. Exploratory data analysis is performed for choosing the subset of the variables as the focus in this dissertation. A state level model is developed in the end which motivates the work in the next two chapters.

\subsection{Data Gathering}
Four sources of data, L8 satellite images, VIIRS estimated flared volumes, NDIC monthly production reports, and county/oilfield shapefiles for North Dakota were gathered for the analysis used in this research.
    \subsubsection{Landsat 8 Images}
    \label{sec:L8-img-specs}
    In total, 167 images (since 2013) were downloaded from Google Cloud using the criteria below:
        \begin{itemize}
            \item From five Path/Row's: 126/216, 126/217, 126/218, 127/216, and 127/217.
            
                According to the Worldwide Reference System (WRS), the satellite imagery of any portion of the world can be queried using Path and Row numbers. These five Path/Row's cover the majority of the areas in North Dakota that have production and flaring activities.
            \item Nighttime images.
            
                Only nocturnal Landsat 8 imagery are used for the purpose of flare detection. 
            \item Cloud cover less than \SI{10}{\percent}.
            
                Images with low cloud cover percentages reveal more clearly land features including gas flares, and thus are ideal for validating the developed methodologies.
            \item GeoTIFF Data Product.
            
                Both the georeferencing information and the raw images of all the spectral bands are preserved through the GeoTIFF format, which are necessary for the analysis.
        \end{itemize}
    
    \subsubsection{VIIRS Estimated Volumes}
    The VIIRS flare inventory and estimated volume dataset obtained from Mikhail N. Zhizhin (personal communication) are used in this dissertation. This dataset includes monthly flare detection records in North America from March 2012 to December 2018 (both inclusive) with their associated:
        \begin{itemize}
            \item Timestamps giving the specific month
            \item Latitudes and longitudes in WGS 84 coordinates
            \item Flared volume estimations in \si{\bcm}
        \end{itemize}

    \subsubsection{NDIC Monthly Production Reports}
\begin{sidewaysfigure}[!htbp]
        \includegraphics[width=\textwidth]{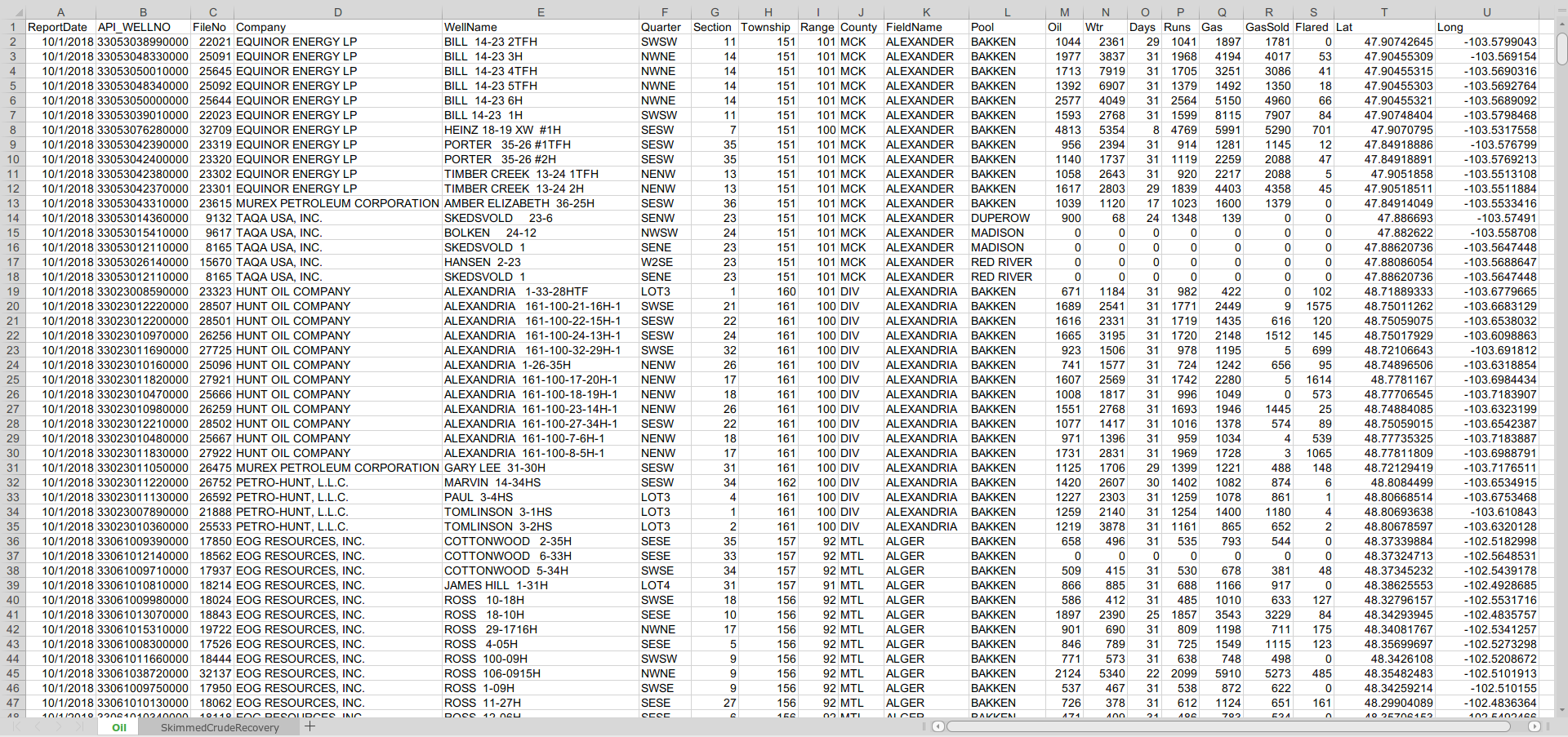}
        \caption{A screenshot of the top ${\sim}50$ rows in the October 2018 production report. Each row corresponds to a well. There are in total \num{17135} rows in this spreadsheet, with the first row being the header.}
        \label{fig:ndic_mp_rep}
    \end{sidewaysfigure}

    All the monthly production reports from May 2015 to April 2020 (both inclusive) which have flaring information have been downloaded from NDIC. There is one Excel spreadsheet per month; each row corresponds to a well (that was active in that month), and columns are for various types of information, including flared gas volume (estimated and reported by operator), oilfield, oil production, etc. A screenshot of the top ${\sim}50$ rows in one of the spreadsheets is displayed in \ref{fig:ndic_mp_rep}.

    \subsubsection{NDIC Shapefiles}\label{sec:shp_file}
    The shapefiles for the counties and oilfields in North Dakota are downloaded from the NDIC GIS Map Server. All the polygons are described in NAD 27 coordinates. The shapefiles are for reverse geocoding the satellite detection locations to readable addresses, specifically which county and oilfield is a flare located in.

\subsection{Satellite Image Processing}\label{sec:l8_glow}
\begin{figure}[!htbp]
	\begin{center}
		\subfigure[VIIRS performance~\autocite{Elvidge2019ExtendingData}]{
			\includegraphics[height=0.40\textheight]{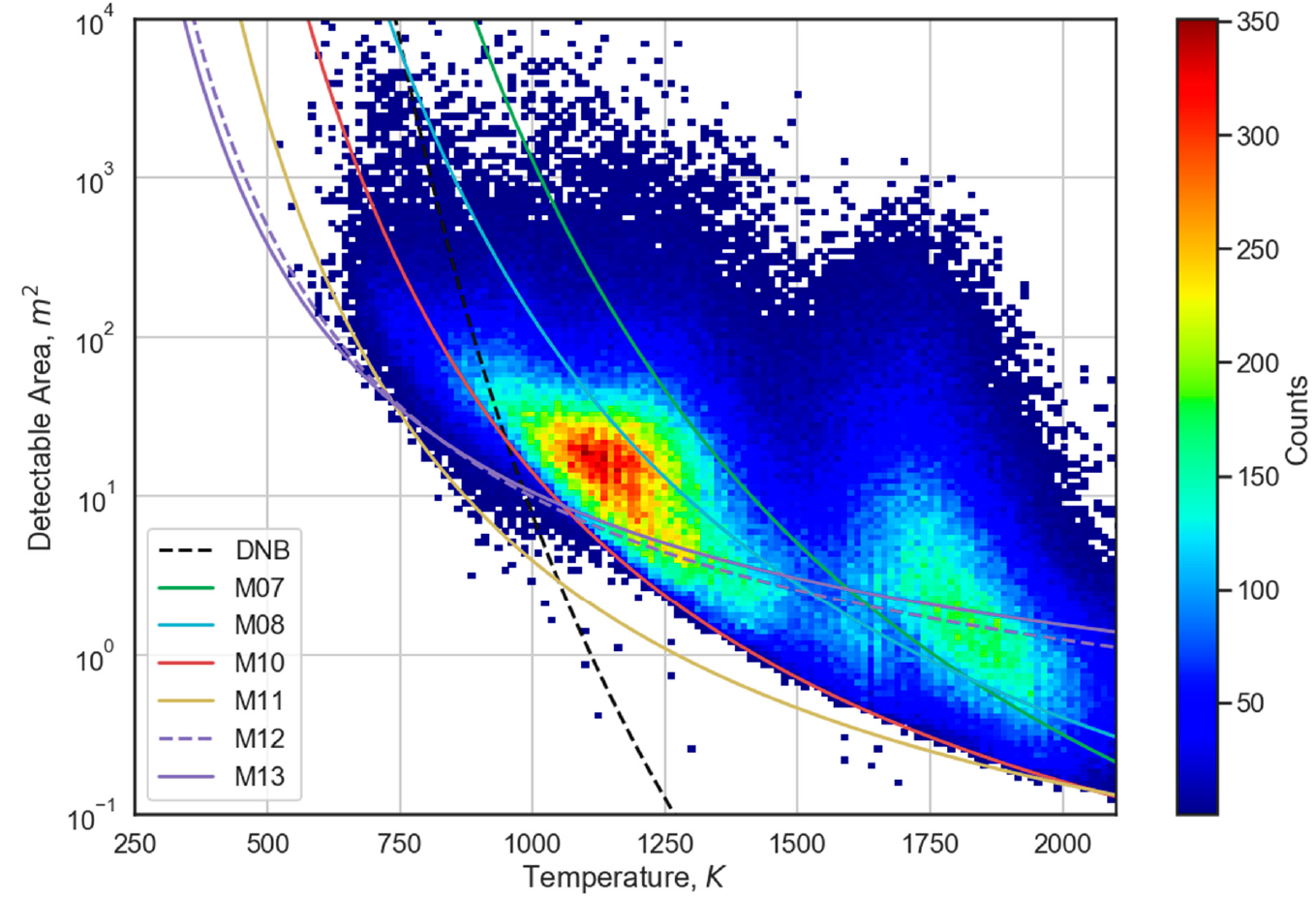}\label{fig:viirs-detect-lim}
		} \\
		\subfigure[L8 performance; figure provided by Mikhail N. Zhizhin (personal communication)]{
			\includegraphics[height=0.40\textheight]{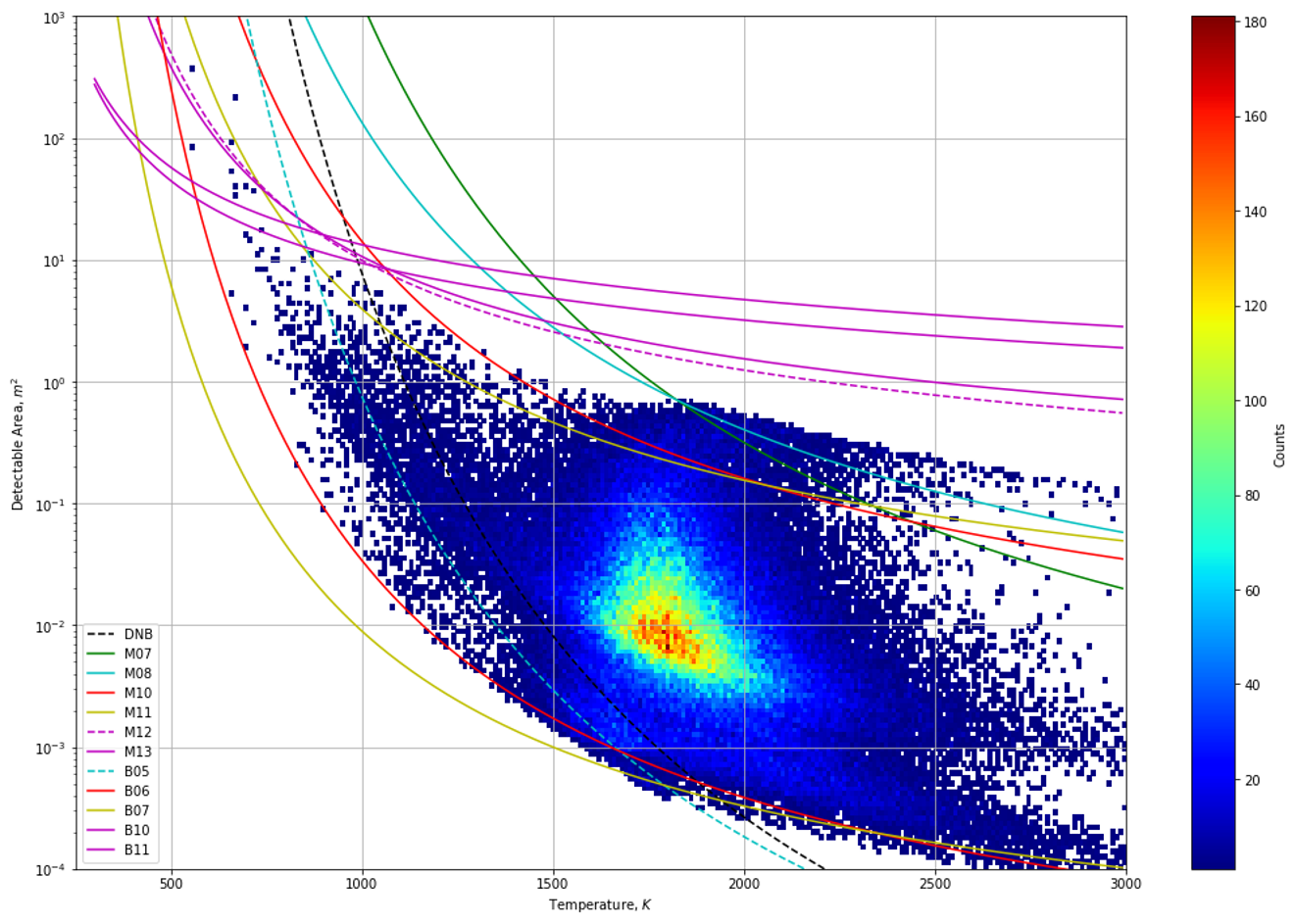}\label{fig:L8-detect-lim}
		}
		\caption{\label{fig:viirs-L8-detect-lim}The nighttime combustion source detection limits of VIIRS (top) and L8 (bottom). For natural gas flaring whose temperature is generally greater than \SI{1500}{\kelvin}, L8 detected flares show source areas (around \SI{d-2}{\square\metre}) orders of magnitude less than that of VIIRS (around \SI{d0}{\square\metre}).}
	\end{center}
\end{figure}

As discussed in Section~\ref{sec:L8-img-specs}, all the available L8 images have been downloaded. They are processed in batch, following the workflow as outlined in Section~\ref{sec:vnf_process_steps}. To compare and contrast with VIIRS' performance, specifically the nighttime combustion source detection limits, all the flares detected from all of the L8 images are gathered and used to generate the source area versus temperature scattergram shown in \ref{fig:viirs-L8-detect-lim}.

Although it is expected that L8 would pick up smaller flares than VIIRS (which is capable of detecting flares around the size of a whole cooktop area), the majority of the detections as indicated on the scattergram are too small for natural gas flaring. To verify if some hot pixels are clustered together and actually representing a single flare or flaring site, HDBSCAN~\autocite{hdbscan} with an implementation due to \textcite{McInnes2017} is executed on every L8 detection map to find out if large blobs of hot pixels are present. HDBSCAN is a density-based clustering algorithm which keeps all the advantages of the original DBSCAN~\autocite{dbscan}, for example the capacity of finding clusters of arbitrary shapes. It also outperforms DBSCAN by being able to build clusters of varying density~\autocite{burkov2019hundred_page_ml_book}. Further, to get the most accurate results in this case, haversine metric is chosen to handle the great-circle distances between the hot pixels; leaf clustering is used instead of the default Excess of Mass method to produce more fine grained clusters. The clustering results are illustrated in \ref{fig:L8_cluster_count_plt}.
\csmlongfigure[!htbp]{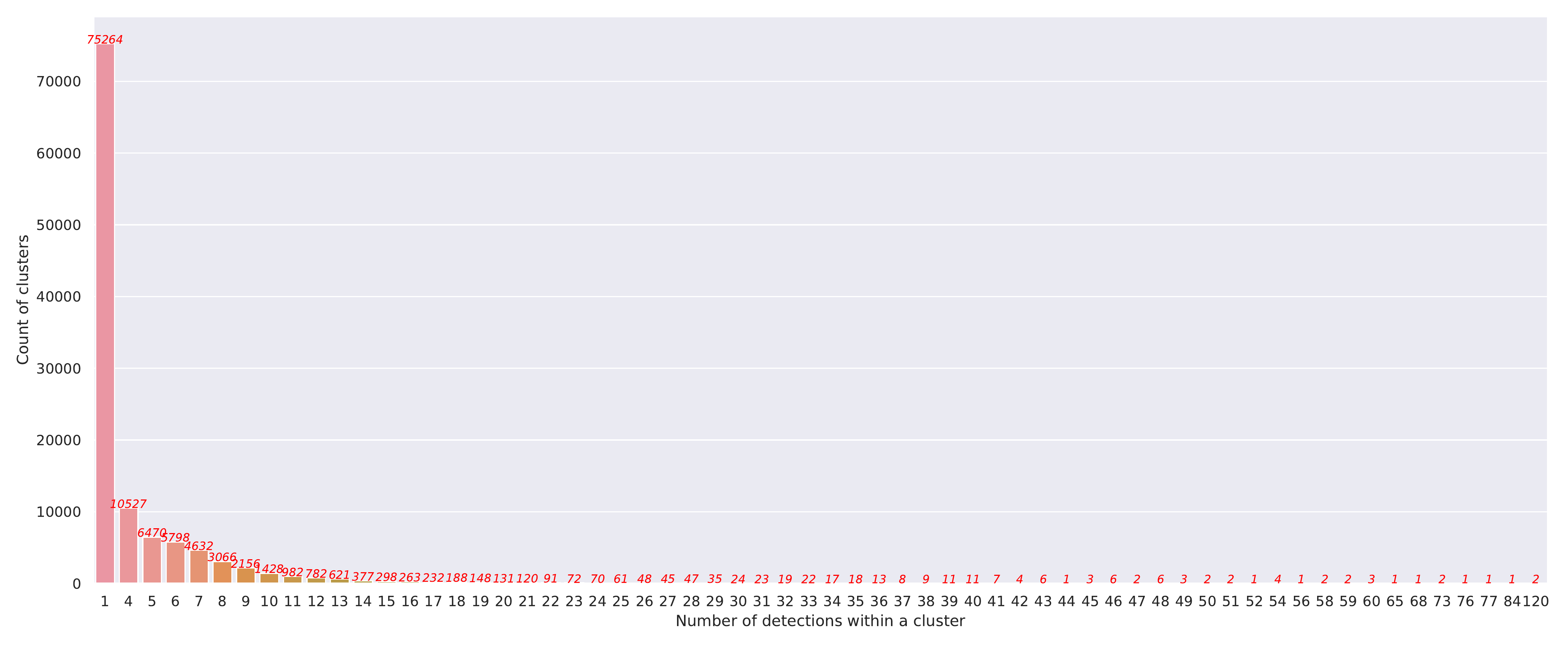}{figures/L8_cluster_count_plt}{\textwidth}{A count plot showing the distribution of cluster sizes: clearly there are a certain number of large clusters (as shown by the tail to the right).}{ For example, there exists \num{2} clusters each of which contains \num{120} hot pixels and there is one cluster with \num{84} hot pixels.}

\begin{figure}[!htbp]
	\begin{center}
		\subfigure[Band 6 (SWIR)]{
			\includegraphics[height=5cm]{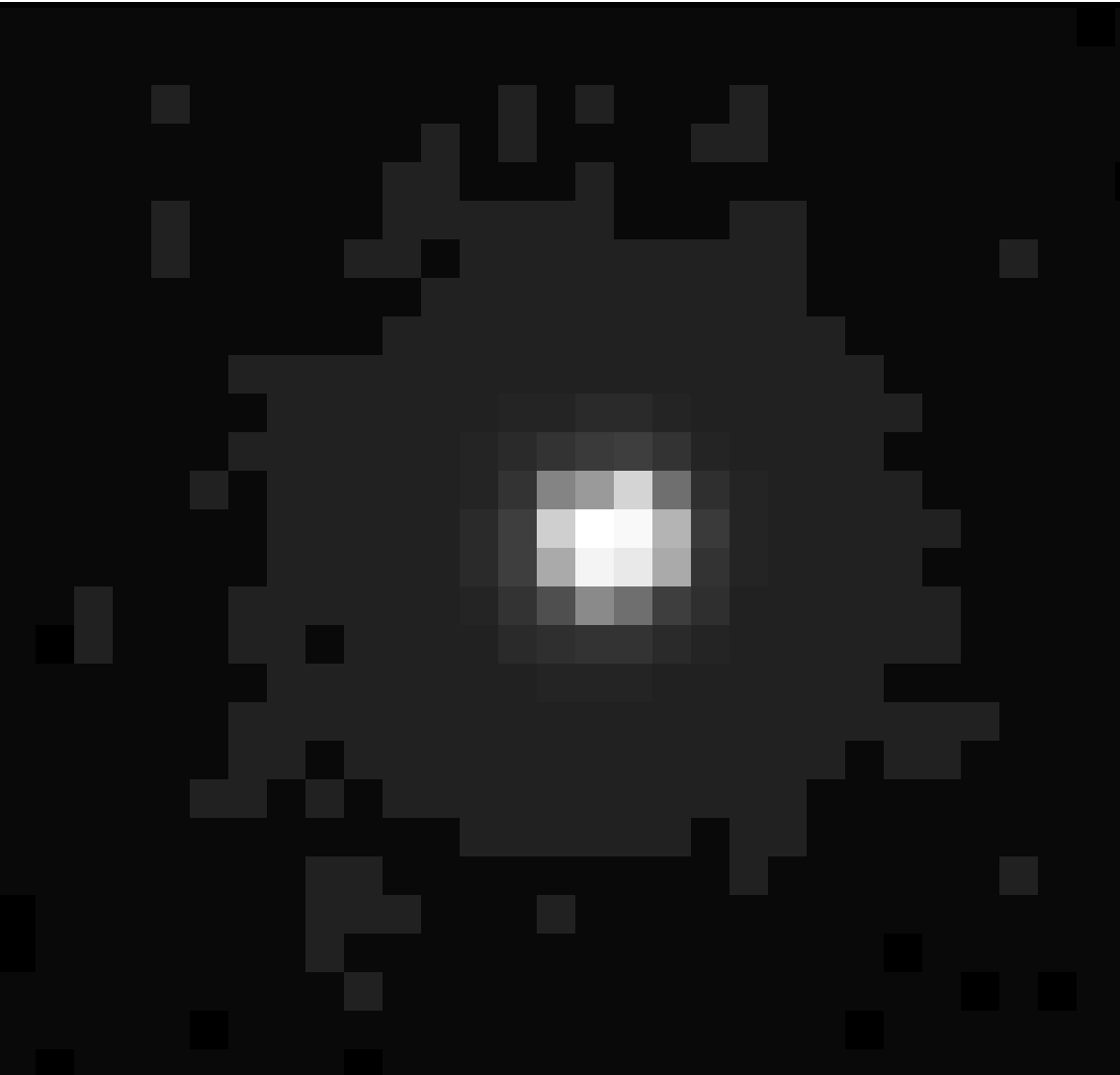}
		}\hspace{0.1\textwidth}
		\subfigure[KMZ view]{
			\includegraphics[height=5cm]{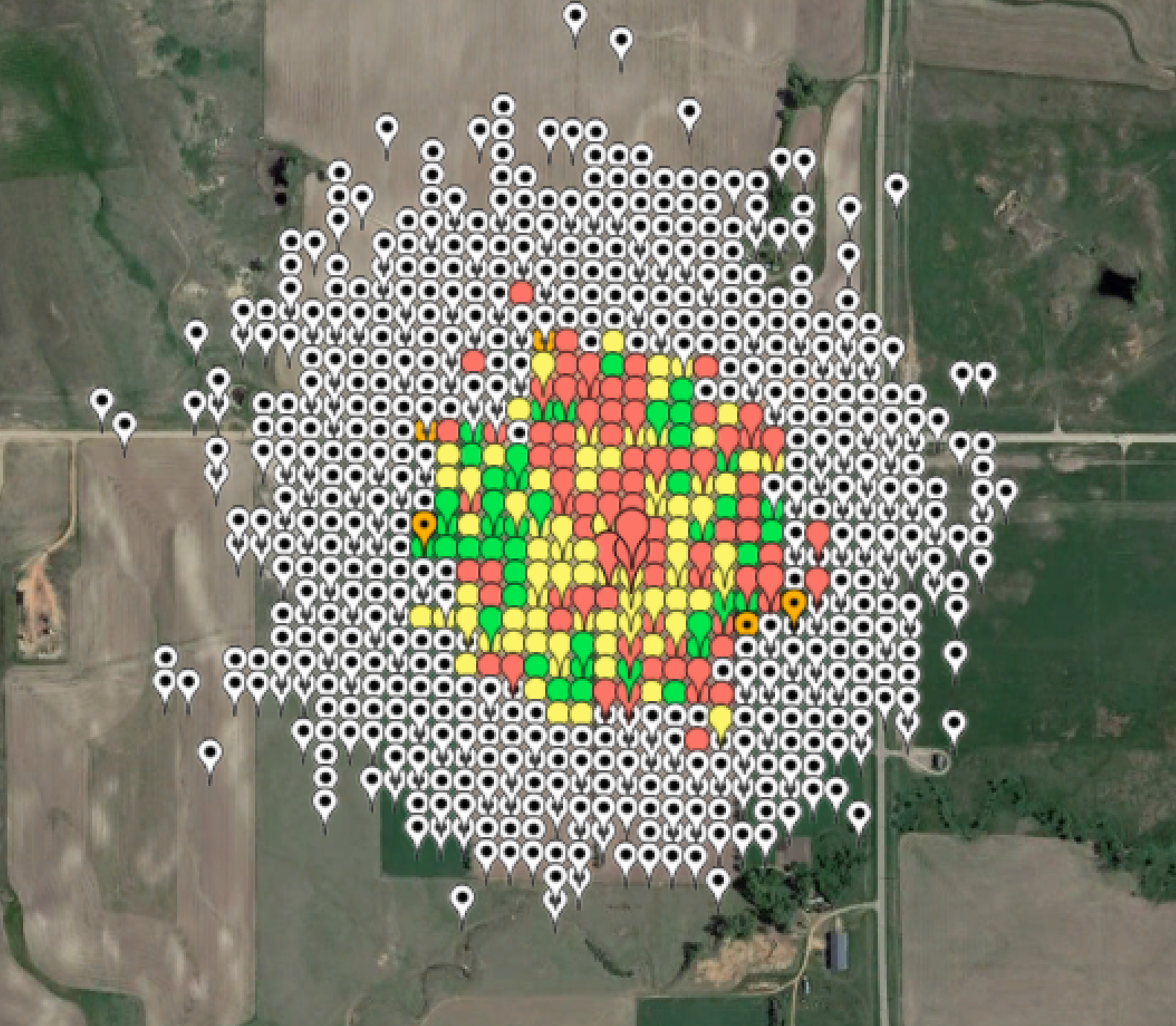}
		}
		\caption{\label{fig:L8-blob}A large flare consisting of many hot pixels (detections), which is found by running the nightfire algorithm on L8 images. Both the Band 6 (grayscale image) and the KMZ view are shown and provided by Christopher D. Elvidge (personal communication).}
	\end{center}
\end{figure}

To verify whether these clusters are really single flares or they are actually a large number of neighboring wells (in which case each hot pixel still represents an individual flare), they are tracked down by looking further into each detection map (KMZ file). It is found that some large blobs of hot pixels are clustered and indeed represent single (huge) flares. One of the examples is shown in \ref{fig:L8-blob}. This poses a challenge to situations where an accurate estimate of the flare count is needed.

The reason for this processing artifact is that, for large flares, there is glow surrounding the flare that was treated as many individual combustion sources. There are potential approaches to mitigate this to make the interpretation and estimation out of L8 more accurate. In this work, the flares detected from VIIRS and the gas volumes estimated out of those are the focus for analytics.

\FloatBarrier

\subsection{Reverse Geocoding}\label{sec:rev_geocode}
By reverse geocoding, the county information of every VIIRS flare that is in North Dakota can be retrieved. For most of the flares, the oilfield information is also retrievable. Thereafter, the flaring statistics from VIIRS and NDIC can be compared and contrasted at different levels, for a certain point or period of time.

Shapefiles as discussed in Section~\ref{sec:shp_file} are used. With the help of GeoPandas, the procedures for extracting counties and oilfields are the same:
\begin{enumerate}
    \item Read the VIIRS records into a geospatial data object, with their original coordinates in WGS 84.
    \item Read the shapefile into a geospatial data object, with its original coordinates in NAD 27.
    \item Transform all the geometries in the shapefile to WGS 84 coordinates.
    \item Perform a spatial join of the two data objects to get the county or oilfield information for each flare, if a specific county/oilfield's polygon and the flare intersect, i.e., having any boundary or interior point in common.
\end{enumerate}

\subsection{Correlational Analysis}\label{sec:corr_analysis}
To study the correlations between oil/gas prices, flaring statistics, and production performance, various time series are extracted for May 2015 to December 2018 (both inclusive). The below list describes all the variables used with their associated labels:
\begin{labeling}[\enskip]{NDIC flaring well count}
    \item[VIIRS flared vol] monthly flared gas volume from VIIRS
    \item[NDIC flared vol] monthly flared gas volume from NDIC
    \item[WTI oil price] WTI crude oil price given by \textcite{EIA2020PetroleumPrices}
    \item[Henry Hub gas price] Henry Hub natural gas price given by \textcite{EIA2020NaturalPrices}
    \item[NDIC oil prod] monthly oil production from NDIC
    \item[NDIC gas prod] monthly gas production from NDIC
    \item[VIIRS flare count] monthly flare detections count from VIIRS
    \item[NDIC flaring well count] monthly wells count which conduct flaring from NDIC
    \item[NDIC GOR] ratio of the NDIC gas production to the NDIC oil production
\end{labeling}

First, the monthly observations are extracted from each time series, and Spearman's $\rho$ is employed to measure the statistical dependence between the variables. Spearman's $\rho$ is a rank correlation, which quantifies the correlation between the rankings of two variables. Compared to Pearson's $r$, it assesses monotonic relationships which can be nonlinear and is more robust to outliers, therefore is used in this section. The pairwise correlations between the variables are presented in \ref{fig:oil_pr_corr}. Since a correlation matrix is always symmetric with unit diagonals, only the lower triangular part without the diagonal is plotted to minimize the information redundancy.
\csmlongfigure[!htbp]{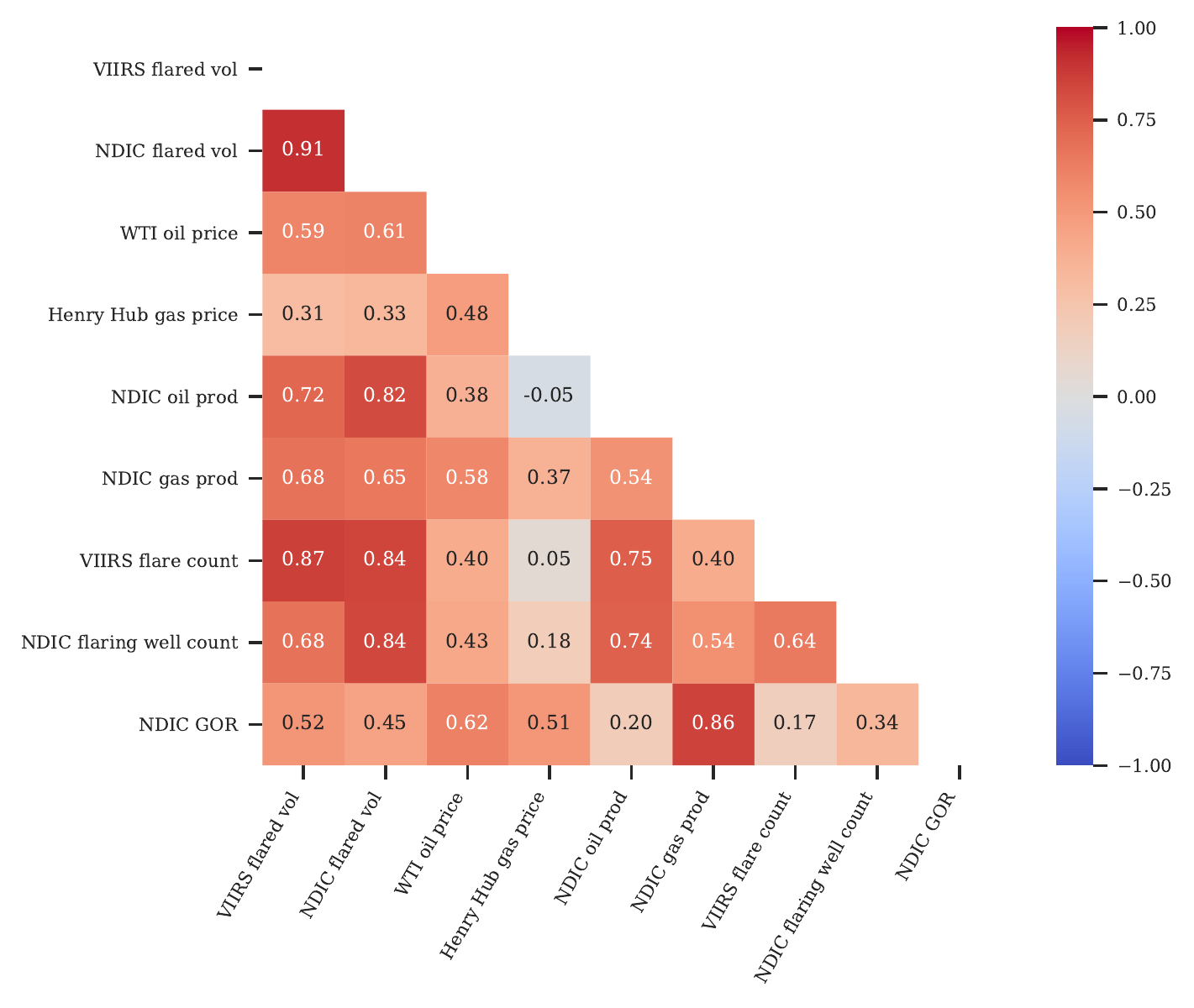}{figures/oil_pr_corr}{\textwidth}{A heat map showing the pairwise Spearman correlations between the original time series' monthly observations.}{ The values are annotated in each cell, the corresponding variables of which can be obtained by reading off the tick labels from the vertical and horizontal axes.}

It can be observed that most pairs show positive correlations. Financial factors (i.e., the oil and gas prices) are not among any of the highly correlated pairs (e.g., above \num{0.80}). Nevertheless, it is indicated that the NDIC and VIIRS reportings have a positive correlation, and oil production is positively correlated with flared gas volume.

In this analysis, due to the nature of the procedure (i.e., extract the monthly data and then measure the rank correlations), all the information on the time scale is neglected. To explore the correlations in the context of time series, the first differences (i.e., lag-$1$ differences) are taken for each variable
\begin{equation}
    y'_t = y_t - y_{t-1},
\end{equation}
and then pairwise Spearman's $\rho$ is evaluated and visualized in \ref{fig:oil_pr_corr_lag1}. In this case, there aren't many pairs of variables which are highly correlated, except the oil and gas production are shown to be monotonically related on the lag-$1$ differences, which is unsurprising. In the remainder of this dissertation, the focus is put on flaring and production related statistics instead of the financial factors. 
\csmlongfigure[!htbp]{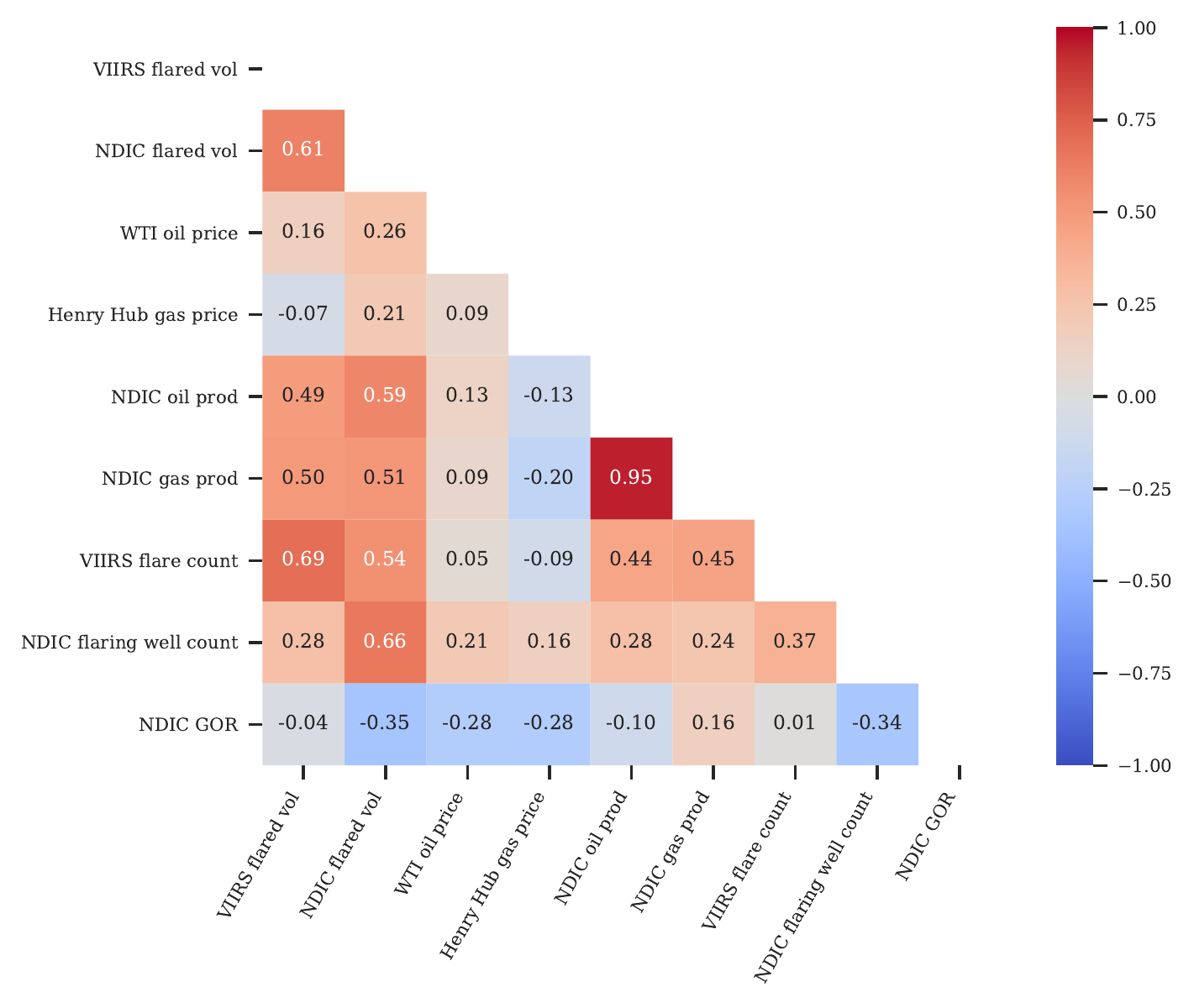}{figures/oil_pr_corr_lag1}{\textwidth}{A heat map showing the pairwise Spearman correlations between the time series after applying the first differences.}{ The values are annotated in each cell, the corresponding variables of which can be obtained by reading off the tick labels from the vertical and horizontal axes.}

\FloatBarrier

\subsection{State Level Flaring Model}
In this section, a regression model is built for the purpose of investigating the statistical relationships between the NDIC and VIIRS reportings. Data from both sources are visualized in \ref{fig:eda_state_data_vis}, which demonstrate a positive correlation.
\csmlongfigure[!htbp]{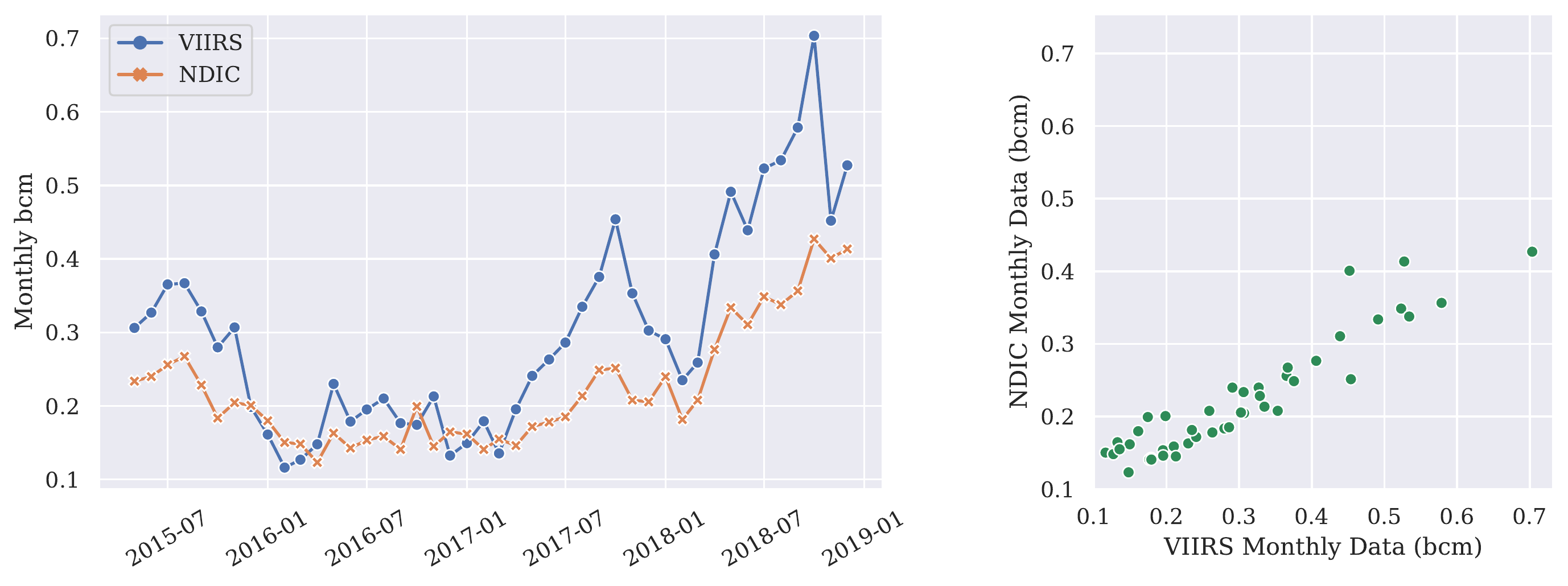}{figures/eda_state_data_vis}{\textwidth}{Visualizations of both the NDIC and VIIRS reportings.}{ Left figure shows the time series. Right figure presents the scatterplot using the data points of each month.}

Assuming a Gaussian observation model for the NDIC reporting with the location parameter encoding VIIRS' information, the model is specified through \crefrange{eda_state_begin}{eda_state_end}:

\begin{subequations}\label{mod:state_lin_mod}
    \begin{align}
        \alpha &\sim \operatorname{Half-Normal}(0.2) \label{eda_state_begin} \\[0.5ex]
        \beta &\sim \operatorname{Gamma}(2, 2) \\[0.5ex]
        \sigma &\sim \operatorname{Half-Cauchy}(0.1) \\[0.5ex]
        \mu_i &= \alpha + \beta \times \mathrm{VIIRS}_i \label{eq:state_lin_mod}\\[0.5ex]
        \mathrm{NDIC}_i &\sim \mathcal{N}(\mu_i, \sigma) \label{eda_state_end}
    \end{align}
\end{subequations}
where $\alpha$ is the intercept and $\beta$ is the slope, both of which are constrained to be non-negative based on the nature of flaring volume; $\sigma$ is the standard deviation in the Gaussian likelihood function, which has to be non-negative as well; $\mu_i$ is the expected NDIC reporting of month $i$, while $\mathrm{NDIC}_i$ and $\mathrm{VIIRS}_i$ are the observed data (i.e., reported volumes) from NDIC and VIIRS in month $i$, respectively. The notation used in defining this model communicates the data generating process unambiguously and is adopted throughout this dissertation. Priors and hyperpriors are on the top while the observation model is at the bottom. The prior distributions for this model and all the others in this dissertation are chosen following the principles below:
\begin{enumerate}
    \item Prefer weakly informative priors, i.e., choose the priors based on the domain expertise at hand before observing any data. They should be strong enough to reflect the domain expertise and be weak enough to ``let the data speak'', i.e., let the likelihood dominate when there is a decent amount of data. For example, a prior of a gamma distribution with mean $\mathds{E}\beta=2/2=1$ is placed on $\beta$, reflecting the assumption that the satellite interpretation workflow gives the same flared volume as the NDIC reporting, before one observes any data.
    \item Prefer priors with soft constraints as opposed to hard constraints, i.e., follow Cromwell's rule. For example, $\alpha$, $\beta$ and $\sigma$ all have prior distributions with support on $\mathds{R}_{>0}$ or $\mathds{R}_{\geq 0}$. Counterexamples include using a triangular distribution or a continuous uniform distribution as the prior for such quantities, for which the author does not recommend.
    \item Prefer maximum entropy distributions, i.e., make the most conservative assumptions based on all the information at hand (obeying all the known constraints). For example, the Gaussian and the binomial distributions are maximum entropy distributions and used in this dissertation, the fact of which can be formally shown leveraging the definition of Kullback--Leibler (KL) divergence.
\end{enumerate}

Once the priors and likelihood are established, four Markov chains of Hamiltonian Monte Carlo are run in parallel to sample from the posterior. The parameter estimates are reported in \ref{tab:eda_st_param}, and the posterior distributions and trace plots are presented in \ref{fig:eda_state_traceplt}. The four chains are plotted separately with different colors. The $x$-axis of the trace plot shows the number of iterations. This layout is used consistently for the remainder of this dissertation.

\begin{table}[!htbp]
\centering
\caption{Parameter Estimates of State Level Flaring Model}
\label{tab:eda_st_param}
\begin{tabular}{c l S[table-format=1.3] c}
\toprule
Parameter & Variable & {Point Estimate} & \SI{90}{\percent} Credible Interval \\ \midrule
$\alpha$ & Intercept & 0.061 & (0.044, 0.079) \\
$\beta$ & Slope & 0.535 & (0.482, 0.590) \\
$\sigma$ & Reporting variability & 0.030 & (0.024, 0.035) \\ \bottomrule
\end{tabular}
\end{table}

\csmlongfigure[!htbp]{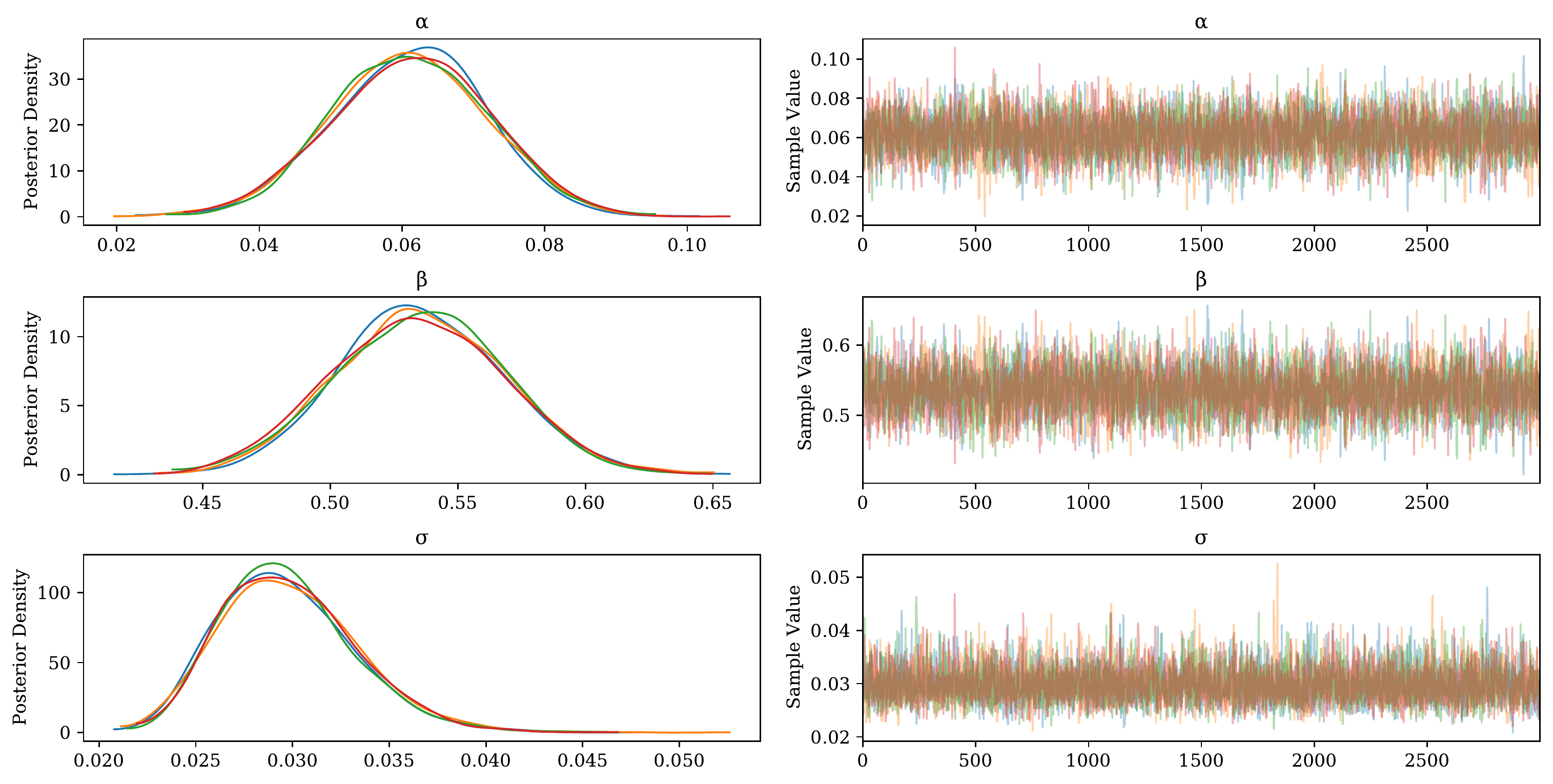}{figures/eda_state_traceplt}{\textwidth}{Posterior distributions (left column) and trace plots (right column) for the state level flaring model.}{ Well mixing and convergence of the Markov chains have been achieved as shown by the trace plots.}

Utilizing the model and the trace, posterior predictive samples are generated to construct the intervals (\ref{fig:eda_state_ppc}). Point estimates and point predictions are easy to obtain for a certain machine learning model, however it is the properly constructed intervals that will provide insights into the uncertainty for decision making. The author would like to emphasize the importance of quantifying uncertainties when using machine learning, no matter for inference, prediction, or building intermediate models for integration into physics-based models. This is unfortunately neglected or ignored in some of the applications/publications in the petroleum engineering domain. The importance of properly quantifying the uncertainties will also be stressed in the following chapters.

\csmlongfigure[!htbp]{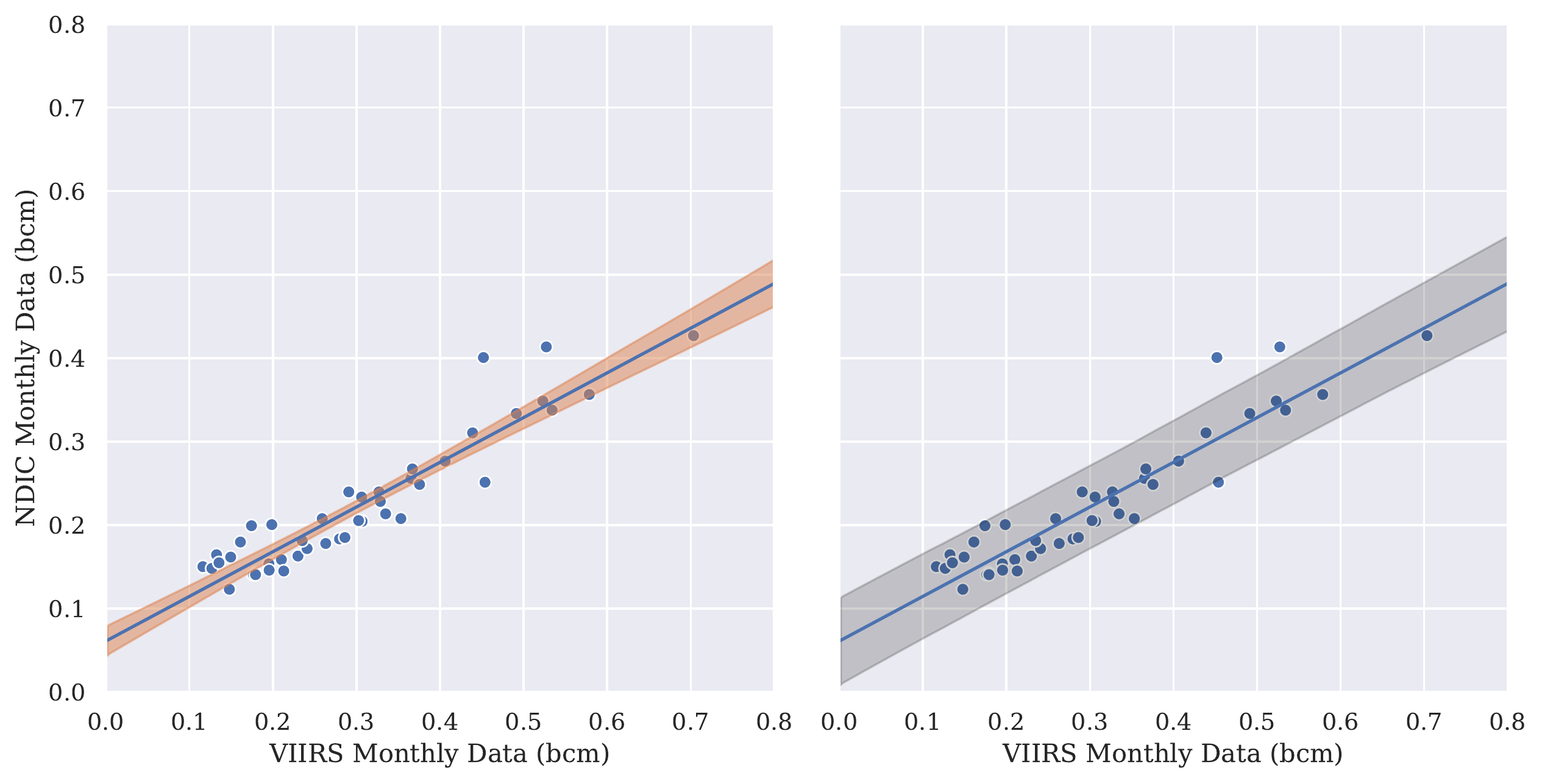}{figures/eda_state_ppc}{\textwidth}{Intervals are constructed using posterior predictive samples.}{ In both figures, the line shows the ``best'' fit using point estimates (posterior means) of $\alpha$ and $\beta$. Shaded area in the left figure presents the \SI{90}{\percent} credible interval (CI) of the regression mean. Shaded area in the right figure demonstrates the \SI{90}{\percent} prediction interval for the future NDIC reporting, for which most of the existing observations fall within.}

Whenever only one model specification is needed for making point predictions, it can be recovered by the parameter estimates from \ref{tab:eda_st_param}:
\begin{equation}\label{eq:state_point_est_mod}
    \mathrm{NDIC}_i = 0.061 + 0.535 \times \mathrm{VIIRS}_i \, ,
\end{equation}
where $\mathrm{NDIC}_i$ and $\mathrm{VIIRS}_i$ are flared volumes in bcm of month $i$. The model also provides clear interpretations for the NDIC reporting regression mean, on the whole state level:
\begin{enumerate}
    \item The intercept indicates on average there is \SI{90}{\percent} probability that \SIrange{0.04}{0.08}{\bcm} reported volume per month will not be captured by the current VIIRS processing workflow. The posterior mean is \SI{0.061}{\bcm} ($\approx \SI{2150}{\mmcf}$).
    \item The slope indicates on average when satellite estimated volume increases by one unit, under \SI{90}{\percent} probability the NDIC reporting will increase by \num{0.48} unit to \num{0.59} unit. The posterior mean is \num{0.535} unit.
\end{enumerate}

This model, while serving as a decent calibration and estimation tool for NDIC reporting on the state level, makes the assumption that the heterogeneity within the state (e.g., among different counties) is negligible and all the monthly observations are conditionally independent and identically distributed (i.i.d.). For the scenarios in which these assumptions do not hold, other types of models can be built and are discussed in Chapter~\ref{ch:hier} and Chapter~\ref{ch:gp}.

\chapter{County Level Flaring Model}\label{ch:hier}

\setlength{\epigraphwidth}{0.7\textwidth}
\setlength{\epigraphrule}{0pt}
\epigraph{``Multilevel regression deserves to be the default form of regression.''}{--- \textcite{mcelreath2016rethink}}

\subsection{Learning the Heterogeneity}
In this chapter, the author explores the heterogeneity in correlations between the state-reported and satellite-detected flaring statistics, among different counties in North Dakota. The motivations are threefold:
    \begin{enumerate}
        \item Provide more granular insights than merely investigating the whole state's flaring statistics.
        \item Compare and contrast different counties' reporting consistencies with the baseline (i.e., the satellite detections).
        \item Develop a dedicated model for each county for calibration and prediction purposes.
    \end{enumerate}

\subsection{Hierarchical Model}\label{sec:hier_mod}
A common problem in learning from data is modeling individuals or units of a population. For example, building models for different counties in a state, or for different well pads in an oilfield. Usually from domain expertise, it is expected that the units would demonstrate some differences, however they do not necessarily represent completely independent data generating processes. In other words, the units are different in some ways, while being similar in others. Unfortunately, the following two common modeling approaches are extreme and not ideal:
    \begin{enumerate}
        \item Complete pooling
            \begin{itemize}
                \item This ignores heterogeneity and assumes that the observations from all the units are generated/described by the exact same process. One set of parameters is learned for the whole population. In this situation, the variance might be smaller, however the bias could be huge.
            \end{itemize}
        \item No pooling
            \begin{itemize}
                \item This lets each unit learn its own set of parameters from its own data. The assumption is that the information from each unit tells one nothing about any other unit. In this situation, the bias might be smaller, however the variance could be huge.
            \end{itemize}
    \end{enumerate}

In practice, neither of these approaches will be able to generalize well for insight extraction or prediction tasks, due to the total generalization error being large. In fact, these two extremes can be compromised by explicitly modeling the entire population of units. That is, in order to investigate the correlations among the individual units, an explicit model is introduced for the population. In the learning phase, the individual posteriors are used to fit some population distribution, while the information of the population is then fed back to the individuals. What happens in this case is that the individuals with diffuse likelihood functions (e.g. with less data) are dragged more towards the population distribution, whereas the individuals which are well informed by their data will have their posteriors mostly unchanged. In this process, dynamic regularization is achieved, i.e., the total generalization error is much smaller by partially pooling the data and balancing between the bias and variance.

In the context of county level model development, the question is now how might one model the population. To motivate the choice of a particular class of models, some characteristics of the counties have to be examined. In this work, the counties are considered to be exchangeable, i.e., the joint probability $p(\bm{\theta}_1, \bm{\theta}_2, \dots, \bm{\theta}_n)$ is invariant to permutation of the indices, where $\bm{\theta}_i$, $i=1,2,\dots,n$ is the parameters for the $i$-th county. That is, for any permutation $\pi$,
\begin{equation}
    p(\bm{\theta}_1, \bm{\theta}_2, \dots, \bm{\theta}_n) = p(\bm{\theta}_{\pi_1}, \bm{\theta}_{\pi_2}, \dots, \bm{\theta}_{\pi_n}) \, .
\end{equation}
Furthermore, the list of counties can grow, i.e., although one might only look at a few counties at this point, in the future new counties in terms of flaring activities might be considered.

If a population being modeled is exchangeable, and the population can grow arbitrarily large, de Finetti's theorem shows that the only distribution that respects exchangeability is a hierarchical distribution:
\begin{equation}
    p(\bm{\theta}_1, \bm{\theta}_2, \dots, \bm{\theta}_n) = \int \Biggl[ \prod_{i=1}^{n} p(\bm{\theta}_i \mid \phi) \Biggr] p(\phi) \diff\phi \, ,
\end{equation}
where $\phi$ is a population parameter (which can be generalized to multiple population parameters) and $p(\phi)$ is a population prior. It asserts an important fact that if exchangeable data is used for analytics, there must exist a population model~\autocite{Jordan2010UCBTheorem}. This provides guidance for the development of the county level flaring models in this chapter.

Equivalently, the individual and population parameters can be fitted jointly, achieving a dynamic pooling of the data:
\begin{equation}\label{eq:de_finetti_joint_mod}
    p(\bm{\theta}_1, \bm{\theta}_2, \dots, \bm{\theta}_n, \phi) = \Biggl[ \prod_{i=1}^{n} p(\bm{\theta}_i \mid \phi) \Biggr] p(\phi) \, ,
\end{equation}
in which process not only the $\bm{\theta}$'s but also $\phi$ are learned. After adding the observations component ($\mathcal{D}=\{(\mathbf{x}_j, y_j) \mid j=1, \ldots , m\}$) to it, the joint model becomes:
\begin{equation}\label{eq:hier_gen_form}
    p\!\left( \left\{ y_j, \mathbf{x}_j, \bm{\theta}_{\mathsmaller{\textsc{county}} [j]}, \psi_j \right\}_{j=1}^{m}, \phi \right) = \Biggl[ \prod_{j=1}^{m} p(y_j \mid \mathbf{x}_j, \bm{\theta}_{\mathsmaller{\textsc{county}} [j]}, \psi_j) \, p(\bm{\theta}_{\mathsmaller{\textsc{county}} [j]} \mid \phi) \Biggr] p(\phi) \, ,
\end{equation}
where $\bm{\theta}_{\mathsmaller{\textsc{county}} [j]}$ stands for the parameters for the $j$-th observation based on its county assignment, and $\psi$ are some other parameters in the likelihood function that are not necessarily distributed according to a population model. \Cref{eq:hier_gen_form} characterizes a hierarchical model that fits nicely into the Bayesian framework and is exploited for building the models in this chapter.

As a fundamental approach to model heterogeneity, hierarchical models have been depended upon routinely in various fields including ecological science~\autocite{bolker2008ecological}, political science~\autocite{gelman2006Hierarchical}, and biological science~\autocite{mcelreath2016rethink}. The author believes that they should be widely accepted and utilized in the petroleum engineering domain as well, where the dataset is usually presented in hierarchies. For example, the shale gas wells in a given basin were completed by different oilfield service companies. The information can then be pooled among the service companies. A further discussion is given in Section~\ref{sec:pe_ml_discuss}. One caveat, though, is that de Finetti's theorem is based on the assumption that the population (of units) is exchangeable and can grow arbitrarily large. Just like every other assumption in machine learning, it should not be taken for granted and does not always hold. In the context of county level flaring model development, one might argue that there are currently \num{53} counties in North Dakota and there might not be many new counties (as administrative divisions) in any finite amount of time. In that regard, the author agrees with the claim of \textcite{Box2009StatisticalAdjustment} that, since assumptions ``are never exactly true'', what shall be sought is the \textit{useful} models as opposed to the \textit{correct} ones. That is the goal for applying the hierarchical models in this chapter.

It is worth noting that the terminologies are not consistent when referring to these types of models: some argue that hierarchical model and multilevel model are different names for the same modeling technique~\autocite{bolker2008ecological,mcelreath2016rethink}, while others tried to differentiate them~\autocite{Carpenter2019AllModeling}. In this dissertation, the model assumptions are communicated via the mathematical structures instead of the terminologies, by writing out the full model definitions whenever possible.

\subsection{Data Description}\label{sec:hier_data_descri}
After performing the reverse geocoding as outlined in Section~\ref{sec:rev_geocode}, there are twelve counties found to have reported flaring activities from both VIIRS and NDIC. For each county's historical data from May 2015 to December 2018 (both inclusive), only the months that have reported volumes from both sources are extracted. A scatterplot for each of the 12 counties is presented in \ref{fig:county_scatter}, where the county abbreviations follow the convention from the NDIC monthly production reports. \ref{tab:nd_county_abbrev} lists the full county names associated with each abbreviation.
\csmlongfigure[!htbp]{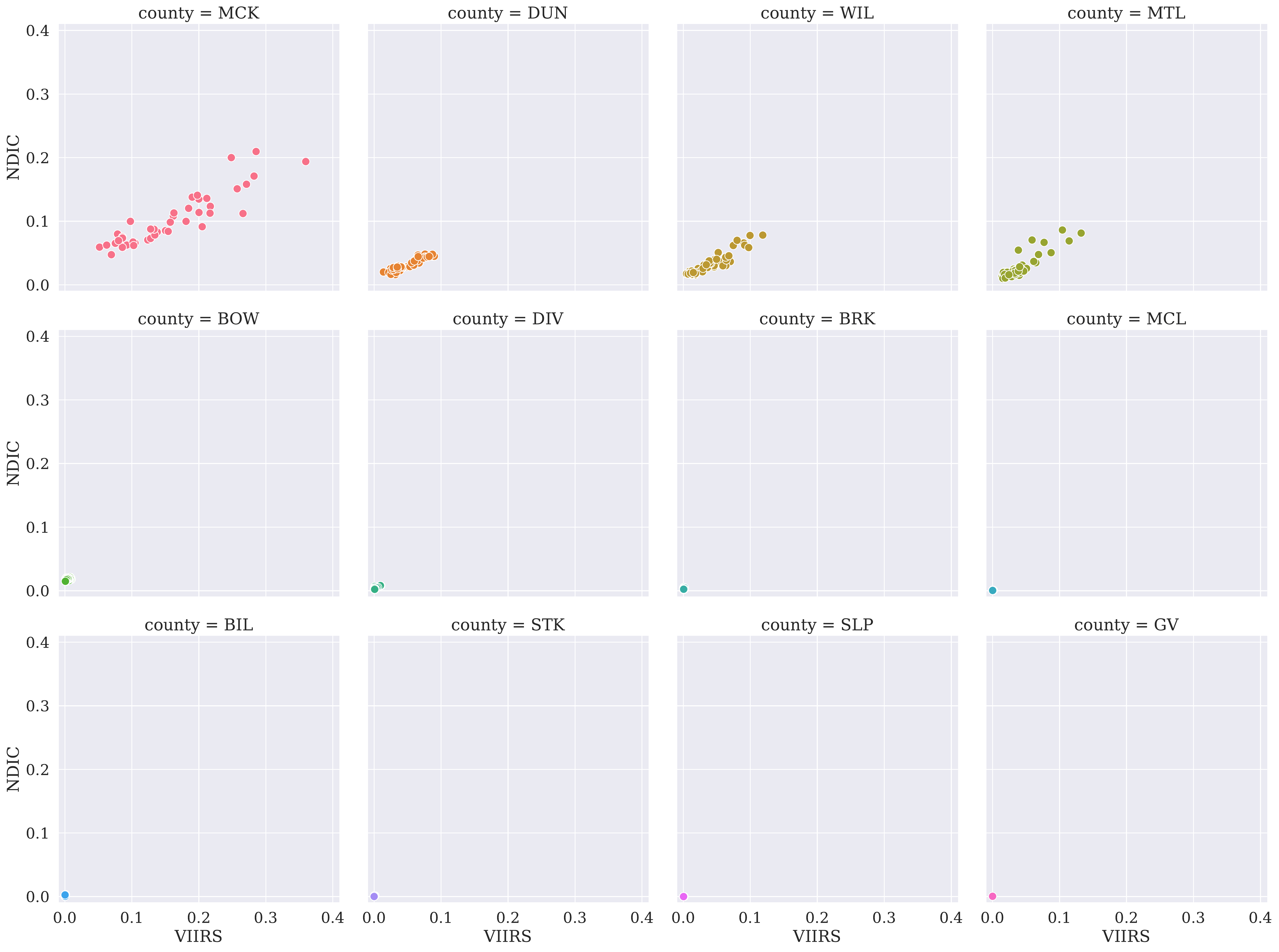}{figures/county_scatter}{\textwidth}{Scatterplots of NDIC and VIIRS reportings for different counties.}{ Both the $x$- and $y$-axis are shared among all the subplots. The $x$-axis is the monthly VIIRS reporting of the flared volume in \si{\bcm}, and the $y$-axis is for the NDIC reporting in the same unit.}
\begin{table}[!htbp]
\centering
\caption{North Dakota County Abbreviations}
\label{tab:nd_county_abbrev}
\begin{tabular}{ll}
\toprule
Abbreviation & County               \\ \midrule
MCK          & McKenzie County      \\
DUN          & Dunn County          \\
WIL          & Williams County      \\
MTL          & Mountrail County     \\
BOW          & Bowman County        \\
DIV          & Divide County        \\
BRK          & Burke County         \\
MCL          & McLean County        \\
BIL          & Billings County      \\
STK          & Stark County         \\
SLP          & Slope County         \\
GV           & Golden Valley County \\ \bottomrule
\end{tabular}
\end{table}

It can be seen that the flaring magnitudes in terms of the flared volumes are quite diverse for the different counties. To better visualize all of them, a zoomed-in view for each county is shown in \ref{fig:county_scatter_ALLzoomed}. It becomes clear that most of the counties except SLP and GV have more than ${\sim}12$ data points; however, only the four counties in the top row (i.e., MCK, DUN, WIL and MTL) have the largest amount of data and indicate stronger positive correlations between VIIRS and NDIC.
\csmlongfigure[!htbp]{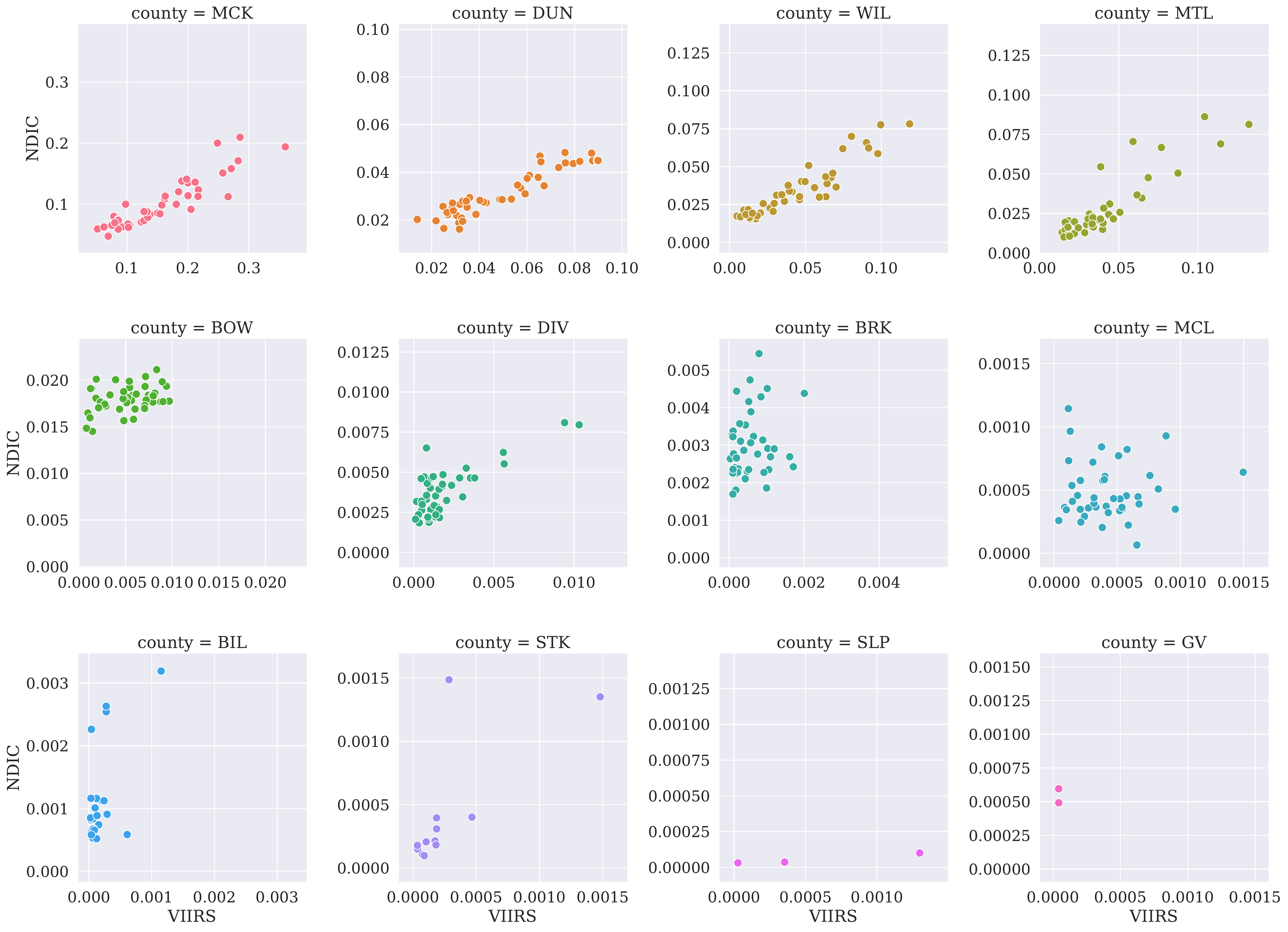}{figures/county_scatter_ALLzoomed}{\textwidth}{Scatterplots of NDIC and VIIRS reportings for different counties, without sharing neither $x$- nor $y$-axis for all the subplots.}{ Within each subplot, equal scaling and limits are set for $x$- and $y$-axis. The axes' meanings are the same as in \ref{fig:county_scatter}.}

For the purpose of building county level models and investigating the heterogeneity among the counties, the no pooling option discussed in the previous section will fail. Especially with counties SLP (which has \num{3} observations) and GV (which has \num{2} observations), if a linear model such as \Cref{eq:state_lin_mod} is fitted, the learned slope parameters $\beta_{\mathsmaller{\textsc{county}}}$ will have point estimates $\hat{\beta}_{\textsc{slp}} \approx 0$ and $\hat{\beta}_{\textsc{gv}} \gg 0$ with their associated samples. The interpretation of the slope parameter (which was discussed right after \Cref{eq:state_point_est_mod}) implies that such inferences are never possible. Some other counties, even with more data points (e.g., MCL), suffer from the noise levels in their observations. Using their own dataset will frustrate accurate inferences. Therefore, in order to build models robustly at a county level, the hierarchical model discussed in the previous section is exploited.


\subsection{Model Specification}
Motivated by the discussions in Section~\ref{sec:hier_mod}, partial pooling is performed by explicitly modeling the entire population of counties. In this way, the counties such as MCL can leverage the information from other counties to learn their own parameters. Counties with ``strong data'' (i.e., very informative data which makes the likelihood dominate the structure of the posterior), such as those in the top row of \ref{fig:county_scatter_ALLzoomed}, indicate a positive correlation between VIIRS and NDIC. Therefore, a similar strategy as in Model~\ref{mod:state_lin_mod} is adopted for the counties, i.e., one set of slope and intercept is learned for each county.

Since the slope and intercept are very interpretable, the meanings of which were discussed right after \Cref{eq:state_point_est_mod}, partial pooling is also enabled across parameter types (i.e., intercepts and slopes). In other words, knowing how much flared volume is missed from VIIRS (i.e., the information carried by the intercept) might improve learning how VIIRS and NDIC will covary (i.e., the information carried by the slope). Specifically, a population model with a multivariate normal density is used for the different counties' parameters.

The hierarchical model is specified through \crefrange{county_cp_begin}{county_cp_end}:
\begin{subequations}\label{mod:county_hier_cp}
    \begin{align}
        \mu_\alpha &\sim \operatorname{Half-Normal}(0.1) \label{county_cp_begin} \\[0.5ex]
        \mu_\beta &\sim \operatorname{Gamma}(2, 2) \\[0.5ex]
        \sigma_\alpha &\sim \operatorname{Half-Normal}(0.1) \\[0.5ex]
        \sigma_\beta &\sim \operatorname{Half-Normal}(0.1) \\[0.5ex]
        \sigma &\sim \operatorname{Half-Normal}(0.05) \\[0.5ex]
        \mathbf{R} &\sim \operatorname{LKJcorr}(2) \label{exp:cp_core_beg} \\[0.5ex]
        \bm{\Sigma} &= \begin{pmatrix} 
                    \sigma_\alpha & 0 \\
                    0 & \sigma_\beta 
                \end{pmatrix} \cdot \mathbf{R} \cdot \begin{pmatrix} 
                                            \sigma_\alpha & 0 \\
                                            0 & \sigma_\beta 
                                        \end{pmatrix} \\[0.5ex]
        \begin{bmatrix} \alpha_{\mathsmaller{\textsc{county}}} \\ \beta_{\mathsmaller{\textsc{county}}} \end{bmatrix} &\sim \operatorname{MVNormal} \!\left( \begin{bmatrix} \mu_\alpha \\ \mu_\beta \end{bmatrix}, \bm{\Sigma} \right) \label{exp:mvn_cp} \\[0.5ex]
        \mu_j &= \alpha_{\mathsmaller{\textsc{county}} [j]} + \beta_{\mathsmaller{\textsc{county}} [j]} \times \mathrm{VIIRS}_j \\[0.5ex]
        \mathrm{NDIC}_j &\sim \mathcal{N}(\mu_j, \sigma) \label{county_cp_end}
    \end{align}
\end{subequations}
where:
\begin{description}[noitemsep,itemindent=-8em,leftmargin=6em]
    \item[\mu_\alpha] is the average intercept for all the counties;
    \item[\mu_\beta] is the average slope for all the counties;
    \item[\sigma_\alpha] is the standard deviation among different counties' intercepts;
    \item[\sigma_\beta] is the standard deviation among different counties' slopes;
    \item[\sigma] is the the standard deviation in NDIC reporting within the counties;
    \item[\mathbf{R}] is the correlation matrix distributed according to an LKJ distribution. It is \num{2}-by-\num{2} in size and encodes the correlation between the intercepts and slopes;
    \item[\bm{\Sigma}] is the covariance matrix for the population model, which is constructed by multiplying the correlation matrix from both sides by a diagonal matrix of standard deviations;
    \item[\alpha_{\mathsmaller{\textsc{county}}}\text{ and }\beta_{\mathsmaller{\textsc{county}}}] are the intercept and slope for each county, whose prior distributions are defined by a two-dimensional Gaussian population model;
    \item[\footnotesize{\textsc{county}}\lbrack j \rbrack] (in the subscript) denotes the county index, i.e., $\footnotesize{\textsc{county}}\lbrack j \rbrack \in \{k\in \mathds{N}_{0} \mid k\leq 11\}$, such that $\alpha_{\mathsmaller{\textsc{county}}[j]}\text{ and }\beta_{\mathsmaller{\textsc{county}}[j]}$ are the intercept and slope for the $j$-th observation based on its county assignment;
    \item[\mathrm{VIIRS}_j] is the VIIRS reported volume of the $j$-th observation;
    \item[\mu_j] denotes the underlying flared volume of the $j$-th observation;
    \item[\mathrm{NDIC}_j] is the NDIC reported volume of the $j$-th observation.
\end{description}

The LKJ distribution due to \AtNextCite{\defcounter{maxnames}{99}}\textcite{Lewandowski2009GeneratingMethod} is a distribution over positive-definite symmetric matrices with unit diagonals, i.e., correlation matrices. In the model specification above, it directly influences the prior for the covariance matrix. Before it was introduced and when HMC was not widely applicable, the usual choices for modeling covariance matrices were Wishart or inverse-Wishart distributions, due to their nice conjugacy properties. However, LKJ is better suited for modern Bayesian computational settings~\autocite{Betancourt2015CommentArguments,lambert2018bayes} and therefore employed in this work.

LKJ has a single parameter $\eta$, which can be interpreted as the shape parameter of a symmetric beta distribution~\autocite{bda3}. As $\eta$ gets larger, the prior is more skeptical of large correlations in the matrix, i.e., providing regularizing effects. The probability density of LKJ with a few $\eta$ values are displayed in \ref{fig:lkj_density}. In this work, $\operatorname{LKJcorr}(\eta=2)$ is chosen to define a weakly informative and regularizing prior.
\csmlongfigure[!htbp]{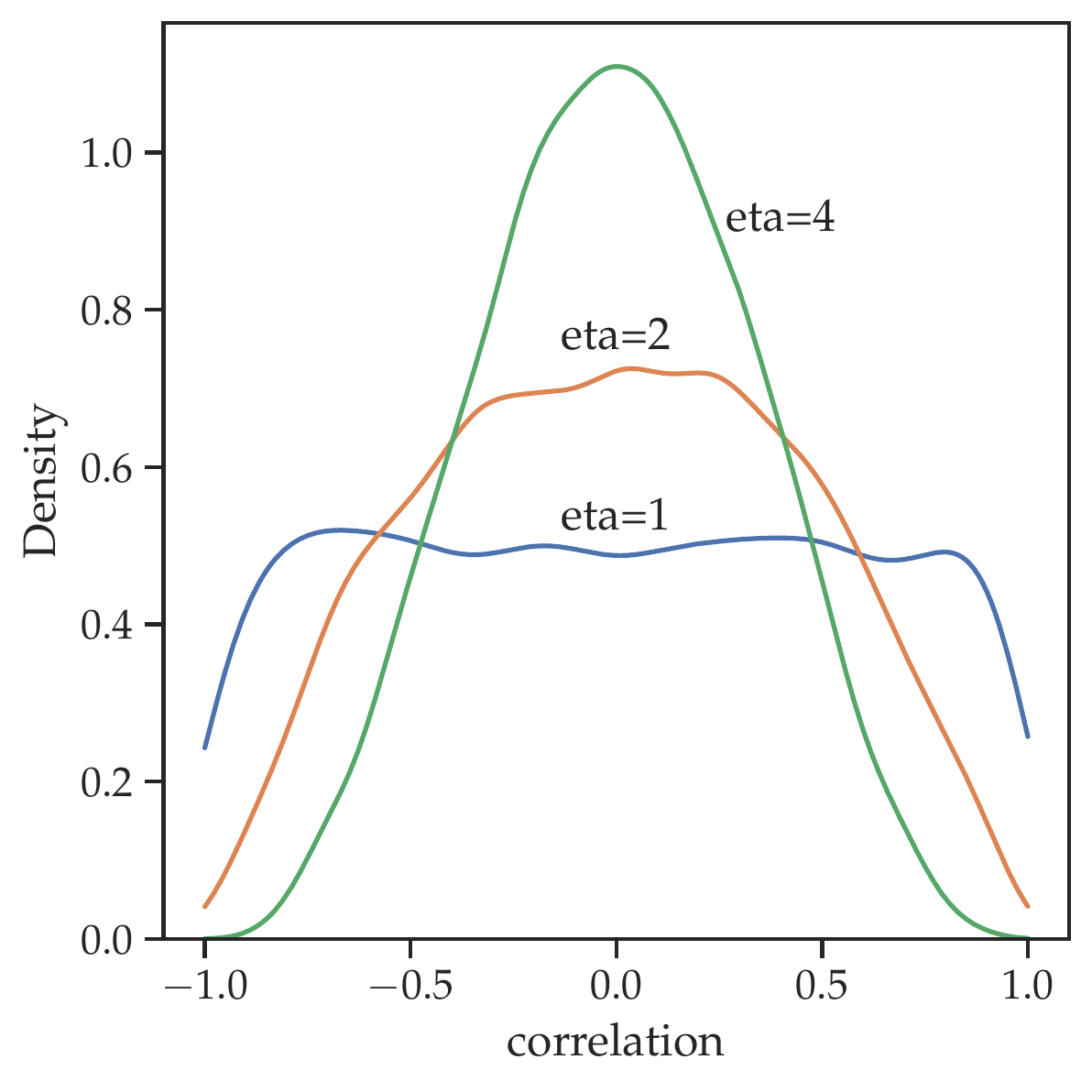}{figures/lkj_density}{0.425\textwidth}{$\operatorname{LKJcorr}(\eta=\mathtt{eta})$ probability density.}{ As $\eta$ increases, larger correlations become less plausible.}

Model~\ref{mod:county_hier_cp}, while being expressive in the data generating process, is a \textit{centered} parameterization of the hierarchical structure~\autocite{Papaspiliopoulos2007AModels}. In this parameterization, the hierarchical parameters (such as $\beta_{\mathsmaller{\textsc{county}}}$) and the lower-level parameters in the prior (e.g., $\mu_{\beta}$ and $\sigma_{\beta}$) are tightly coupled, and they are highly correlated in the posterior. Since this model involves complex geometries and interactions in the posterior, HMC is leveraged for sampling. When there is not a lot of data (which is the case for the current NDIC and VIIRS reportings), this parameterization leads to very inefficient sampling and non-convergences~\autocite{stan_man}. The \textit{noncentered} parameterization is preferable in these cases and therefore employed for building the county level models.

\subsection{Model Reparameterization}
Reparameterization of hierarchical models can be applied to any distribution in the location-scale family, for which the normal distribution is a good candidate. In the case of reparameterizing a multivariate normal prior, suppose the prior for $\bm{\theta}$ is a multivariate normal with mean vector $\bm{\mu}$ and covariance matrix $\bm{\Sigma}$ (such as \cref{exp:mvn_cp}), then a noncentered parameterization is given by:
    \begin{subequations}
        \begin{align}
        \widetilde{\bm{\theta}} &\sim \operatorname{MVNormal}(\mathbf{0}_n, \, \mathbf{I}_n) \\[0.5ex]
        \bm{\varphi} &= \bm{\mu} + \mathbf{L} \cdot \widetilde{\bm{\theta}}
        \end{align}
    \end{subequations}
where $\widetilde{\bm{\theta}}$ has the same dimensions as $\bm{\theta}$ and all of its elements i.i.d.\ according to $\mathcal{N}(0,1)$, $\mathbf{L}$ satisfies $\mathbf{L} \cdot \mathbf{L}^\top=\bm{\Sigma}$, and $\bm{\varphi}$ recovers the exact same prior distribution for $\bm{\theta}$. This reparameterization leads to more efficient sampling by reducing the dependence between $\bm{\mu}$, $\mathbf{L}$, and $\widetilde{\bm{\theta}}$. One choice for $\mathbf{L}$ is the Cholesky factor of $\bm{\Sigma}$, which provides implementation convenience for the multivariate normal cases~\autocite{stan_man} and is adopted in this work.

The noncentered county level model is specified through \crefrange{county_ncp_begin}{county_ncp_end}, with the reparameterized part (corresponding to Model~\ref{mod:county_hier_cp}) highlighted in blue:
\begin{subequations}\label{mod:county_hier_ncp}
    \begin{align}
        \mu_\alpha &\sim \operatorname{Half-Normal}(0.1) \label{county_ncp_begin} \\[0.5ex]
        \mu_\beta &\sim \operatorname{Gamma}(2, 2) \\[0.5ex]
        \sigma_\alpha &\sim \operatorname{Half-Normal}(0.1) \\[0.5ex]
        \sigma_\beta &\sim \operatorname{Half-Normal}(0.1) \\[0.5ex]
        \sigma &\sim \operatorname{Half-Normal}(0.05) \\[0.5ex]
        \color{Blue}\mathbf{L} &\color{Blue}\sim \operatorname{LKJCholeskyCov}\!\left(\eta=2, \begin{bmatrix} \sigma_\alpha & \sigma_\beta \end{bmatrix}^\intercal \right) \label{exp:ncp_core_beg} \\[0.5ex]
        \color{Blue}\begin{bmatrix} z_\alpha \\ z_\beta \end{bmatrix} &\color{Blue}\sim \operatorname{MVNormal} \!\left( \begin{bmatrix} 0 \\ 0 \end{bmatrix}, \begin{pmatrix} 
                    1 & 0 \\
                    0 & 1 
                \end{pmatrix} \right) \\[0.5ex]
        \color{Blue}\begin{bmatrix} \alpha_{\mathsmaller{\textsc{county}}} \\ \beta_{\mathsmaller{\textsc{county}}} \end{bmatrix} &\color{Blue}= \begin{bmatrix} \mu_\alpha \\ \mu_\beta \end{bmatrix} + \mathbf{L} \cdot \begin{bmatrix} z_\alpha \\ z_\beta \end{bmatrix} \label{exp:ncp_core_end} \\[0.5ex]
        \mu_j &= \alpha_{\mathsmaller{\textsc{county}} [j]} + \beta_{\mathsmaller{\textsc{county}} [j]} \times \mathrm{VIIRS}_j \\[0.5ex]
        \mathrm{NDIC}_j &\sim \mathcal{N}(\mu_j, \sigma) \label{county_ncp_end}
    \end{align}
\end{subequations}
where:
\begin{description}[noitemsep,itemindent=-8em,leftmargin=6em]
    \item[\mathbf{L}] is the Cholesky factor of the covariance matrix which has LKJ distributed correlations;
    \item[z_\alpha\text{ and }z_\beta] are the standardized intercept and slope for each county.
\end{description}
The rest of the symbols have the same meaning as in Model~\ref{mod:county_hier_cp}. The noncentered model imposes the exact same probabilistic structure as in Model~\ref{mod:county_hier_cp}, and is implemented for making inference on each county's parameters.

\subsection{Model Fitting}
\begin{figure}[p]
    \centering
    \includegraphics[height=\dimexpr\textheight-39pt\relax]{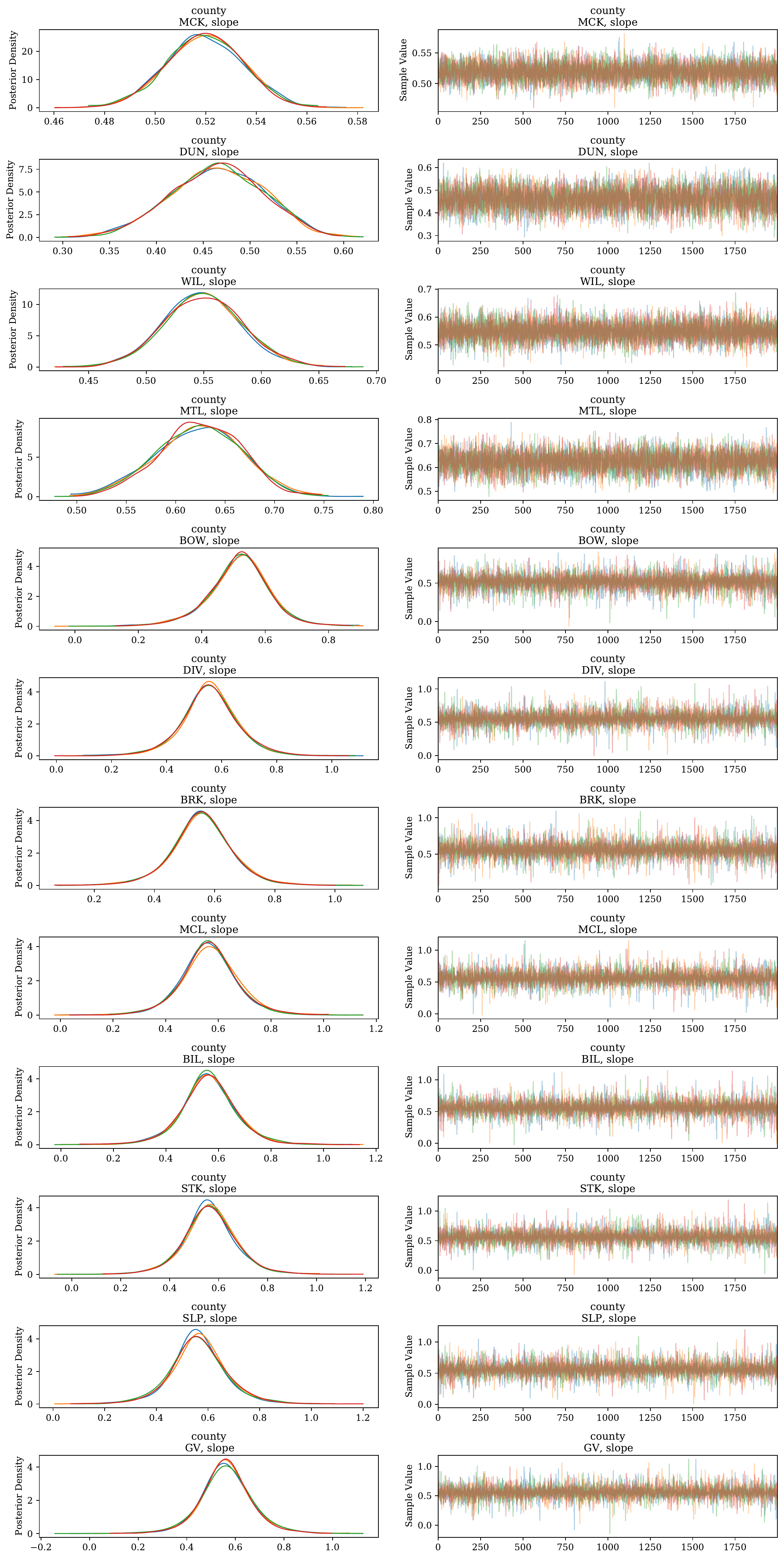}
    \caption{Posterior distributions and trace plots of the slopes for each county.}
    \label{fig:county_slope_traceplt}
\end{figure}

\begin{figure}[p]
    \centering
    \includegraphics[height=\dimexpr\textheight-39pt\relax]{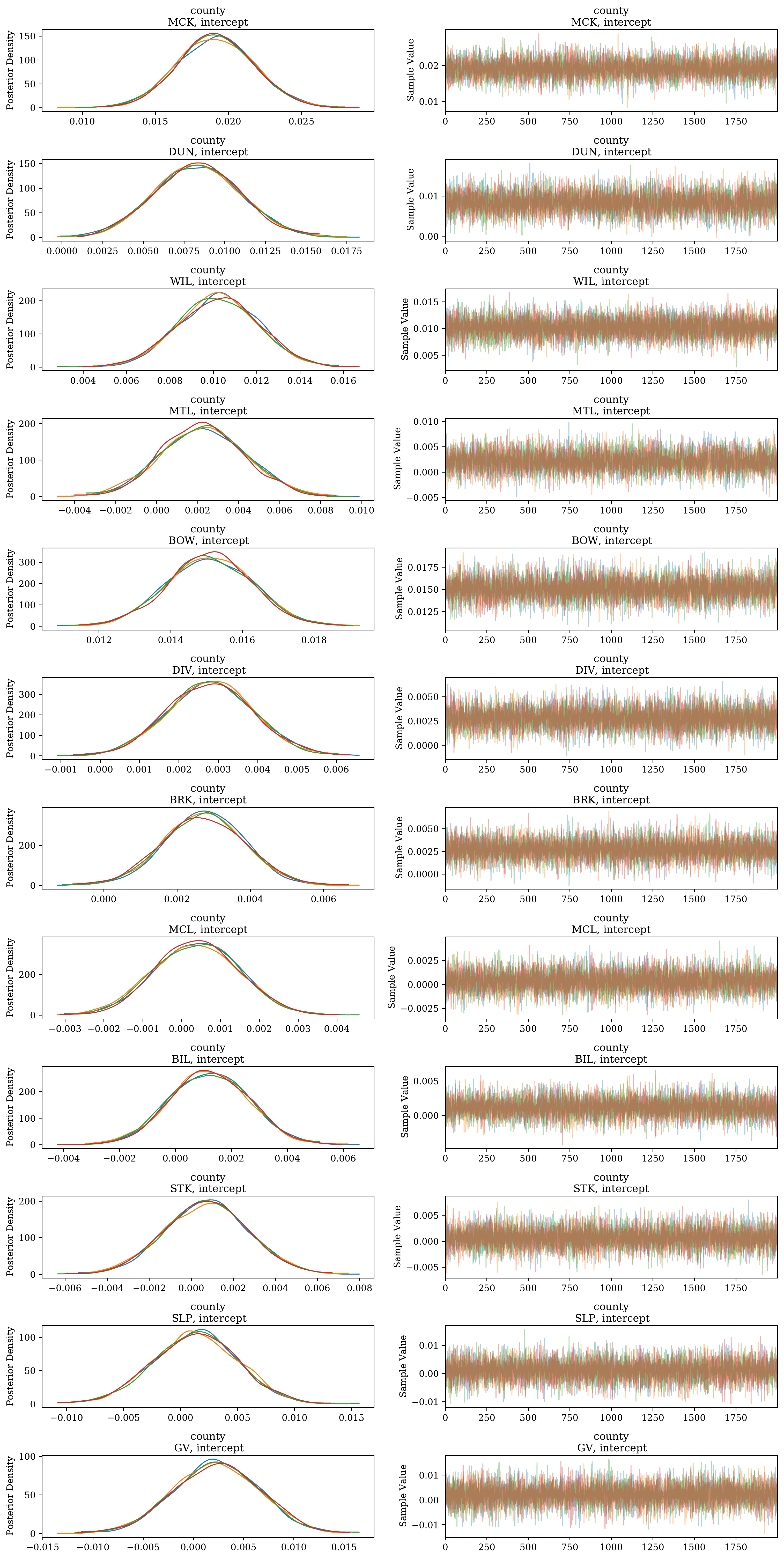}
    \caption{Posterior distributions and trace plots of the intercepts for each county.}
    \label{fig:county_intercept_traceplt}
\end{figure}

Four chains are sampled from the posterior distributions. The posterior distributions and trace plots for the slopes and intercepts are presented in \ref{fig:county_slope_traceplt} and \ref{fig:county_intercept_traceplt}, respectively. Well mixing and convergence have been achieved as shown by the trace plots.

To better compare and contrast the different counties' parameters, the forest plots of \SI{90}{\percent} highest density intervals (HDI) for the slopes and intercepts are given in \ref{fig:county_slope_forestplt} and \ref{fig:county_intercept_forestplt}, respectively. In both figures, counties are ordered by the VIIRS reported volumes, and those with the least amount of estimated volumes (such as SLP and GV) are at the bottom. The thin lines present the \SI{90}{\percent} HDI's and the thicker line segments stand for the interquartile ranges (IQR). The points represent the posterior means.
\csmlongfigure[!htbp]{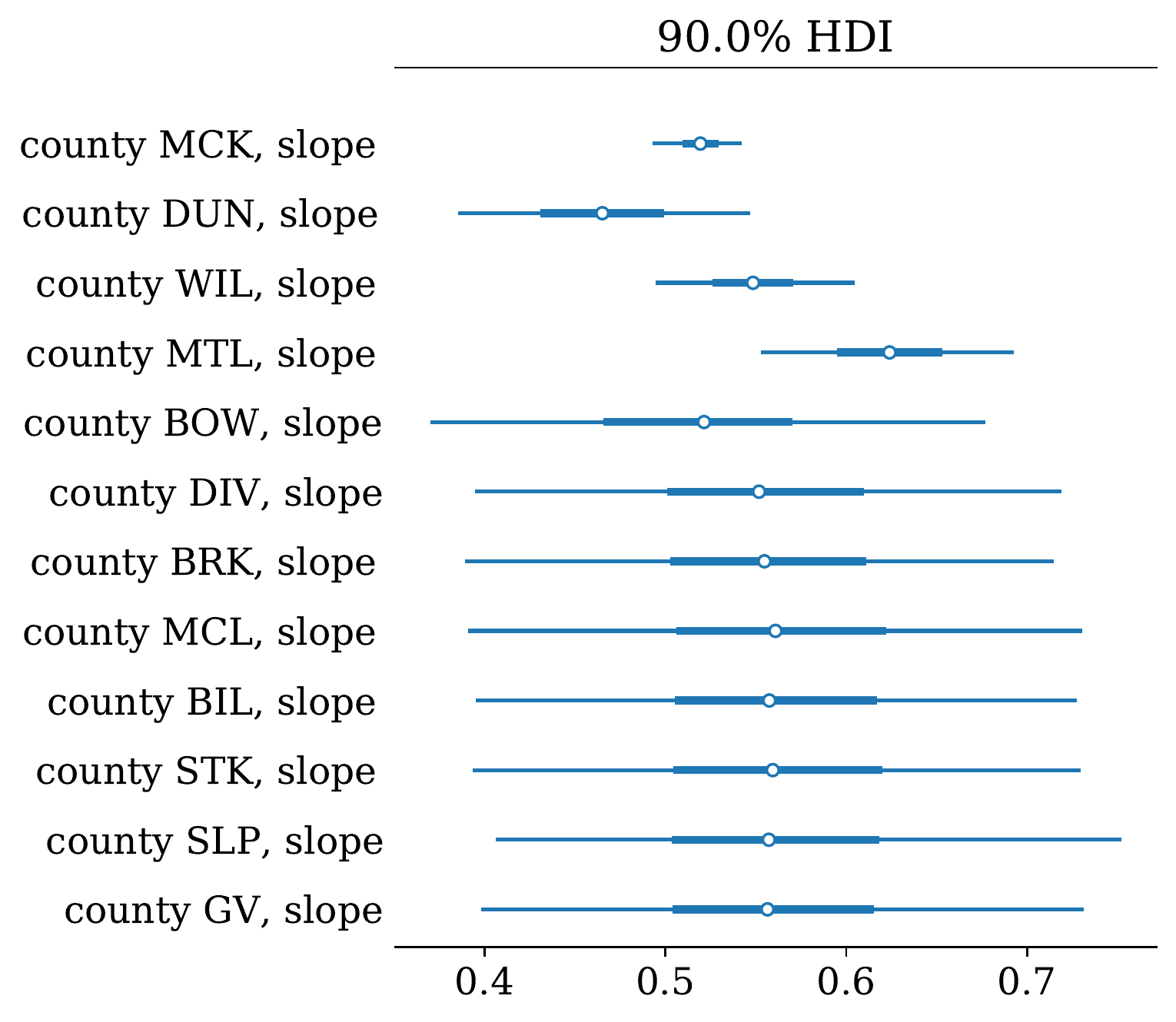}{figures/county_slope_forestplt}{0.5\textwidth}{A forest plot showing the uncertainties around each county's slope estimate.}{ The counties at the bottom have insufficient or noisy datasets, therefore their estimates are largely pulled towards the partially-pooled mean.}
\csmlongfigure[!htbp]{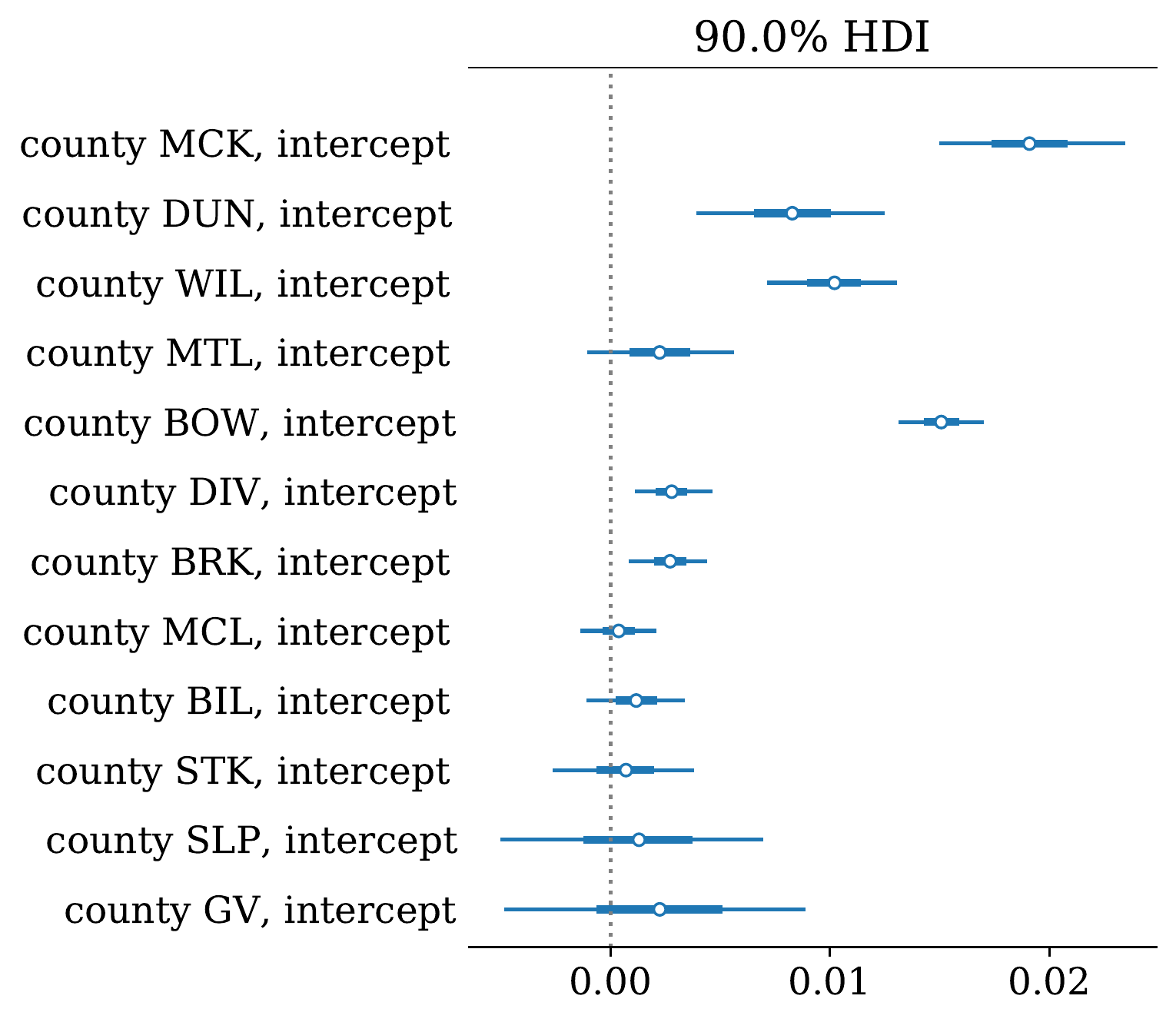}{figures/county_intercept_forestplt}{0.5\textwidth}{A forest plot showing the uncertainties around each county's intercept estimate.}{ The dotted line labels the zero intercept, for which some counties' estimates are not significantly different from.}

In the case of the slopes (\ref{fig:county_slope_forestplt}), it can be seen the top four counties are quite diverse. MTL has the largest point estimate in the entire population ($\hat{\beta}_{\textsc{mtl}} > 0.6$) while DUN has the smallest one ($\hat{\beta}_{\textsc{dun}} < 0.5$). Furthermore, the HDI's for DUN and MTL rarely overlap, indicating that it is almost certain that MTL has a larger slope than DUN. The counties with fewer observations (remaining eight counties) have greater uncertainties in their parameter estimates, while all of their point estimates are pulled towards the partially-pooled mean which is between \num{0.5} and \num{0.6}. When there is not enough data for some counties, the hierarchical model strives to reinforce information sharing among different counties, thus providing more sensible results and also quantifying the uncertainties in such processes. From domain expertise, these results make more physical sense than the no-pooling estimates discussed in Section~\ref{sec:hier_data_descri} (i.e., $\hat{\beta}_{\textsc{slp}} \approx 0$ and $\hat{\beta}_{\textsc{gv}} \gg 0$).

In the case of the intercepts (\ref{fig:county_intercept_forestplt}), there is also heterogeneity among the counties. In particular, by plotting a dotted line labeling the zero intercept, some counties are found to likely have zero intercept (e.g., zero is covered by the IQR or HDI) while others have intercepts that are significantly different from zero. It might not be surprising to get close-to-zero intercepts and greater uncertainties for those counties with less data (such as SLP and GV), however it is interesting to obtain the HDI for MTL that covers zero. Recall that the intercept parameter can be interpreted as the NDIC reported volume which is not captured by VIIRS. This finding for MTL, along with the fact that MTL has the largest slope point estimate (where a larger slope denotes closer proximity to the satellite estimation), convinces the author that MTL used to have persistent and stronger gas flares. They kept VIIRS from missing the flaring events in general, and lead to the reported volumes from NDIC and VIIRS being closer to each other. On the contrary, DUN's smaller slope and larger intercept characterize its flares as sporadic and weaker. One thing worth mentioning is that, with the current interpretation of the intercept, it does not make much physical sense to have negative intercepts. Although every county has positive point estimates for their intercepts, some counties' HDI's show coverage over the negative values. This is a limitation of choosing a 2D Gaussian population model for the intercepts and slopes. Since the 2D Gaussian is supported on $\mathds{R}^2$, in the context of some counties having ``weak data'', negative values make an appearance in their HDI's.

The discussions above naturally lead to the question of whether the slopes and intercepts are correlated. It turns out that, by partially pooling the different types of parameters, a probable negative correlation between the slopes and intercepts is revealed (\ref{fig:rho_prior_vs_post}). The correlation is learned from the heterogeneity in flare characteristics among the counties:
\begin{itemize}
    \item Persistent flares yield smaller intercepts and larger slopes.
    \item Sporadic flares yield larger intercepts and smaller slopes.
\end{itemize}
In other words, intercepts and slopes covary in the entire population of counties. By pooling information across parameter types, what the model learns in the intercept can improve learning about slopes, and vice versa. With this ``experience'' or ``knowledge'', the hierarchical model will be able to quickly update its expectation for any new counties' parameters even with just a few observations in the beginning. It should be noted that there is also some probability mass for the positive correlation values, i.e., the negative correlation is not very strong. This could be due to that some counties do not have a lot of data at this time. The posterior will be updated as more data is brought in.
\csmlongfigure[!htbp]{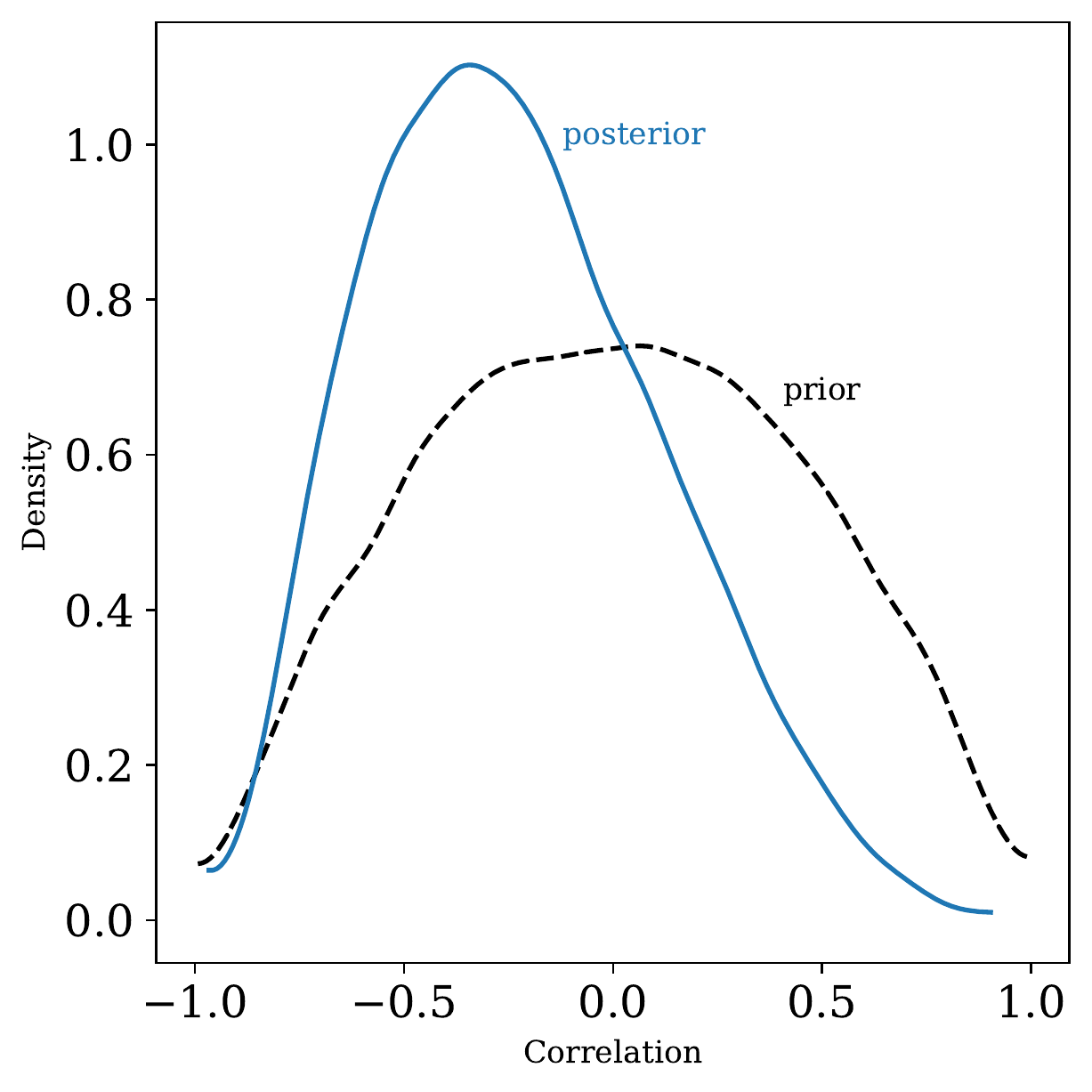}{figures/rho_prior_vs_post}{0.5\textwidth}{Correlation between the intercepts and slopes.}{ Blue: Posterior distribution of the correlation, the mode of which is below zero. Dashed: Prior distribution, the $\operatorname{LKJcorr}(2)$ density.}

Finally, the parameter estimates are reported in \ref{tab:county_param}, from which the parametric model for each county can be recovered, and then deployed in calibration and prediction usage scenarios.
\begin{table}[!htbp]
\centering
\caption{Parameter Estimates of County Level Flaring Model}
\label{tab:county_param}
\begin{tabular}{cclS[table-format=1.3]R}
\toprule
Parameter                  & Variable                    & County & {Point Estimate} & \thead{\SI{90}{\percent} \textnormal{ CI}} \\ \midrule
\multirow{12}{*}{$\alpha_{\mathsmaller{\textsc{county}}}$} & \multirow{12}{*}{Intercept} & MCK    & 0.019            & (0.015, 0.023)                      \\
                           &                             & DUN    & 0.008            & (0.004, 0.013)                      \\
                           &                             & WIL    & 0.010            & (0.007, 0.013)                      \\
                           &                             & MTL    & 0.002            & (-0.001, 0.006)                      \\
                           &                             & BOW    & 0.015            & (0.013, 0.017)                      \\
                           &                             & DIV    & 0.003            & (0.001, 0.005)                      \\
                           &                             & BRK    & 0.003            & (0.001, 0.004)                      \\
                           &                             & MCL    & 0.000            & (-0.001, 0.002)                      \\
                           &                             & BIL    & 0.001            & (-0.001, 0.003)                      \\
                           &                             & STK    & 0.001            & (-0.003, 0.004)                      \\
                           &                             & SLP    & 0.001            & (-0.005, 0.007)                      \\
                           &                             & GV     & 0.002            & (-0.005, 0.009)                      \\ \arrayrulecolor{black!30}\midrule
\multirow{12}{*}{$\beta_{\mathsmaller{\textsc{county}}}$}  & \multirow{12}{*}{Slope}     & MCK    & 0.519            & (0.493, 0.542)                      \\
                           &                             & DUN    & 0.464            & (0.385, 0.547)                      \\
                           &                             & WIL    & 0.549            & (0.495, 0.605)                      \\
                           &                             & MTL    & 0.623            & (0.553, 0.693)                      \\
                           &                             & BOW    & 0.516            & (0.370, 0.677)                      \\
                           &                             & DIV    & 0.554            & (0.395, 0.719)                      \\
                           &                             & BRK    & 0.556            & (0.389, 0.715)                      \\
                           &                             & MCL    & 0.563            & (0.391, 0.730)                      \\
                           &                             & BIL    & 0.560            & (0.395, 0.727)                      \\
                           &                             & STK    & 0.562            & (0.393, 0.729)                      \\
                           &                             & SLP    & 0.561            & (0.406, 0.752)                      \\
                           &                             & GV     & 0.560            & (0.398, 0.731)                      \\ \arrayrulecolor{black}\bottomrule
\end{tabular}
\end{table}

\subsection{Model Extensibility}
Looking back at the hierarchical model and the reparameterization strategy from the previous sections, there are four potential deployment scenarios that are worth discussing. They demonstrate the extensibility and flexibility of the chosen approach in the context of flaring data analytics:
\begin{enumerate}
    \item New counties are present in terms of the reported flaring statistics from both VIIRS and NDIC.
    
    At this time, there are \num{12} counties that have reported flaring statistics from both VIIRS and NDIC. If flaring data becomes available for some other counties in the future, the hierarchical model allows the population to be immediately expanded to accommodate the new counties. This can be seen from the conditional structure in \Cref{eq:de_finetti_joint_mod}: by taking a model for $n+1$ counties
    \begin{equation}
        p(\bm{\theta}_1, \bm{\theta}_2, \dots, \bm{\theta}_n, \bm{\theta}_{n+1},\phi) = \Biggl[ \prod_{i=1}^{n+1} p(\bm{\theta}_i \mid \phi) \Biggr] p(\phi) \, ,
    \end{equation}
    then pulling out the term for the $(n+1)$-th county from the right-hand side (RHS)
    \begin{equation}
        p(\bm{\theta}_1, \bm{\theta}_2, \dots, \bm{\theta}_n, \bm{\theta}_{n+1},\phi) = p(\bm{\theta}_{n+1} \mid \phi) \Biggl[ \prod_{i=1}^{n} p(\bm{\theta}_i \mid \phi) \Biggr] p(\phi) \, ,
    \end{equation}
    it can be recognized that the remaining part on the RHS is the hierarchical model for $n$ counties
    \begin{equation}
        p(\bm{\theta}_1, \bm{\theta}_2, \dots, \bm{\theta}_n, \bm{\theta}_{n+1},\phi) = p(\bm{\theta}_{n+1} \mid \phi) \, p(\bm{\theta}_1, \bm{\theta}_2, \dots, \bm{\theta}_n, \phi) \, .
    \end{equation}
    This indicates the newly introduced counties will only depend on the population parameters $\phi$, i.e., how the new counties interact with the existing ones (from the initial dataset) is not explicitly specified but being mediated through $\phi$. This mechanism allows the population (of counties) to be expanded arbitrarily. In practice, without any modification, Model~\ref{mod:county_hier_ncp} can be re-fitted with the new dataset as a whole.
    \item More data are available for those counties which used to have very few observations.
    
    In the event of more data becoming available for those counties with wide HDI's such as SLP and GV, the posteriors will be updated according to that information. Their HDI's would become narrower and narrower as more and more data are available, and since the hierarchical model pools information among the counties, these counties will contribute to updating the population model's and other counties' parameters. Similar to Item \num{1} above, Model~\ref{mod:county_hier_ncp} does not need modifying and can be re-fitted with the new data.
    \item Sample sizes among counties become more unbalanced.
    
    In general, when there is a lot of data for each county, the centered parameterization (Model~\ref{mod:county_hier_cp}) is more efficient. When the sample size is not large, which is the case for the current VIIRS and NDIC reportings, the noncentered parameterization (Model~\ref{mod:county_hier_ncp}) is better. However, the parameterization for hierarchical models is not a monolithic tactic. If the reported flaring data becomes very unbalanced across counties, e.g., some counties have a huge amount of data whereas others have very little data, then each county can be parameterized differently. More specifically,
    \begin{itemize}
        \item For the counties that have strong data such that their likelihood functions dominate, centered parameterization can be applied through \crefrange{exp:cp_core_beg}{exp:mvn_cp}.
        \item For the counties that have weak data such that their prior models dominate, noncentered parameterization can be applied through \crefrange{exp:ncp_core_beg}{exp:ncp_core_end}.
    \end{itemize}
    All in all, this is still one hierarchical model which defines the exact same probabilistic structure as Model~\ref{mod:county_hier_cp} or Model~\ref{mod:county_hier_ncp}, but avoids inefficiencies and non-convergences in the sampling from posteriors.
    \item Oilfield level heterogeneity needs to be examined.
    
    Under the assumptions that the oilfields in North Dakota are exchangeable and the population of oilfields (which conduct flaring) can grow, the hierarchical model developed in this chapter can be directly applied to investigate the heterogeneity in different oilfields' parameters. Following the reverse geocoding as discussed in Section~\ref{sec:rev_geocode}, there are \num{258} oilfields that have both NDIC and VIIRS reportings for the same study period as in this chapter. Some oilfields have very few observations and can benefit from the hierarchical model through pooling information among the entire population of oilfields. Furthermore, due to the number of oilfields being relatively large, the population model could be learned with more ease (because more information is available for the population). In the case of the county level model developed in this chapter, since there are only \num{12} individuals (counties) in the population, some uncertainties about the population are inevitably present and reflected through the posteriors.
\end{enumerate}

The models developed in this chapter, while capturing the heterogeneity among the different counties in North Dakota, rely on the assumption that all the monthly observations within a certain county are conditionally i.i.d. For situations where the temporal structure has to be taken into consideration, other types of models can be built and are discussed in the next chapter.

\chapter{Flaring Time Series Analytics}\label{ch:gp}

\setlength{\epigraphwidth}{0.65\textwidth}
\setlength{\epigraphrule}{0pt}
\epigraph{``Were neural networks over-hyped, or have we underestimated the power of smoothing methods? 

I think both these propositions are true.''}{--- \textcite{mackay_IT}}

\subsection{Learning the Flaring Pattern and Behavior}
In this chapter, the author develops a generic framework for revealing flaring patterns and behaviors. The main challenges are fourfold:
    \begin{enumerate}
        \item Observed data are noisy.
            \begin{itemize}
                \item Companies estimate the flaring volumes and conduct self-reporting. Satellites could miss some events. However, having knowledge about the underlying process is vital in lots of situations including when the state and local governments need to make key decisions based on the data. In the meantime, understanding the underlying process helps with anomaly detection by differentiating between true anomalies in reporting and ordinary noise or stochasticity.
            \end{itemize}
        \item A probabilistic approach is desirable to be adopted.
            \begin{itemize}
                \item A set of most probable functions (characterizing the underlying process) are preferable over one single best fit function.
            \end{itemize}
        \item The observations of a certain entity are time series.
            \begin{itemize}
                \item The temporal structure is intrinsic to the dataset and thus must be harnessed.
            \end{itemize}
        \item The framework should be generic enough for automated insights extraction.
            \begin{itemize}
                \item There are more than \num{200} operators and \num{500} oilfields operating in North Dakota. Choosing a specific parametric form of model (e.g., ARIMA or LSTM) for each entity and then fitting the model to the data is not only time consuming, but also prevents easy integration into automation pipelines (for extracting insights for example).
            \end{itemize}
    \end{enumerate}

It is striking that the elegant properties of Gaussian process make it a natural choice to tackle all of these challenges and is therefore employed in this chapter.

\subsection{Gaussian Process}
A Gaussian process (GP) can be viewed as a distribution over infinite-dimensional Hilbert space of functions. It is formally defined as ``a collection of random variables, any finite number of which have a joint Gaussian distribution''~\autocite{gpml}. Gaussian processes are extremely powerful nonparametric learning techniques, which provide a composite of flexibility and interpretability. They are well suited to problems which necessitate principled handling of uncertainty and interpretation, in the presence of noisy and dynamic datasets. Such scenarios include smoothing~\autocite{Deisenroth2012RobustProcesses} and time series modeling~\autocite{Roberts2013GaussianModellingb}. They are also well established in different fields under various names, for example kriging in geostatistics and Kalman filters both correspond to Gaussian processes~\autocite{mackay1998introductionGP}.

In this work, the motivation is to develop a generic framework for recognizing the underlying unknown processes $f(\mathbf{x})$ which reflect flaring strategies and behaviors. Thus inference is conducted directly in the function space employing GP as a prior. A Gaussian process is completely specified by its mean function $m(\mathbf{x})$ and covariance function $k(\mathbf{x}, \mathbf{x}')$~\autocite{Bandyopadhyay2018CSMStatistics}, which are defined as:
\begin{align}
    m(\mathbf{x}) &= \mathds{E} [f(\mathbf{x})] \,, \\[0.5ex]
    k(\mathbf{x}, \mathbf{x}') &= \mathds{E} [ (f(\mathbf{x}) - m(\mathbf{x})) (f(\mathbf{x}') - m(\mathbf{x}'))] \,,
\end{align}
and the function distributed as a Gaussian process is denoted by
\begin{equation}
    f(\mathbf{x}) \sim \mathcal{GP}\!\left(m(\mathbf{x}), \, k(\mathbf{x}, \mathbf{x}') \right).
\end{equation}

    \subsubsection{Mean Function}
    In this work, the mean functions are always chosen to be zero, since there is no prior knowledge on the mean of the latent processes. In the meantime, for GPs with a zero mean function, the mean of the posterior process is not confined to be zero~\autocite{gpml}. All the latent functions modeled with a GP prior in this dissertation follow
    \begin{equation}
        f(\mathbf{x}) \sim \mathcal{GP}\!\left(0, \, k(\mathbf{x}, \mathbf{x}') \right),
    \end{equation}
    where $k$ is some covariance function.
    
    \subsubsection{Covariance Function}
    Covariance function, also known as kernel, is the crucial ingredient in a GP, as it encodes one's assumptions about how the function should behave by defining similarity. The fundamental assumption is that data points with inputs $\mathbf{x}$ which are close would have similar target values $y$. This assumption is usually very reasonable in areas including time series modeling, and it is theoretically backed by Tobler's first law of geography. The covariance functions used in this dissertation include:
        \begin{enumerate}
            \item The Mat\'ern class of covariance functions, which is given by:
            \begin{equation}
            k_{\nu }(r)={\frac {2^{1-\nu }}{\Gamma (\nu )}}{\Bigg (}{\sqrt {2\nu }}{\frac {r}{\ell }}{\Bigg )}^{\!\nu} K_{\nu }{\Bigg (}{\sqrt {2\nu }}{\frac {r}{\ell }}{\Bigg )} \, ,
            \end{equation}
            where $\Gamma(\cdot)$ is the gamma function, $K_{\nu }$ is a modified Bessel function of the second kind of order $\nu$, $r=\norm{\mathbf{x}-\mathbf{x}'}$, and $\ell$ is the lengthscale controlling the smoothness from one perspective: large $\ell$ characterizes functions which change slowly and can be reliably extrapolated further away.
            
            The Mat\'ern covariance functions can be written as a product of an exponential and a polynomial of order $p$, when $\nu$ is half-integer: $\nu=p+1/2, \, p \in \mathds{N}_{0}$. The hyperparameter $\nu$ controls the smoothness from another perspective: when $\nu=1/2$, the Mat\'ern kernel becomes the exponential kernel (continuous but not differentiable); as $\nu \to \infty$, it becomes the exponentiated quadratic kernel (infinitely differentiable). \textcite{gpml} argued that the most interesting cases for machine learning would be $\nu=3/2$ and $\nu=5/2$.
            
            For gas flaring time series, as operators might change flaring strategy at any given time due to policy changes, gas processing facility deployment, gas price fluctuation, etc., the latent process might not be as smooth as infinitely differentiable. Instead the Mat\'ern kernel is harnessed which is capable of inducing non-smooth function realizations to handle those discontinuities. Specifically the Mat\'ern kernel with $\nu=5/2$ is chosen for this dissertation with the input space $\mathcal{X} \subseteq \mathds{R}^1$:
            \begin{equation}
            k_{\textsc{mat\'ern52}}(x, x'; \ell) \coloneqq \left(1 + \frac{\sqrt{5(x - x')^2}}{\ell} +
            \frac{5(x-x')^2}{3\ell^2}\right)
            \exp\!\left[ - \frac{\sqrt{5(x - x')^2}}{\ell} \right],
            \end{equation}
            where $x$ vary over the time domain.
            
            \item The standard periodic kernel due to \textcite{mackay1998introductionGP}:
            \begin{equation}
            k_{\textsc{periodic}}(x, x'; T, \ell) \coloneqq \exp\!\left( -\frac{\sin^2(\uppi |x-x'| \frac{1}{T})}{2\ell^2} \right),
            \end{equation}
            where $T$ denotes the period. This kernel is used for modeling seasonal behaviors.
            
            \item The white noise kernel, which is given by:
            \begin{equation}
            k_{\mathsmaller{\textsc{WhiteNoise}}}(x, x'; \delta) \coloneqq \delta^2 \mathbf{I}_n,
            \end{equation}
            where $\delta^2$ is the variance of the noise. In this dissertation, the usage of the white noise kernel is for stabilizing the computation of the covariance matrix. Adding a small value of diagonal shift will try to guarantee the resulting covariance matrix is always positive semi-definite.
        \end{enumerate}
        
    A nice property is that the sum and product of the established kernels are still valid kernels. This fact is also exploited in the model building process in this work.
    
    \subsubsection{Inference and Model Reparameterization}
    In practice, one always works with a dataset of finite size. In such situations, a multivariate normal prior distribution is placed on the vector of function values $\mathbf{f}$,
    \begin{equation}
    \mathbf{f} \sim \operatorname{MVNormal}(\mathbf{m}_{x},\, \mathbf{K}_{xx}) \, , \label{eq:gp_latent_cp}
    \end{equation}
    where the vector $\mathbf{m}_x$ and the matrix $\mathbf{K}_{xx}$ are the mean function and covariance function evaluated over the inputs $\mathbf{x}$.
    
    A key question which has significant impact on the inference is how to learn the hyperparameters from data. A natural (and popular) approach is to conduct maximum likelihood estimation, i.e., generating point estimates leveraging the data. However, as \textcite{Betancourt2017RobustStan} showed with experiment results, both regularized and unregularized maximum marginal likelihood have limited performance in terms of fitting robustly and recovering the true data generating process. Technically, given a particular kernel with particular hyperparameters, a GP does not support an entire Hilbert space but only a slice through that space; changing the hyperparameters by an infinitesimal amount yields a different slice which has no overlap with the original one. Therefore in this dissertation, a full Bayesian approach is taken for the GP inference, i.e., the entire Hilbert space of functions is considered by taking into account all of the possible hyperparameters for a specific kernel. 
    
    For the class of problems which have Gaussian observation models, GP has nice closed-form posterior results. However, for the situations which do not have Gaussian observation models, for examples the ones in this dissertation which employ $\stut$ or Poisson likelihood, there does not exist analytical solutions. HMC as discussed in Section~\ref{sec:mcmc} is used to sample from the posteriors.
    
    Specifically, the noncentered parameterization of the latent multivariate Gaussian is exploited. The reparameterized model is
    \begin{subequations}
        \begin{align}
        \widetilde{\mathbf{f}} &\sim \operatorname{MVNormal}(\mathbf{0}_n, \, \mathbf{I}_n) \label{eq:gp_repar_begin} \\[0.5ex]
        \mathbf{L} &= \operatorname{Cholesky}(\mathbf{K}_{xx}) \\[0.5ex]
        \mathbf{f} &= \mathbf{m}_{x} + \mathbf{L} \cdot \widetilde{\mathbf{f}} \label{eq:gp_repar_end}
        \end{align}
    \end{subequations}
    which defines the same distribution as \cref{eq:gp_latent_cp} but induces a nicer posterior geometry for HMC to explore and sample from~\autocite{Betancourt2017RobustStan}.
    
    Once the learning on hyperparameters is done, posterior predictive distribution of the latent function values which are not part of the original dataset is obtained by
    \begin{equation}\label{eq:gp-cond}
    \mathbf{f}_{*} \mid \mathbf{f} \sim \operatorname{MVNormal} \!\left(
      \mathbf{m}_{*} + \mathbf{K}_{x*}^{\top} \mathbf{K}_{xx}^{-1} (\mathbf{f}-\mathbf{m}_{x}), \,
      \mathbf{K}_{**} - \mathbf{K}_{x*}^{\top} \mathbf{K}_{xx}^{-1}\mathbf{K}_{x*} \right),
    \end{equation}
    where $\mathbf{m}_{*}$ is the mean function evaluated at the new inputs, $\mathbf{K}_{**}$ is the covariance between the new inputs, and $\mathbf{K}_{x*}$ is the covariance between the original inputs and the new inputs.

\subsection{Suite of Models for Pattern Recognition}
This section presents models built from various angles, with the goal of providing a coherent framework for learning the flaring pattern and behavior in a principled manner. Each model is tested on real flaring data from North Dakota. Whenever more granular analytics capabilities are demonstrated through investigations at oilfield level or operator level, the data from a major producing field, the Blue Buttes Oilfield~\autocite{Alexeyev2017WellFormation}, and one operator, denoted by `Operator A' are used.

\subsubsection{Modeling Proportion of Gas Flared}\label{sec:gp_gas_p}
The proportion of gas production that is flared is an indicator of flaring intensity and energy efficiency. It is interesting to investigate whether the proportion has changed over a period of time for certain operators and oilfields. The model is specified through \crefrange{gas_p_begin}{gas_p_end}:

\begin{subequations}\label{eq:gp_gas_p}
    \begin{align}
        \ell &\sim \operatorname{Gamma}(2, 1) \label{gas_p_begin} \\[0.5ex]
        \eta &\sim \operatorname{Half-Cauchy}(5) \\[0.5ex]
        \nu &\sim \operatorname{Gamma}(2, 0.1) \\[0.5ex]
        \hat{\sigma}^2 &\sim \operatorname{Half-Cauchy}(5) \\[0.5ex]
        k &= \eta^2 \times k_{\textsc{mat\'ern52}}(x, x'; \ell) \\[0.5ex]
        f &\sim \mathcal{GP}(0, k) \\[0.5ex]
        \pi_i &= \operatorname{logit}^{-1}\!\left(f(x_i) \right) \\[0.5ex]
        \mu_i &= \pi_i \times G_i \\[0.5ex]
        F_i &\sim \stut(\nu, \mu_i, 1/\hat{\sigma}^2) \label{gas_p_end}
    \end{align}
\end{subequations}
where:
\begin{description}[noitemsep,itemindent=-8em,leftmargin=6em]
    \item[\ell] is the lengthscale for the Mat\'ern kernel;
    \item[\eta] is the marginal deviation parameter controlling how strongly the latent functions vary in the output space;
    \item[\nu] is the degrees of freedom for the $\stut$ likelihood;
    \item[\hat{\sigma}^2] controls the inverse scaling parameter of the $\stut$ likelihood (analogous to the precision of a Gaussian distribution);
    \item[k] is the covariance function for the GP;
    \item[f] denotes the latent process, which is distributed according to the GP;
    \item[\pi_i] is the underlying flaring gas proportion of month $i$. Since proportion is bounded between \num{0} and \num{1}, the inverse-logit function is applied to the latent process;
    \item[G_i] is the total gas production of month $i$;
    \item[\mu_i] denotes the underlying flared volume of month $i$;
    \item[F_i] is the reported flared volume, which is modeled using a $\stut$ observation model.
\end{description}

The reasoning behind choosing a $\stut$ observation model is to make the model specification be able to generalize to as many entities as possible and be robust to (potentially many) outliers and noisy data points. This is due to the fact that at this time, operators have to estimate the flared volume by their own procedures and conduct reporting, in which case inaccuracies are introduced unintentionally or intentionally. The heavier tail of Student's $t$-distribution is a natural decision in modeling to deal with those phenomena. This line of thought, i.e., design models that are generic and robust, is indeed reflected in choosing the half-Cauchy priors (which are heavy-tailed and very weakly informative) and GP as a nonparametric regression technique.

To demonstrate this model's capability on real data, both the Blue Buttes Oilfield and Operator A are tested. The production and flared volumes coming from NDIC are used. For the oilfield, the posterior distributions and trace plots of the hyperparameters are presented in \ref{fig:field_gas_proportion_trace}. The posterior predictive samples for the underlying process of gas flaring proportions ($\pi_i$) are demonstrated in \ref{fig:field_gas_proportion_ppc}, which depict the trend very clearly. The colored bands have the below coverage for the posterior samples:
    \begin{itemize}
        \item The darkest colored band (in the center at a certain $x$ location) represents the \nth{49} percentile to \nth{51} percentile;
        \item The lightest colored band (characterized by the widest interval at a certain $x$ location) represents the \nth{1} percentile to \nth{99} percentile.
    \end{itemize}

Additionally, \num{30} random samples are drawn from the GP posterior and plotted on the same figure, showing as thin lines. The latent functions do not go through all the observed data points, in which case the model would have been overfitted; instead they present the possible functions which are most compatible with the data as well as the assumptions inherent in the model. On one hand, the insights are already obtained, i.e., the underlying process is inferred. On the other hand, this serves as an anomaly detection tool. For example, the state government might be interested to look into that observed data in the second half of 2019 which deviated quite a lot from the ``true'' process, e.g. to audit the reporting for that month or to investigate what had happened that led to a sudden huge drop in flaring in just one month.
\csmlongfigure[!htbp]{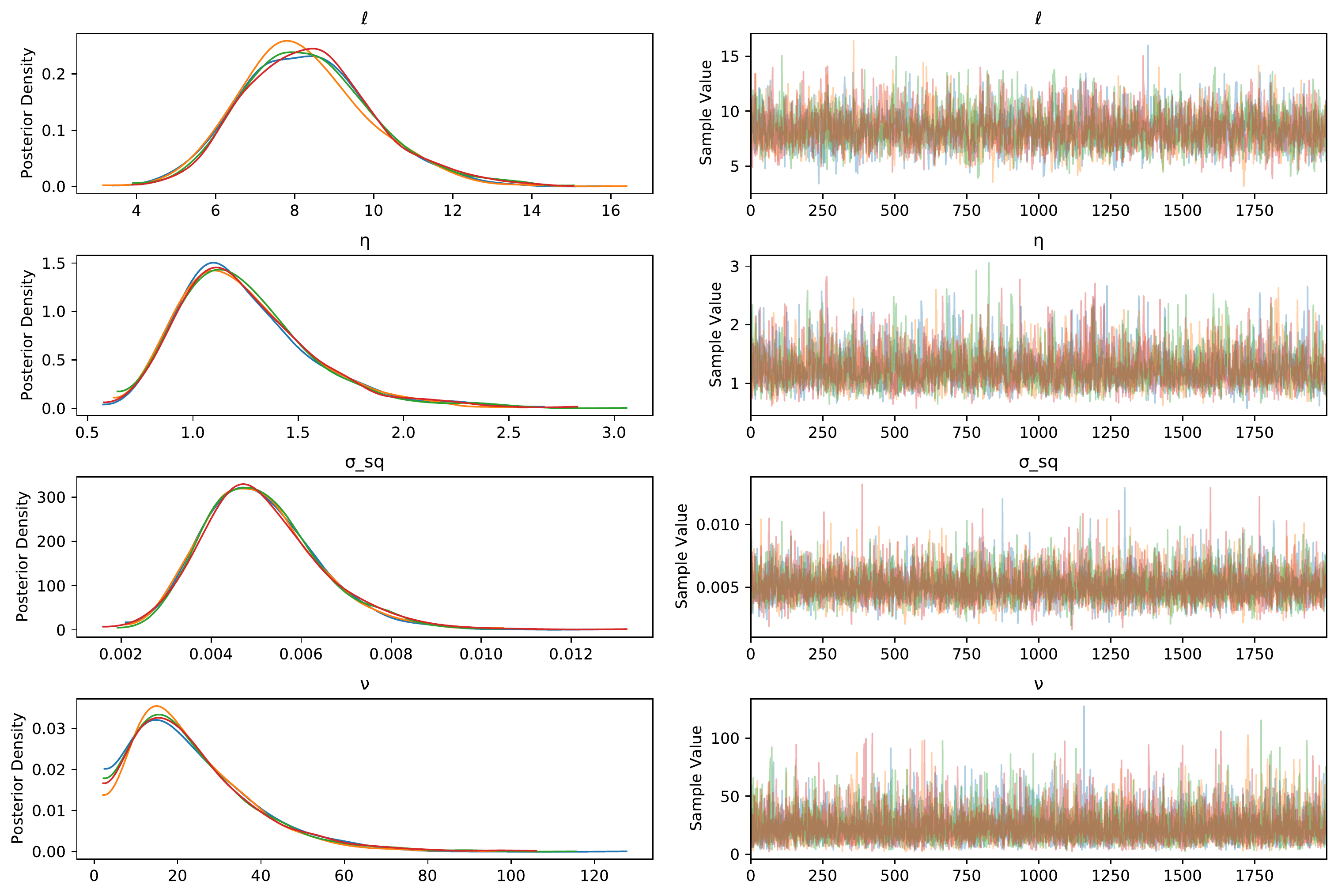}{figures/field_gas_proportion_trace}{\textwidth}{Posterior distributions and trace plots for the Blue Buttes Oilfield gas flaring proportion model.}{ Well mixing and convergence have been achieved.}
\csmlongfigure[!htbp]{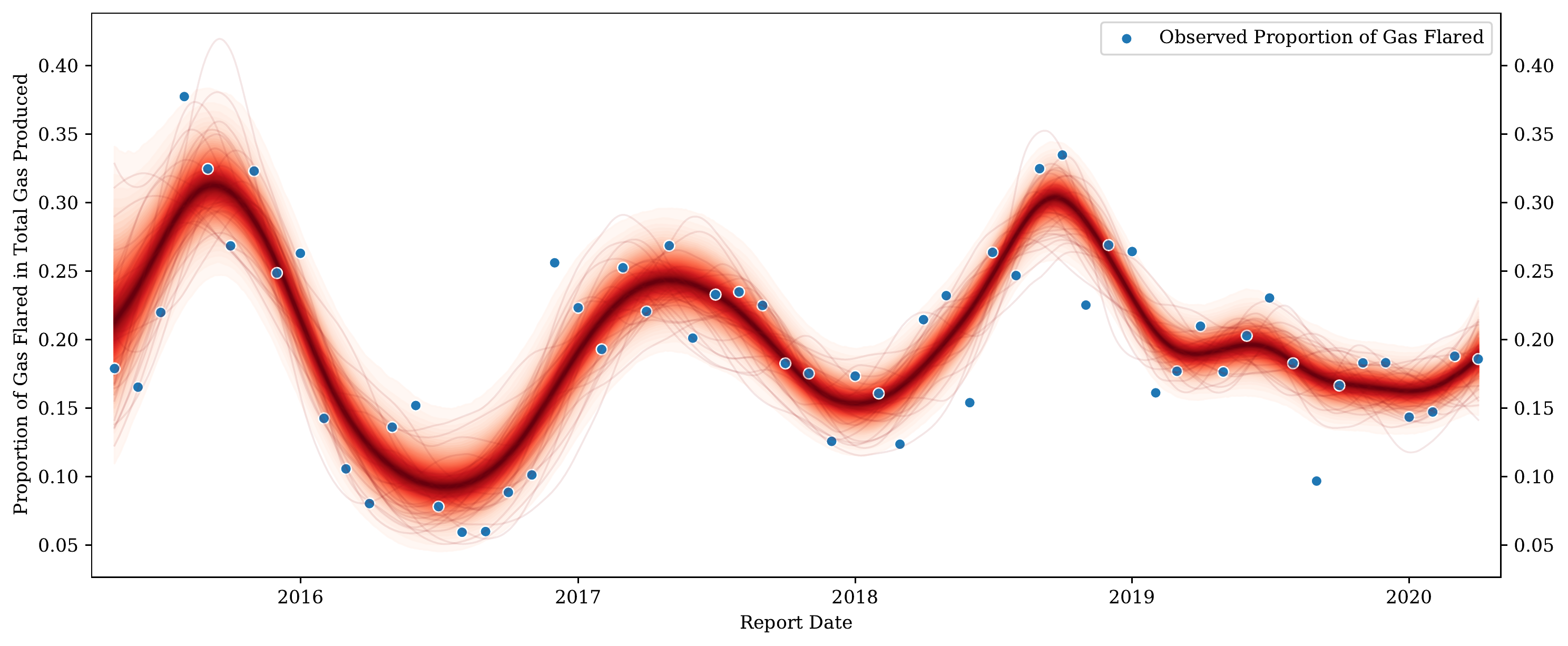}{figures/field_gas_proportion_ppc}{\textwidth}{Posterior predictive samples showing the gas flaring proportion variations at the Blue Buttes Oilfield.}{ Blue points are the observed data while red lines present the posterior samples.}

With the exact same model specification, the model is also run with the operator's data. The posterior distributions and trace plots of the hyperparameters are presented in \ref{fig:operator_gas_proportion_trace}. The posterior predictive samples for the underlying process of gas flaring proportions ($\pi_i$) are demonstrated in \ref{fig:operator_gas_proportion_ppc}. It can be seen this operator's flaring proportion time series is more jagged than the Blue Buttes Oilfield (which is operated by more than five companies). A operator can change flaring strategies more swiftly which can be captured as well. Nevertheless the long-term trend is also available. Comparing \ref{fig:field_gas_proportion_trace} and \ref{fig:operator_gas_proportion_trace}, it can be seen the posterior distributions are very different. However the priors for them were specified in the exact same way. This showcases the power of Bayesian approach. Taking $\ell$ as an example, a $\operatorname{Gamma}(2,1)$ prior is placed on it. However, after conditioning on the data, the operator model reports smaller lengthscale values on average (indicating jagged processes), whereas the oilfield model reports larger lengthscale values (suggesting smoother processes).
\csmlongfigure[!htbp]{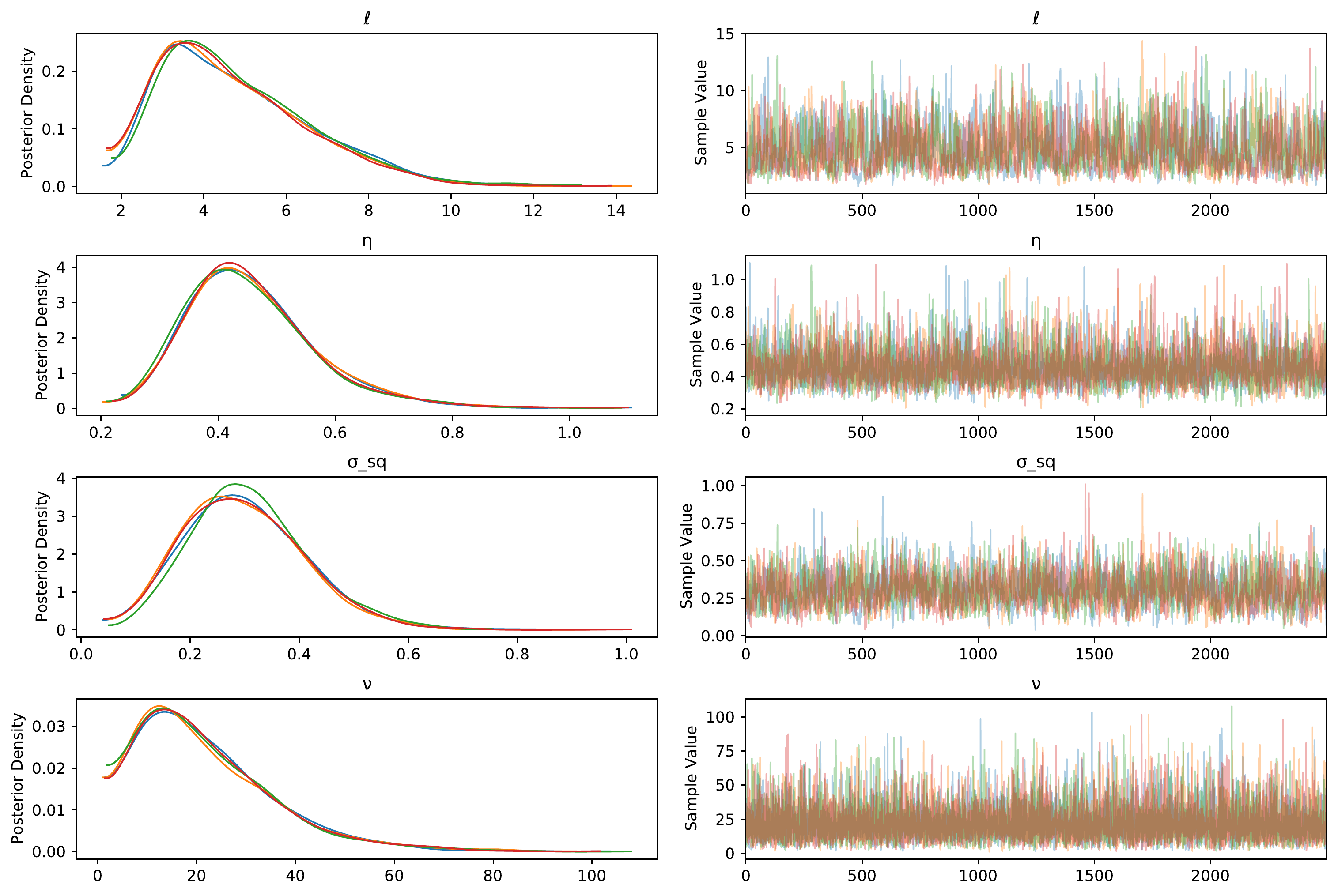}{figures/operator_gas_proportion_trace}{\textwidth}{Posterior distributions and trace plots for the Operator A gas flaring proportion model.}{ Well mixing and convergence have been achieved. Notice the differences between these inference results and those in \ref{fig:field_gas_proportion_trace}, both of which are based on exactly the same priors and likelihood, demonstrating the model specification's wide applicability.}
\csmlongfigure[!htbp]{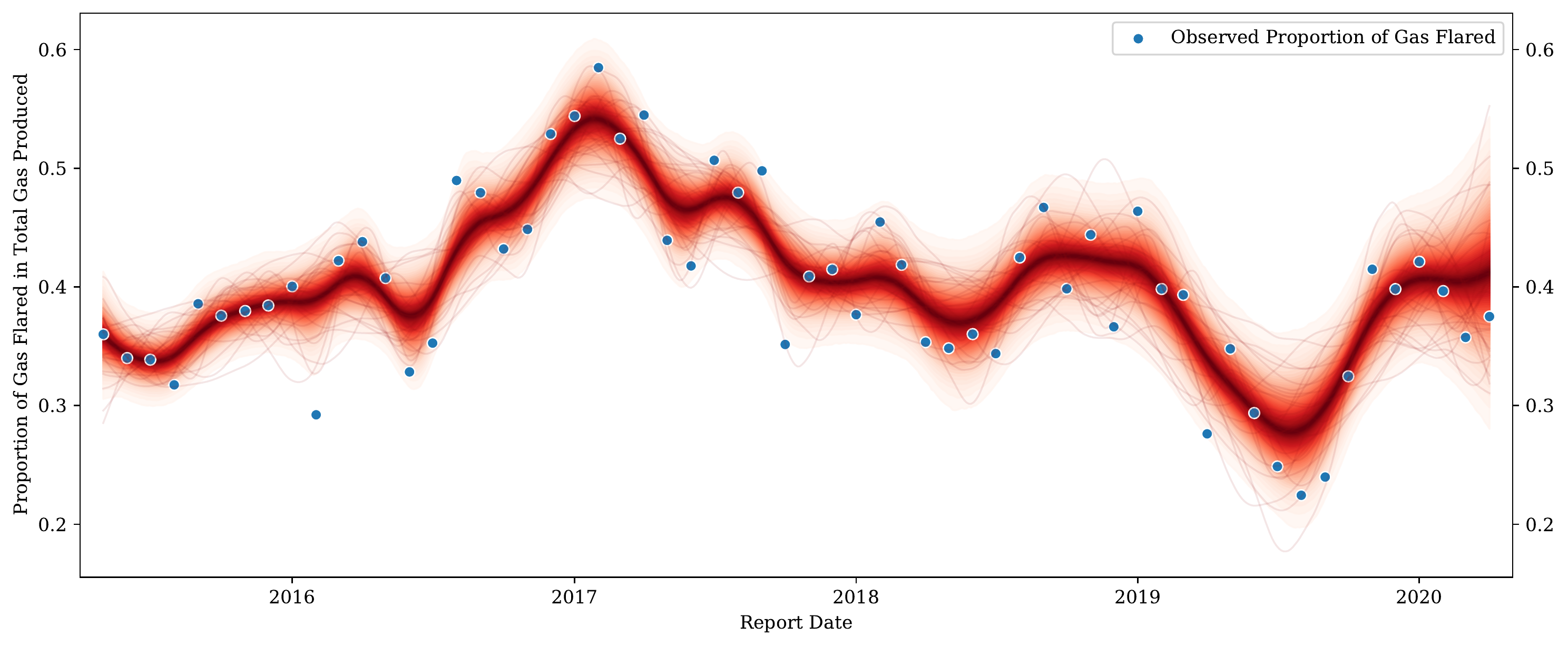}{figures/operator_gas_proportion_ppc}{\textwidth}{Posterior predictive samples showing the gas flaring proportion variations of Operator~A\@.}{ Blue points are the observed data while red lines present the posterior samples.}

Order 24665, which is established by the \citeauthor{NorthDakotaIndustrialCommission2014North041718}, defines the gas capture percentage $p_{\textsc{cap}}$ as
\begin{equation}
    p_{\textsc{cap}} = \frac{G_{\textsc{sold}} + G_{\textsc{used}} + G_{\textsc{proc}}}{G_{\textsc{prod}}} \, ,
\end{equation}
where:
\begin{description}[noitemsep,itemindent=-8em,leftmargin=6em]
    \item[G_{\textsc{sold}}] is the monthly gas sold;
    \item[G_{\textsc{used}}] is the monthly gas used on lease;
    \item[G_{\textsc{proc}}] is the monthly gas processed;
    \item[G_{\textsc{prod}}] is the monthly gas produced.
\end{description}

Since North Dakota bans the venting of natural gas~\autocite{U.S.DepartmentofEnergy2019NorthSheet}, it is obvious the model developed in this section provides a powerful tool for NDIC to evaluate compliance with the gas capture goals: at a given month $i$, $p_{\textsc{cap}}=1-\pi_i$. Furthermore, when looking at the model specification, there is nothing special that encodes the data sources and location information. A user of this model is free to use satellite estimation as the observed data or apply it to the Permian Basin, and conduct inference on the flaring proportion. This is a benefit from using nonparametric and interpretable models as opposed to black box models (such as the neural networks, in which case the learned weights and bias inside the network provide little or no domain insights). The author hopes this section provides a comprehensive view in terms of how and why to use GP, with real data. Models built and presented in later sections follow a similar flow.

\subsubsection{Modeling Proportion of Wells Flaring}
The proportion of wells that conduct flaring in a month can reflect a company's flaring strategy and is an indicator of flaring magnitude. It is interesting to investigate how this indicator varies for a certain entity in a certain time period. The model is specified through \crefrange{well_p_begin}{well_p_end}:
\begin{subequations}\label{mod:well_proportion_flare}
    \begin{align}
        \ell &\sim \operatorname{Gamma}(2, 1) \label{well_p_begin} \\[0.5ex]
        \eta &\sim \operatorname{Half-Cauchy}(5) \\[0.5ex]
        k &= \eta^2 \times k_{\textsc{mat\'ern52}}(x, x'; \ell) \\[0.5ex]
        f &\sim \mathcal{GP}(0, k) \\[0.5ex]
        p_i &= \operatorname{logit}^{-1}\!\left(f(x_i) \right) \\[0.5ex]
        W_i &\sim \operatorname{Binomial}(N_i, p_i) \label{well_p_end}
    \end{align}
\end{subequations}
where $p_i$ is the unobserved ``true'' proportion of wells that conduct flaring in month $i$, $N_i$ is the total number of active wells in month $i$, and $W_i$ is the observed (i.e., estimated and reported by company) number of wells that conduct flaring in month $i$. The rest of the symbols have the same meaning as in Model~\ref{eq:gp_gas_p}.

To demonstrate this model's capability on actual data, both the Blue Buttes Oilfield and Operator A are tested. For the oilfield, the posterior distributions and trace plots of the hyperparameters are presented in \ref{fig:field_well_proportion_trace}. The posterior predictive samples for the underlying process of well flaring proportion ($p_i$) are demonstrated in \ref{fig:field_well_proportion_ppc}. The visualization strategy (different colors represent different percentiles, etc.) is the same as in Section~\ref{sec:gp_gas_p}.
\csmlongfigure[!htbp]{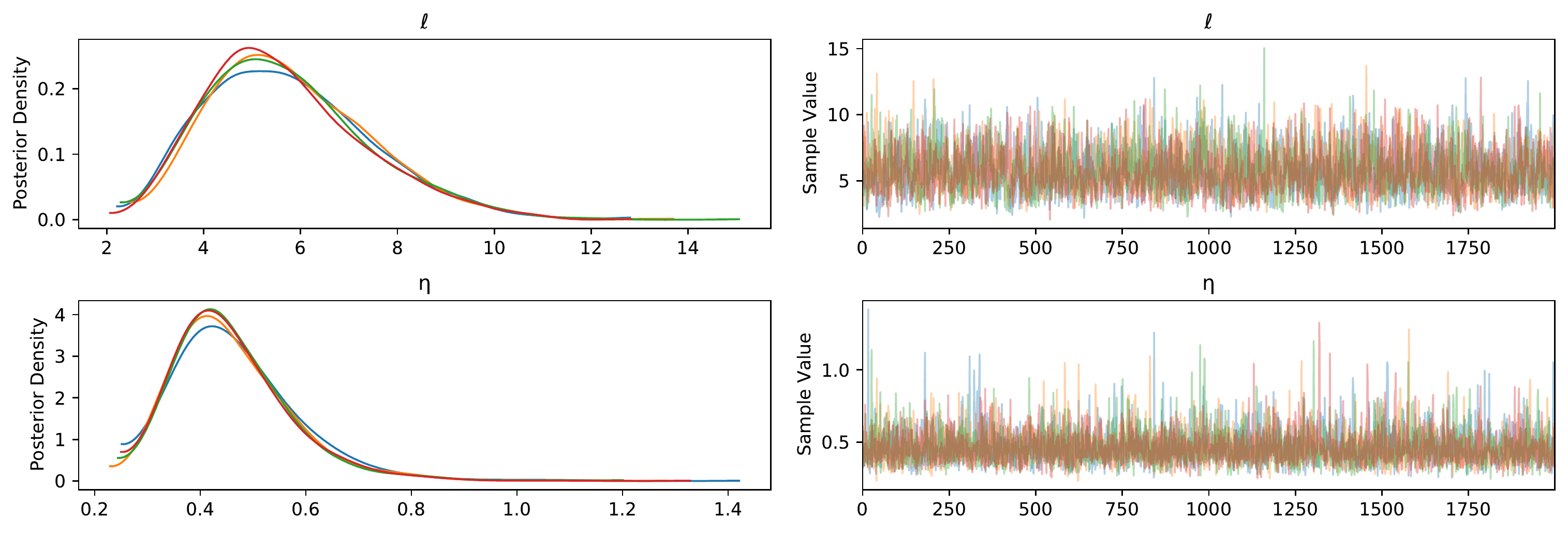}{figures/field_well_proportion_trace}{\textwidth}{Posterior distributions and trace plots for the Blue Buttes Oilfield well flaring proportion model.}{ Well mixing and convergence have been achieved.}
\csmlongfigure[!htbp]{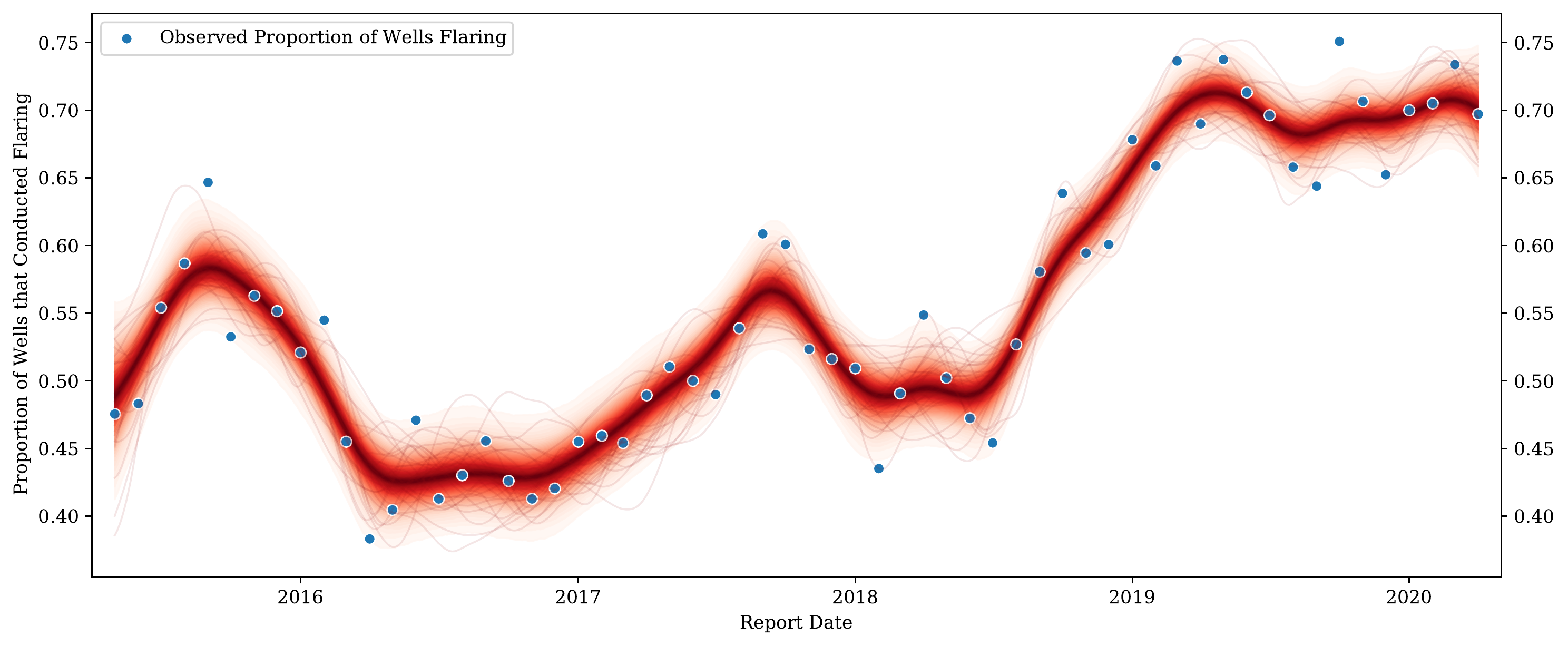}{figures/field_well_proportion_ppc}{\textwidth}{Posterior predictive samples showing the well flaring proportion variations at the Blue Buttes Oilfield.}{ Blue points are the observed data while red lines present the posterior samples.}

With the exact same model specification, this model is also tested with the operator's data. The posterior distributions and trace plots of the hyperparameters are presented in \ref{fig:operator_well_proportion_trace}. The posterior predictive samples for the underlying process of well flaring proportion ($p_i$) are demonstrated in \ref{fig:operator_well_proportion_ppc}. Comparing the two sets of figures from the oilfield and the operator, it can be seen:
    \begin{enumerate}
        \item With the same prior placed on the lengthscale $\ell$, the oilfield model learns from the data and gives a posterior mode around \num{5.5}, whereas the operator model gives a posterior mode around \num{10.0}. This is also reflected in the posterior samples time series plot: the oilfield experienced some well flaring proportion changes in relative shorter time periods, whereas the operator underwent changes on a longer time span.
        \item The oilfield's posterior samples time series show narrower percentile bands while the operator's show wider percentile bands. This is due to the fact that the operator chosen here had smaller number of wells than the oilfield. Since the binomial observation model is used for each month's flaring well count, this naturally represents and quantifies the uncertainties (i.e., binary data contains less information especially when the sample size is small), as well as aligns with the expectation that when there is more data, there should be less uncertainties; when there is less data, there should be more uncertainties.
    \end{enumerate}
\csmlongfigure[!htbp]{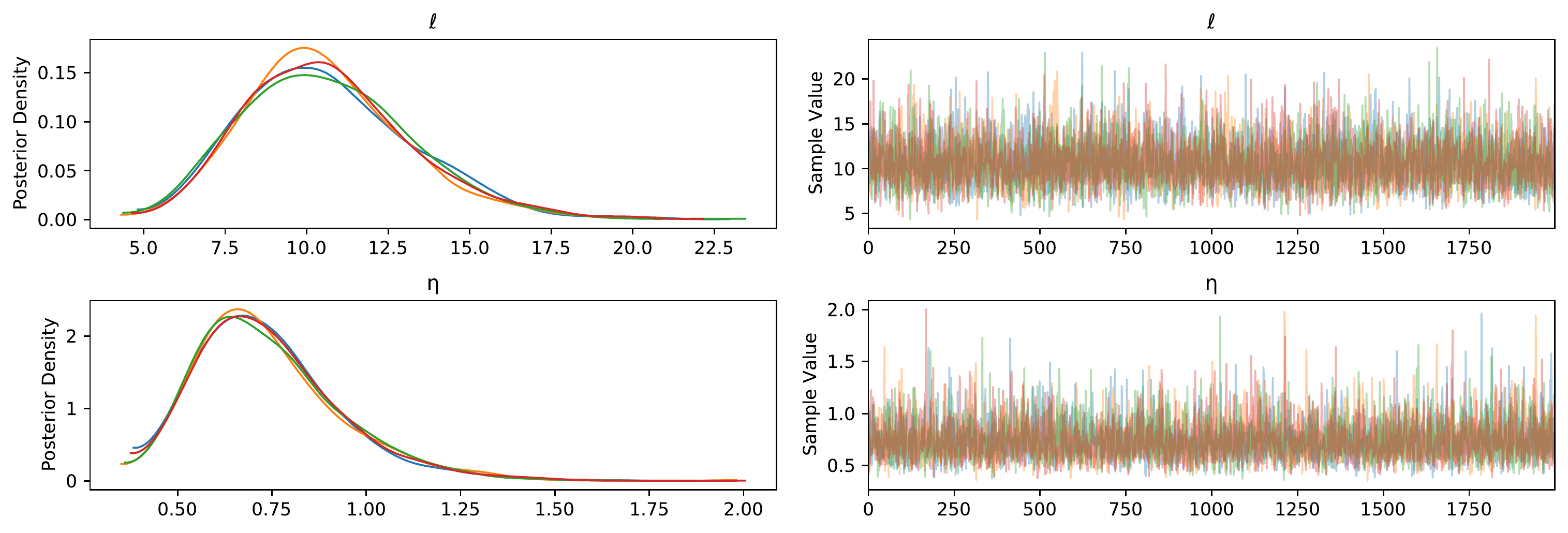}{figures/operator_well_proportion_trace}{\textwidth}{Posterior distributions and trace plots for the Operator A well flaring proportion model.}{ Well mixing and convergence have been achieved. Notice the differences between these inference results and those in \ref{fig:field_well_proportion_trace}, both of which are based on exactly the same priors and likelihood, demonstrating the model specification's wide applicability.}
\csmlongfigure[!htbp]{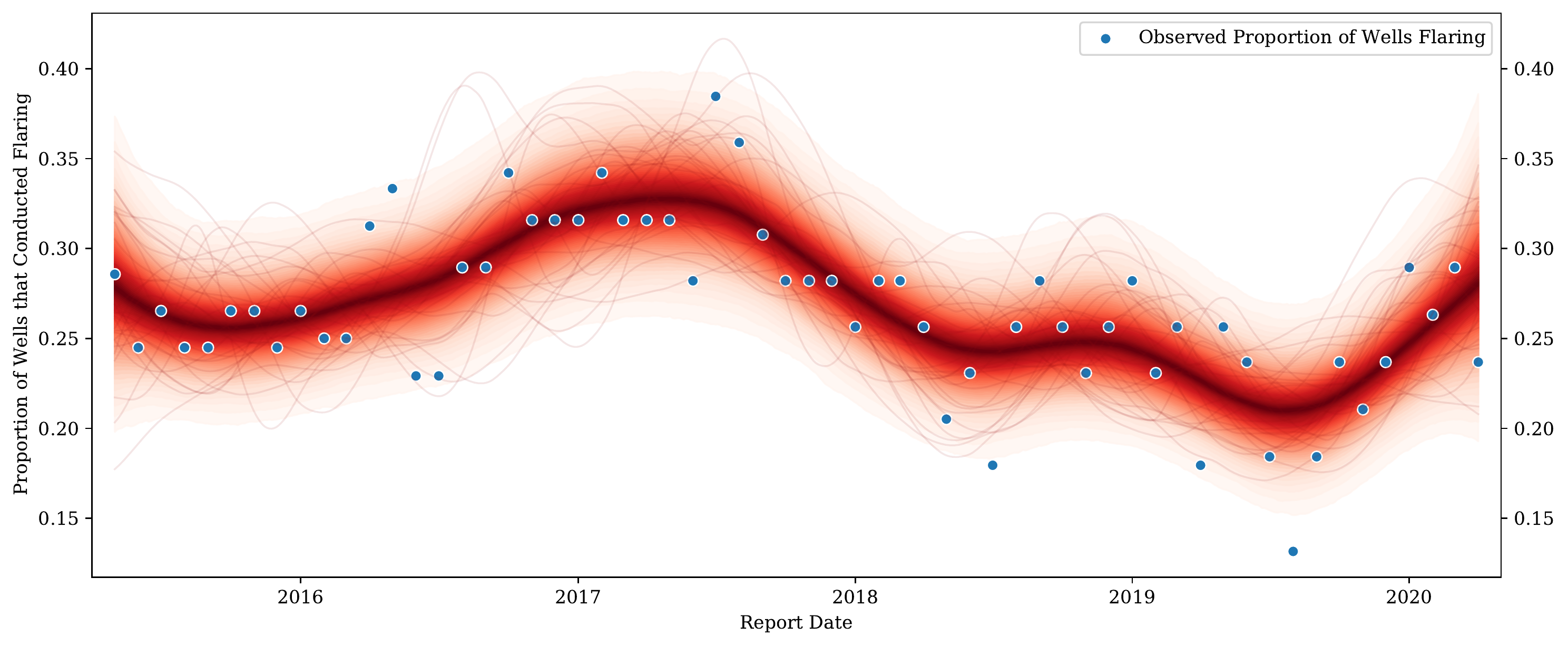}{figures/operator_well_proportion_ppc}{\textwidth}{Posterior predictive samples showing the well flaring proportion variations of Operator~A\@.}{ Blue points are the observed data while red lines present the posterior samples.}

This really showcases how and why to encode domain expertise in flaring data analytics while exploiting machine learning models, which is also the reason to choose the Bayesian approach. One could fit a black box model either with target values $W_i \in \mathds{R}$, or without any probabilistic view (e.g., to optimize for the best deterministic function mapping in the hypothesis space). But either of those would be fundamentally flawed. Domain expertise indicates the well count has to be a non-positive integer, i.e., $W_i \in \mathds{N}_{0}$. Furthermore, neither the NDIC reporting nor the satellite estimation is ever produced in a noise-free environment, and therefore probabilistic modeling is a must. Compared to frequentist machine learning, Bayesian learning is entirely probabilistic and gives one the capability and freedom to encode his/her domain expertise.

\subsubsection{Modeling Flare Detection Count}\label{sec:gp_cox_process}
Satellite detected flare count provides an unbiased indicator of flaring intensity. How this indicator varies in a certain time period for a certain entity is valuable information to obtain. The model is specified through \crefrange{cox_begin}{cox_end}. Essentially the latent process is modeled as a Gaussian Cox process~\autocite{cox_process}, where the Poisson process has varying intensity across time domain and a GP prior is placed on this intensity.
\begin{subequations}\label{mod:cox_flare_ct}
    \begin{align}
        \ell &\sim \operatorname{Gamma}(2, 1) \label{cox_begin} \\[0.5ex]
        \eta &\sim \operatorname{Half-Cauchy}(5) \\[0.5ex]
        k &= \eta^2 \times k_{\textsc{mat\'ern52}}(x, x'; \ell) \\[0.5ex]
        f &\sim \mathcal{GP}(0, k) \\[0.5ex]
        \lambda_i &= \exp\!\left(f(x_i) \right) \\[0.5ex]
        C_i &\sim \operatorname{Poisson}(\lambda_i) \label{cox_end}
    \end{align}
\end{subequations}
where $\lambda_i$ is the unobserved flaring intensity (``true'' count) in month $i$ and $C_i$ is the reported VIIRS detection count in month $i$. Since $\lambda_i$ is bounded to be positive, the natural exponential function is applied to the latent process. The rest of the symbols have the same meaning as in Model~\ref{eq:gp_gas_p}. 

For the task of flaring pattern recognition, the author believes this approach (leveraging a Gaussian Cox process) is a nicer surrogate than a popular change point model presented in~\autocite{bmh,Salvatier2016ProbabilisticPyMC3,stan_man}, which is specified by:
\begin{subequations}\label{eq:classical_cox_process}
    \begin{align}
        e   &\sim  \operatorname{Exponential}(r_e) \\[0.5ex]
        l   &\sim  \operatorname{Exponential}(r_l) \\[0.5ex]
        s   &\sim  \operatorname{Uniform}(1, T) \\[0.5ex]
        C_i &\sim  \operatorname{Poisson}(i < s \; ? \; e \; : \; l)
    \end{align}
\end{subequations}
where $e$ and $l$ are the early and late rates respectively, $r_e$ and $r_l$ controls the priors for the early and late rates, $s$ is the change point, $T$ is the total time period, and the rate in the Poisson likelihood is decided through a ternary conditional operator (?:). The reason is that, although this model could be generalized to more than one change point, its usage is restricted by the assumption that any period between two adjacent change points has a constant rate. This limitation becomes obvious when analyzing the actual flaring data in the discussions below, and is a major disadvantage of the change point model.

The Gaussian Cox process model is tested with the Blue Buttes Oilfield's data. Since only VIIRS data is used, the whole time series is analyzed beginning in 2012. The posterior distributions and trace plots of the hyperparameters are presented in \ref{fig:field_coxProcess_trace}. The posterior predictive samples for the underlying process of flare count ($C_i$) are demonstrated in \ref{fig:field_coxProcess_ppc}. The visualization strategy (different colors represent different percentiles, etc.) is the same as in Section~\ref{sec:gp_gas_p}. From the time series plot, it can be seen the observations from 2014 to 2017 can possibly be described by a change point model (with late 2015 being a potential change point), but the steady growth before and after that time span will frustrate accurate inference with such a model.
\csmlongfigure[!htbp]{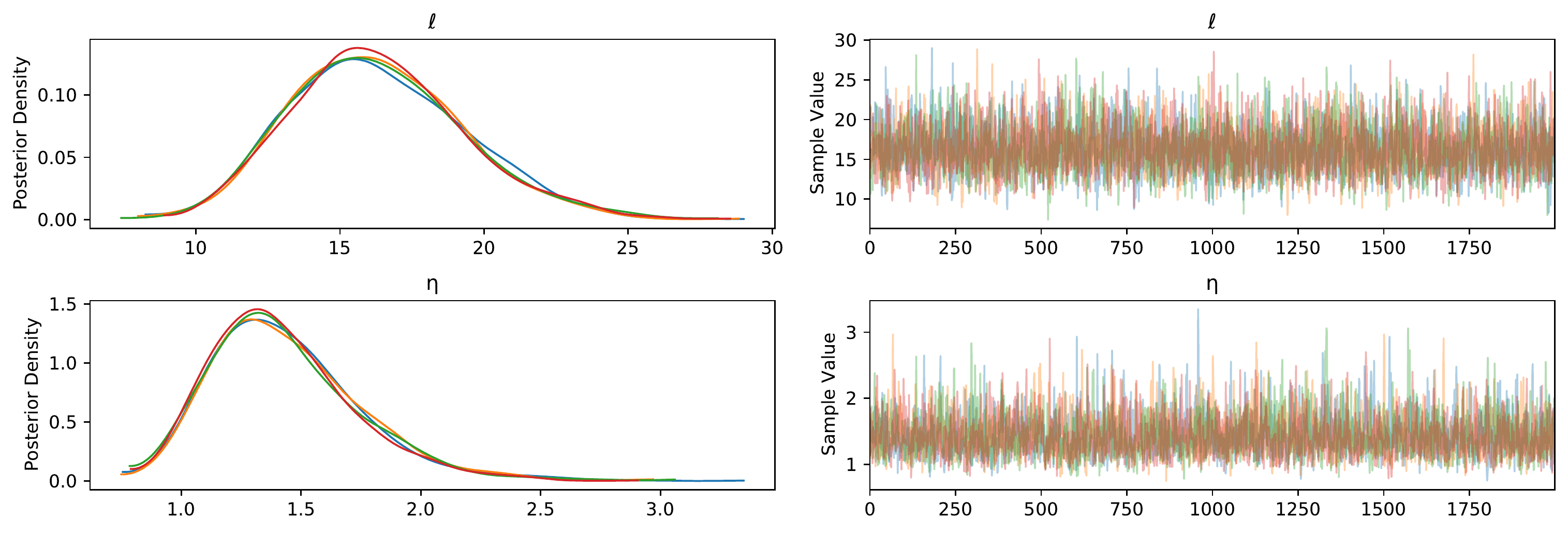}{figures/field_coxProcess_trace}{\textwidth}{Posterior distributions and trace plots for the Blue Buttes Oilfield flare count model.}{ Well mixing and convergence have been achieved.}
\csmlongfigure[!htbp]{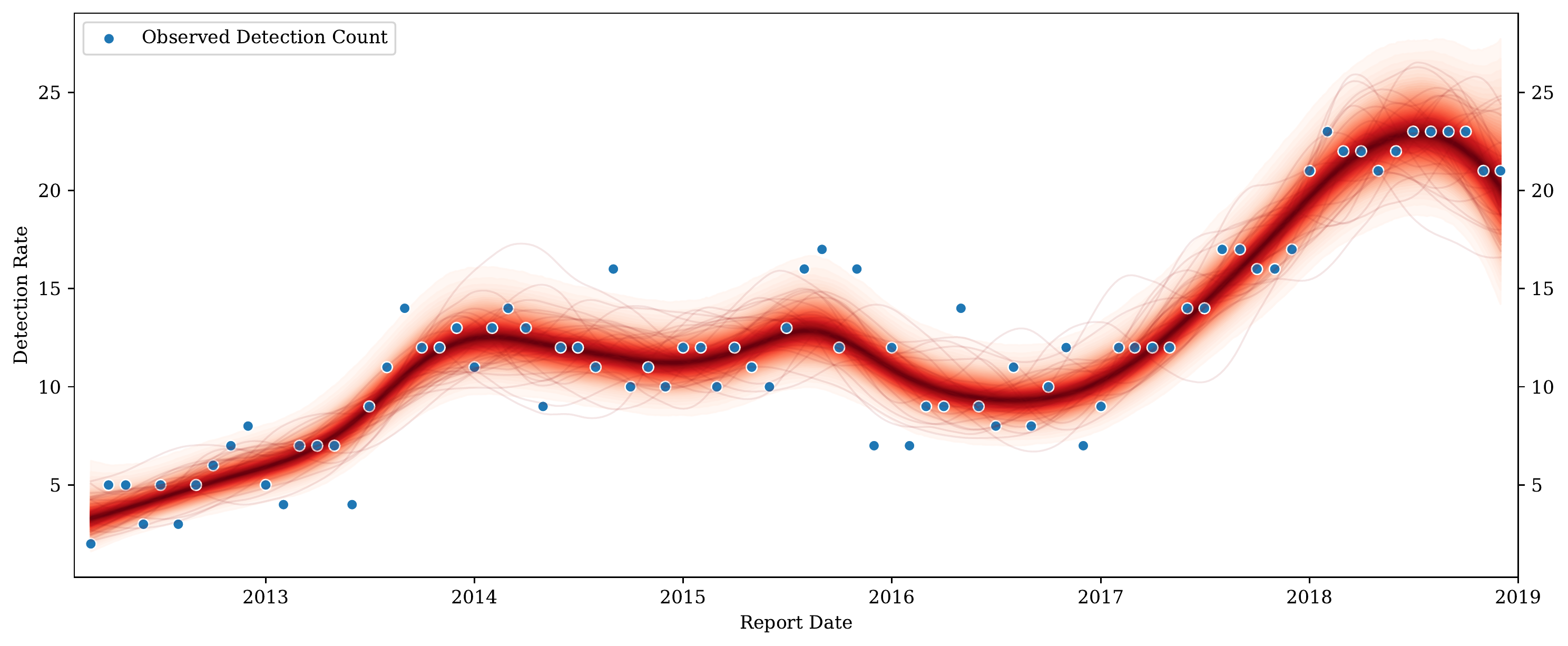}{figures/field_coxProcess_ppc}{\textwidth}{Posterior predictive samples showing the flare count variations at the Blue Buttes Oilfield.}{ Blue points are the observed data while red lines present the posterior samples.}

This model's inference results serve as a type of confirmation, if not evidence, in terms of whether or not an entity achieves the goal/target in reducing the number of wells flaring, when the detection count is used as a surrogate for the number of wells flaring. In practice, reducing the number of wells flaring is exactly the second goal of the regulatory policy introduced by the \citeauthor{NorthDakotaIndustrialCommission2014North041718} in 2014. If the state government is interested in this order's effectiveness from a macroscopic standpoint, the model can also be used to conduct inferences with the state level data. In this case, the posterior distributions and trace plots of the hyperparameters are presented in \ref{fig:stateAllYr_coxProcess_trace}. The posterior predictive samples for the underlying process of flare count ($C_i$) are demonstrated in \ref{fig:stateAllYr_coxProcess_ppc}.
\csmlongfigure[!htbp]{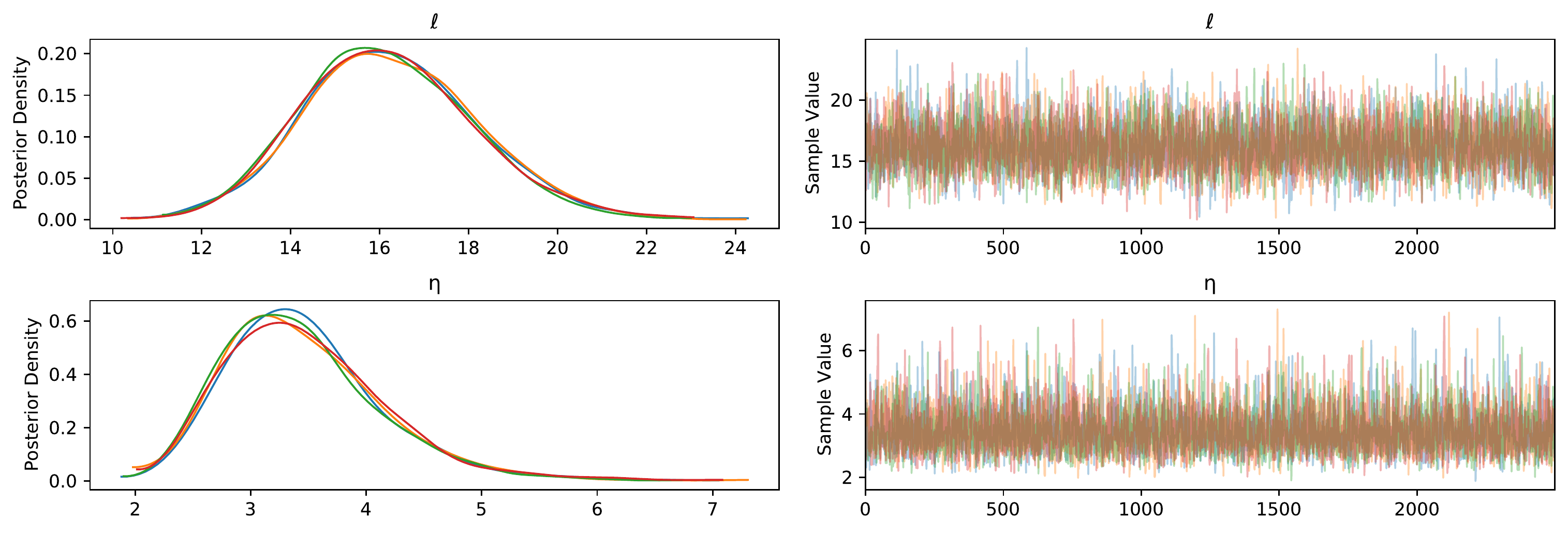}{figures/stateAllYr_coxProcess_trace}{\textwidth}{Posterior distributions and trace plots for the North Dakota flare count model.}{ Well mixing and convergence have been achieved. Notice the differences between these inference results and those in \ref{fig:field_coxProcess_trace}, both of which are based on exactly the same priors and likelihood, demonstrating the model specification's wide applicability.}
\csmlongfigure[!htbp]{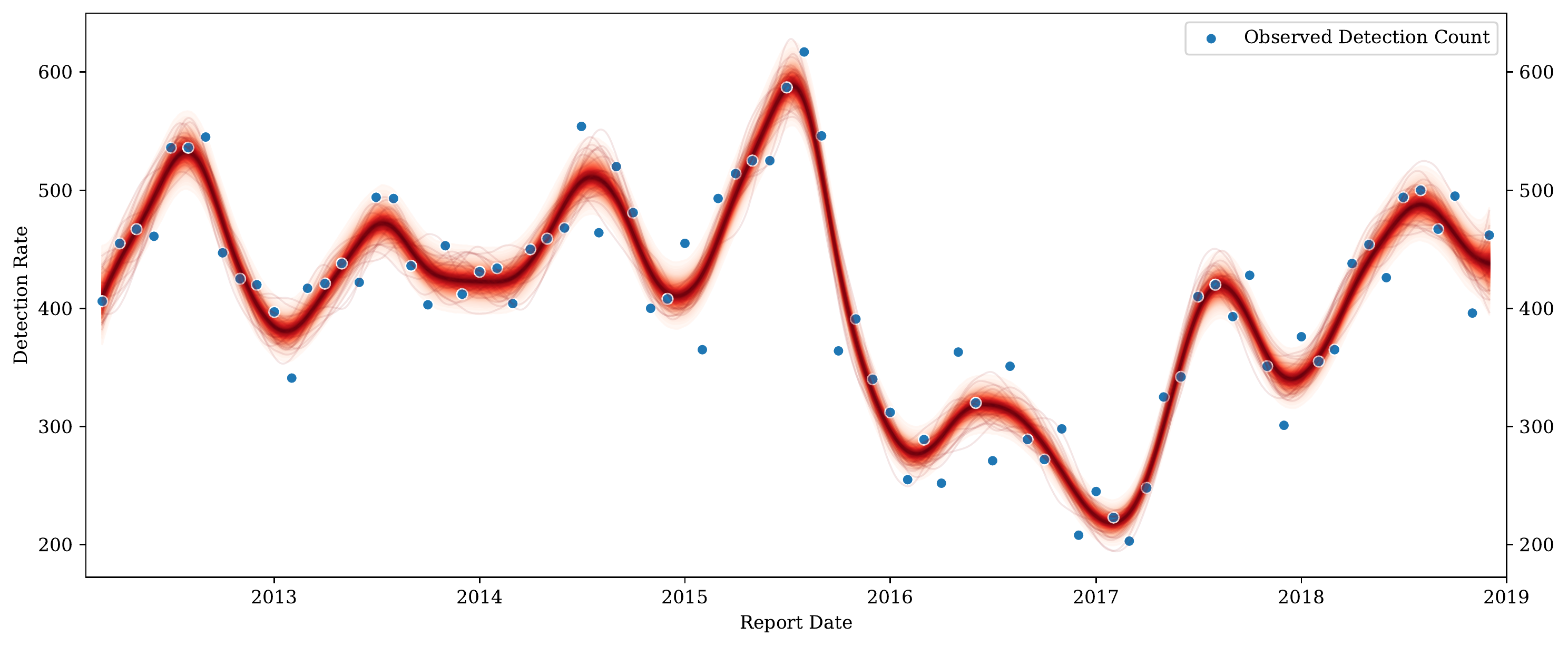}{figures/stateAllYr_coxProcess_ppc}{\textwidth}{Posterior predictive samples showing the flare count variations in North Dakota.}{ Blue points are the observed data while red lines present the posterior samples.}

The percentile bands in this case are quite narrow, which indicate greater confidence in the inferences about the data generating process given the model assumptions. By not (over)fitting to each and every observation, interesting patterns are discovered, for example in every year there is one and only one peak that happened around June. It is worth pointing out that there is no model that can tell the modeler if his/her assumptions are good, only domain expertise might. This model employing a Poisson observation model could be considered ``rigid'' due to the fact that a Poisson likelihood has only one parameter $\lambda$ (to control both the mean and variance) and, furthermore, when $\lambda$ is large as in this scenario, a Poisson distribution is well approximated by a normal distribution. Whenever the state government believes that overdispersion might exist, other observation models such as the negative binomial distribution could be considered. In such cases, only \cref{cox_end} needs to be changed to the negative binomial likelihood, with a prior added for the overdispersion parameter. The specific parameterization is given by \Cref{eq:neg-bin} in Section~\ref{sec:neg-bin}. This really showcases both the flexibility and interpretability of taking a Bayesian approach for high-stakes decision making areas including flaring data analytics.

\subsubsection{Modeling Proportion of Oil Flared}
As crude oil (as opposed to natural gas) is the main commodity at this time, the amount of gas in a barrel of oil equivalent (\si{\boe}) that is flared provides an indicator of production efficiency due to flaring. In this work, the normalized quantity, proportion of oil production being flared, is used such that the model specification is generic for large and small entities. The model is specified through \crefrange{oil_p_begin}{oil_p_end}:
\begin{subequations}\label{mod:oil_proportion_flare}
    \begin{align}
        \ell &\sim \operatorname{Gamma}(2, 1) \label{oil_p_begin} \\[0.5ex]
        \eta &\sim \operatorname{Half-Cauchy}(5) \\[0.5ex]
        \nu &\sim \operatorname{Gamma}(2, 0.1) \\[0.5ex]
        \hat{\sigma}^2 &\sim \operatorname{Half-Cauchy}(5) \\[0.5ex]
        k &= \eta^2 \times k_{\textsc{mat\'ern52}}(x, x'; \ell) \\[0.5ex]
        f &\sim \mathcal{GP}(0, k) \\[0.5ex]
        \pi_i &= \operatorname{logit}^{-1}\!\left(f(x_i) \right) \\[0.5ex]
        \mu_i &= \pi_i \times O_i \\[0.5ex]
        c &= \frac{\SI{6}{\mcf}}{\SI{1}{\boe}} \\[0.5ex]
        F_i/c \eqqcolon E_i &\sim \stut(\nu, \mu_i, 1/\hat{\sigma}^2) \label{oil_p_end}
    \end{align}
\end{subequations}
where:
\begin{description}[noitemsep,itemindent=-8em,leftmargin=6em]
    \item[\pi_i] is the underlying flaring \si{\boe} proportion of month $i$;
    \item[O_i] is the total oil production of month $i$;
    \item[\mu_i] denotes the ``true'' flared \si{\boe} of month $i$;
    \item[c] denotes the conversion factor that \SI{6}{\mcf} equals \SI{1}{\boe}, given by the \textcite{usgs_boe};
    \item[E_i] is the reported flared \si{\boe}, which is modeled using a $\stut$ observation model.
\end{description}
The rest of the symbols have the same meaning as in Model~\ref{eq:gp_gas_p}. To test this model's performance on real data, both the Blue Buttes Oilfield and Operator A are used. For the oilfield, the posterior distributions and trace plots of the hyperparameters are presented in \ref{fig:field_BOE_proportion_trace}. The posterior predictive samples for the underlying process of \si{\boe} flaring proportion ($\pi_i$) are demonstrated in \ref{fig:field_BOE_proportion_ppc}. The visualization strategy (different colors represent different percentiles, etc.) is the same as in Section~\ref{sec:gp_gas_p}.
\csmlongfigure[!htbp]{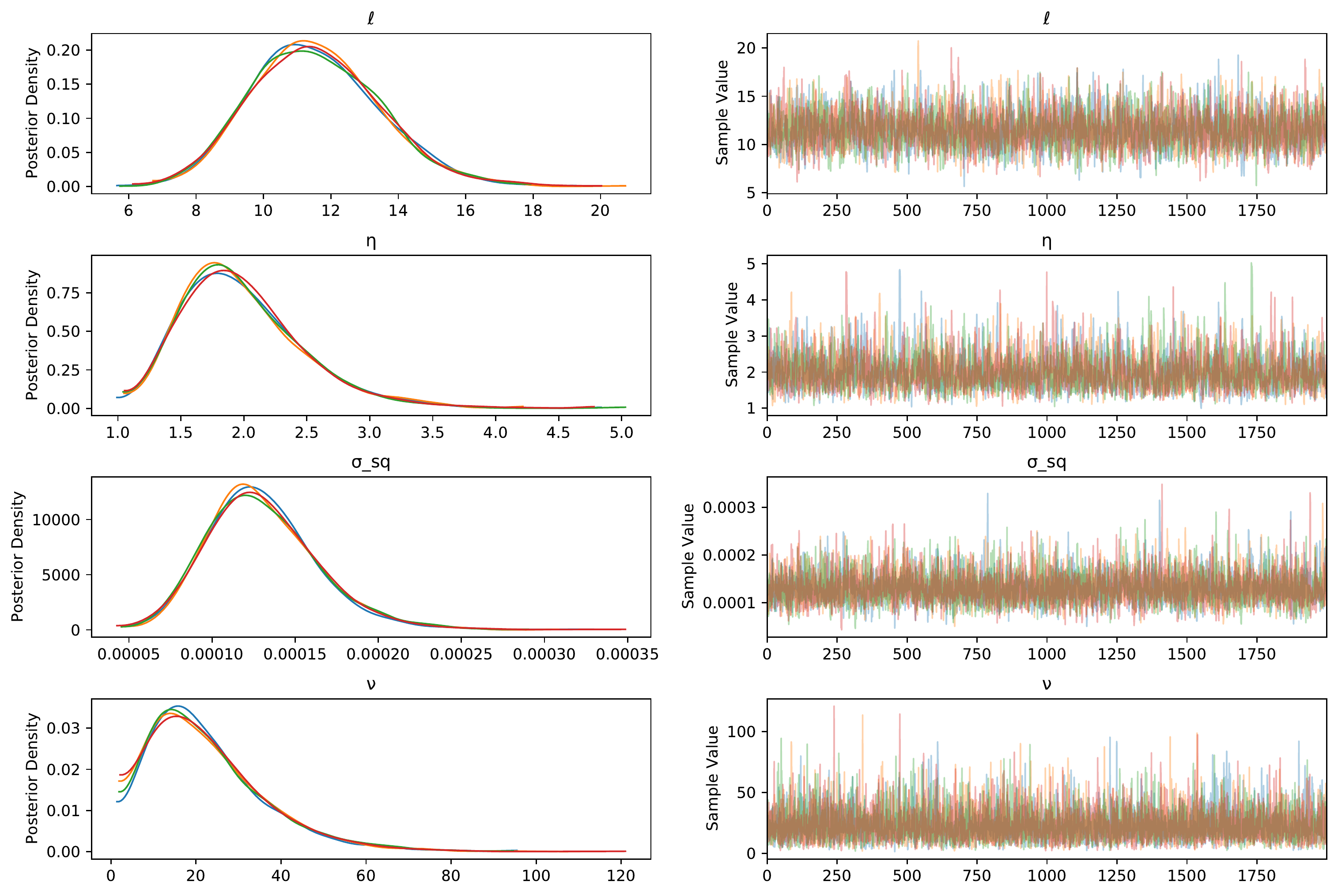}{figures/field_BOE_proportion_trace}{\textwidth}{Posterior distributions and trace plots for the Blue Buttes Oilfield \si{\boe} flaring proportion model.}{ Well mixing and convergence have been achieved.}
\csmlongfigure[!htbp]{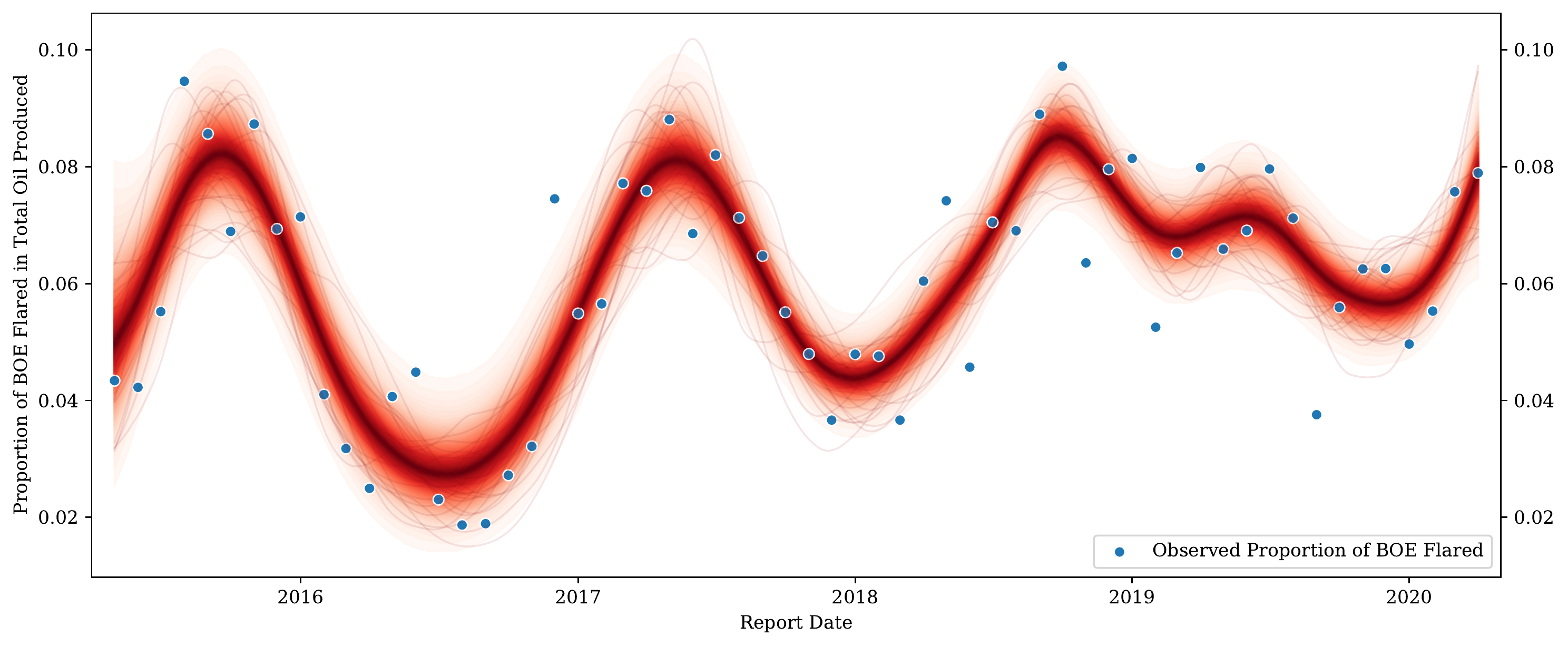}{figures/field_BOE_proportion_ppc}{\textwidth}{Posterior predictive samples showing the \si{\boe} flaring proportion variations at the Blue Buttes Oilfield.}{ Blue points are the observed data while red lines present the posterior samples.}

With the exact same model specification, this model is also tested with the operator's data. The posterior distributions and trace plots of the hyperparameters are presented in \ref{fig:operator_BOE_proportion_trace}. The posterior predictive samples for the underlying process of \si{\boe} flaring proportion ($\pi_i$) are demonstrated in \ref{fig:operator_BOE_proportion_ppc}. 
\csmlongfigure[!htbp]{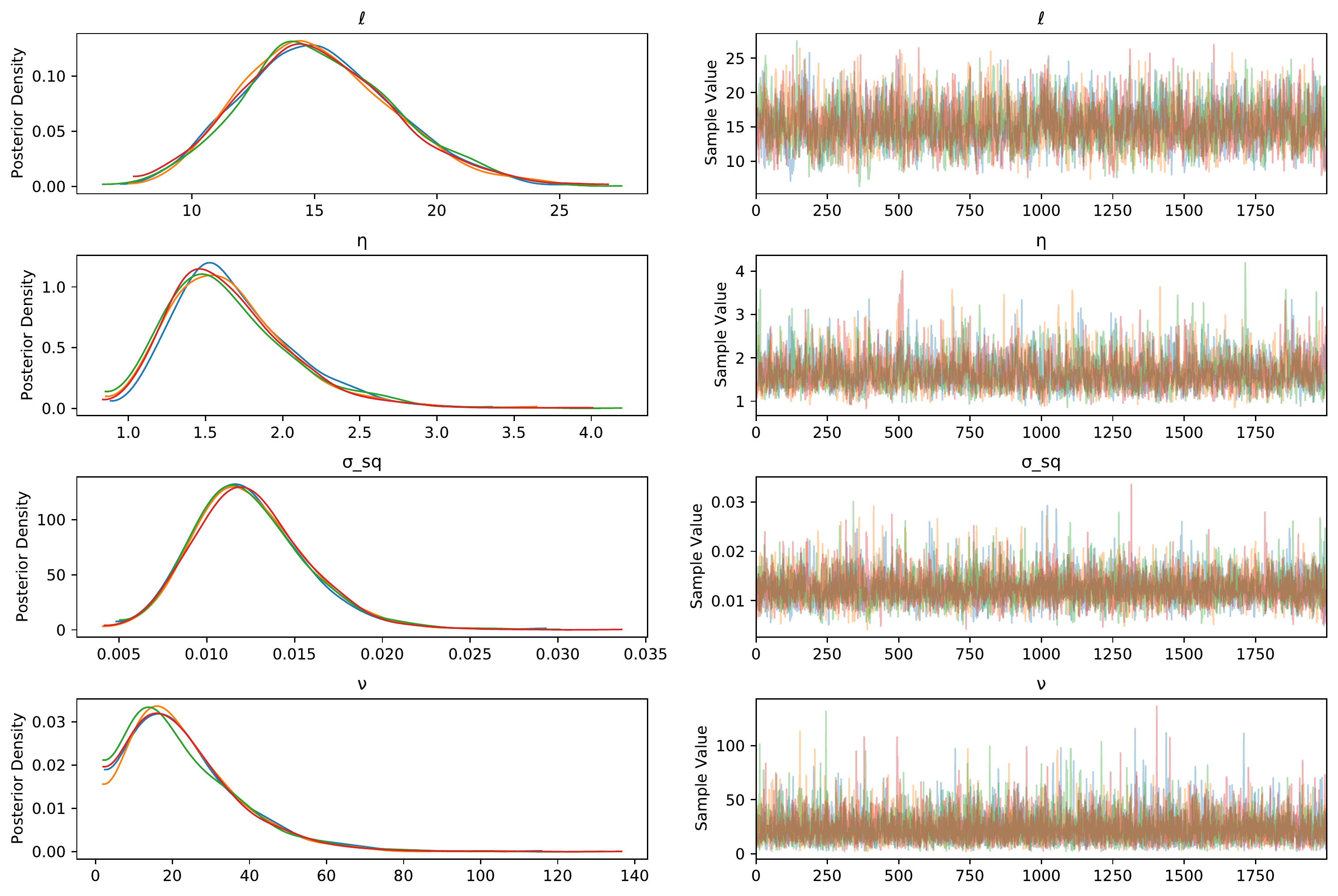}{figures/operator_BOE_proportion_trace}{\textwidth}{Posterior distributions and trace plots of the \si{\boe} flaring proportion model for Operator~A\@.}{ Well mixing and convergence have been achieved. Notice the differences between these inference results and those in \ref{fig:field_BOE_proportion_trace}, both of which are based on exactly the same priors and likelihood, demonstrating the model specification's wide applicability.}
\csmlongfigure[!htbp]{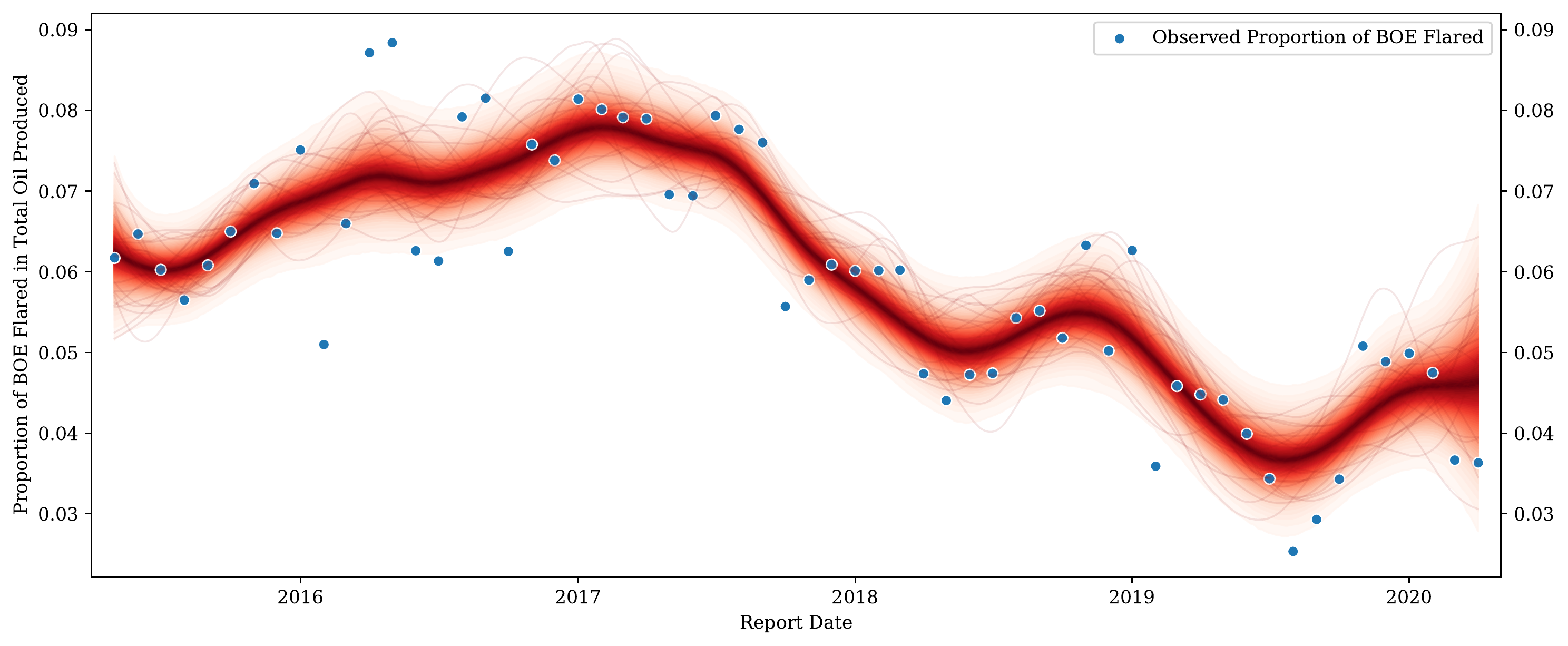}{figures/operator_BOE_proportion_ppc}{\textwidth}{Posterior predictive samples showing the \si{\boe} flaring proportion variations of Operator~A\@.}{ Blue points are the observed data while red lines present the posterior samples.}

Comparing the two sets of figures from the oilfield and the operator, it can be observed:
    \begin{enumerate}
        \item With the same prior placed on the lengthscale $\ell$, which has a mean of \num{2} (months), both models have updated the posterior to move away from this mean, reflecting a long range variation. The oilfield has a posterior mode about \num{1} year while the operator has a mode around \num{15} months. The operator has much larger reporting variability, shown by the parameter $\hat{\sigma}^2$.
        \item With a $\stut$ likelihood, both models demonstrate robustness to outliers and overfitting. This can be seen from the oilfield's late 2019 observations and the operator's early 2016 observations. For the posterior function samples, shown as the thin lines, some of them are indeed pulled towards those ``outliers''. However, the percentile plots (shown as the colored bands) are not impacted and those really can be interpreted as the trend which is most compatible with the data and the assumptions. This built-in Occam's razor of the Bayesian approach when choosing appropriate priors is very impressive. In many of the frequentist machine learning methods, if the regularization strategy is not implemented well especially when the sample size is not huge enough for the asymptotic properties to kick in, outliers become ``influential observations'' that will have a huge undesirable effect on the inference results.
    \end{enumerate}

\subsubsection{Modeling Scale Factor between VIIRS and NDIC}
Both NDIC and VIIRS reporting give (estimated) flared gas volume. The scale factor between the two sources provides insights into whether NDIC reporting is consistent:
\begin{enumerate}
    \item for different entities (e.g., among a group of operators), and 
    \item for one entity when looking at a certain time period.
\end{enumerate}
This is based on the fact that the satellite detection processing algorithm is unbiased and consistent. Item 2 is particularly interesting in terms of time series analytics. The model is specified through \crefrange{sf_begin}{sf_end}:
\begin{subequations}\label{mod:sf_viirs_ndic}
    \begin{align}
        \ell_{\textsc{mat}} &\sim \operatorname{Gamma}(8, 2) \label{sf_begin} \\[0.5ex]
        \eta_{\textsc{mat}} &\sim \operatorname{Half-Cauchy}(5) \\[0.5ex]
        T &\sim \mathcal{N}(12, 1) \\[0.5ex]
        \ell_{\textsc{per}} &\sim \operatorname{Gamma}(4, 3) \\[0.5ex]
        \eta_{\textsc{per}} &\sim \operatorname{Half-Cauchy}(5) \\[0.5ex]
        \nu &\sim \operatorname{Gamma}(2, 0.1) \\[0.5ex]
        \hat{\sigma}^2 &\sim \operatorname{Half-Cauchy}(5) \\[0.5ex]
        k_{\textsc{mat}} &= \eta_{\textsc{mat}}^2 \times k_{\textsc{mat\'ern52}}(x, x'; \ell_{\textsc{mat}}) \\[0.5ex]
        k_{\textsc{per}} &= \eta_{\textsc{per}}^2 \times k_{\textsc{periodic}}(x, x'; T, \ell_{\textsc{per}}) \\[0.5ex]
        k_{\textsc{wn}} &= k_{\mathsmaller{\textsc{WhiteNoise}}}(x, x'; \delta=\num[output-exponent-marker=\ensuremath{\mathrm{e}}]{1e-6}) \\[0.5ex]
        f &\sim \mathcal{GP}(0, k_{\textsc{mat}} + k_{\textsc{per}} + k_{\textsc{wn}}) \\[0.5ex]
        \beta_i &= \exp\!\left(f(x_i) \right) \\[0.5ex]
        \mu_i &= \beta_i \times \mathrm{VIIRS}_i \\[0.5ex]
        \mathrm{NDIC}_i &\sim \stut(\nu, \mu_i, 1/\hat{\sigma}^2) \label{sf_end}
    \end{align}
\end{subequations}
where:
\begin{description}[noitemsep,itemindent=-8em,leftmargin=6em]
    \item[\ell_{\textsc{mat}}] is the lengthscale for the Mat\'ern kernel;
    \item[\eta_{\textsc{mat}}] is the marginal deviation for the Mat\'ern kernel;
    \item[T] is the period for the periodic kernel;
    \item[\ell_{\textsc{per}}] is the lengthscale for the periodic kernel;
    \item[\eta_{\textsc{per}}] is the marginal deviation for the periodic kernel;
    \item[k_{\textsc{mat}}] is the Mat\'ern kernel (component);
    \item[k_{\textsc{per}}] is the periodic kernel (component);
    \item[k_{\textsc{wn}}] is the white noise kernel (component);
    \item[f] denotes the latent process, which is distributed according to a GP whose covariance function is the sum of \num{3} kernels;
    \item[\beta_i] is the underlying scale factor between VIIRS and NDIC of month $i$. Since this scale factor is bounded to be positive, the natural exponential function is applied to the latent process;
    \item[\mathrm{VIIRS}_i] is the VIIRS reported volume of month $i$;
    \item[\mu_i] denotes the underlying flared volume of month $i$;
    \item[\mathrm{NDIC}_i] is the NDIC reported volume of month $i$, which is modeled using a $\stut$ observation model.
\end{description}
The rest of the symbols have the same meaning as in Model~\ref{eq:gp_gas_p}. The reason for adding a periodic kernel is to investigate if there are any seasonal patterns. Maintaining a proper Bayesian workflow lets the data speak for itself, i.e., whether there exists seasonal behaviors or not, as shown by the two case studies in this section.

The model is first fitted with the state level data to investigate the macroscopic reporting consistency. The posterior distributions and trace plots of the hyperparameters are presented in \ref{fig:nd_SF_trace}. The posterior predictive samples for the underlying process of the scale factor variations ($\beta_i$) are demonstrated in \ref{fig:nd_SF_ppc}. The visualization strategy (different colors represent different percentiles, etc.) is the same as in Section~\ref{sec:gp_gas_p}. From the posterior time series plot, it can be seen in general the volumes from NDIC reporting is smaller than that of VIIRS reporting, except for the times when the total flaring magnitude was small (indicated by the smaller points). More importantly, within each and every year from 2015 to 2018, there is a decreasing trend in the values of the scale factor ($\beta_i$) around midyear. Each year's latent process from Q2 to Q3 can be viewed as a ``seesaw'', with July being the middle pivot point and the months after July always going down. Note that within each year, the NDIC reporting of flared volumes might increase steadily or a lot (which was actually happening from the time series plot in \ref{fig:eda_state_data_vis}), however this scale factor declining trends indicate the satellites observed much greater flaring activities than what was reported by the companies! This finding suggests that the NDIC reporting is very likely not consistent throughout the year, and the state government should be concerned that some companies might underreport their flared volumes especially in the second half of the year.
\csmlongfigure[!htbp]{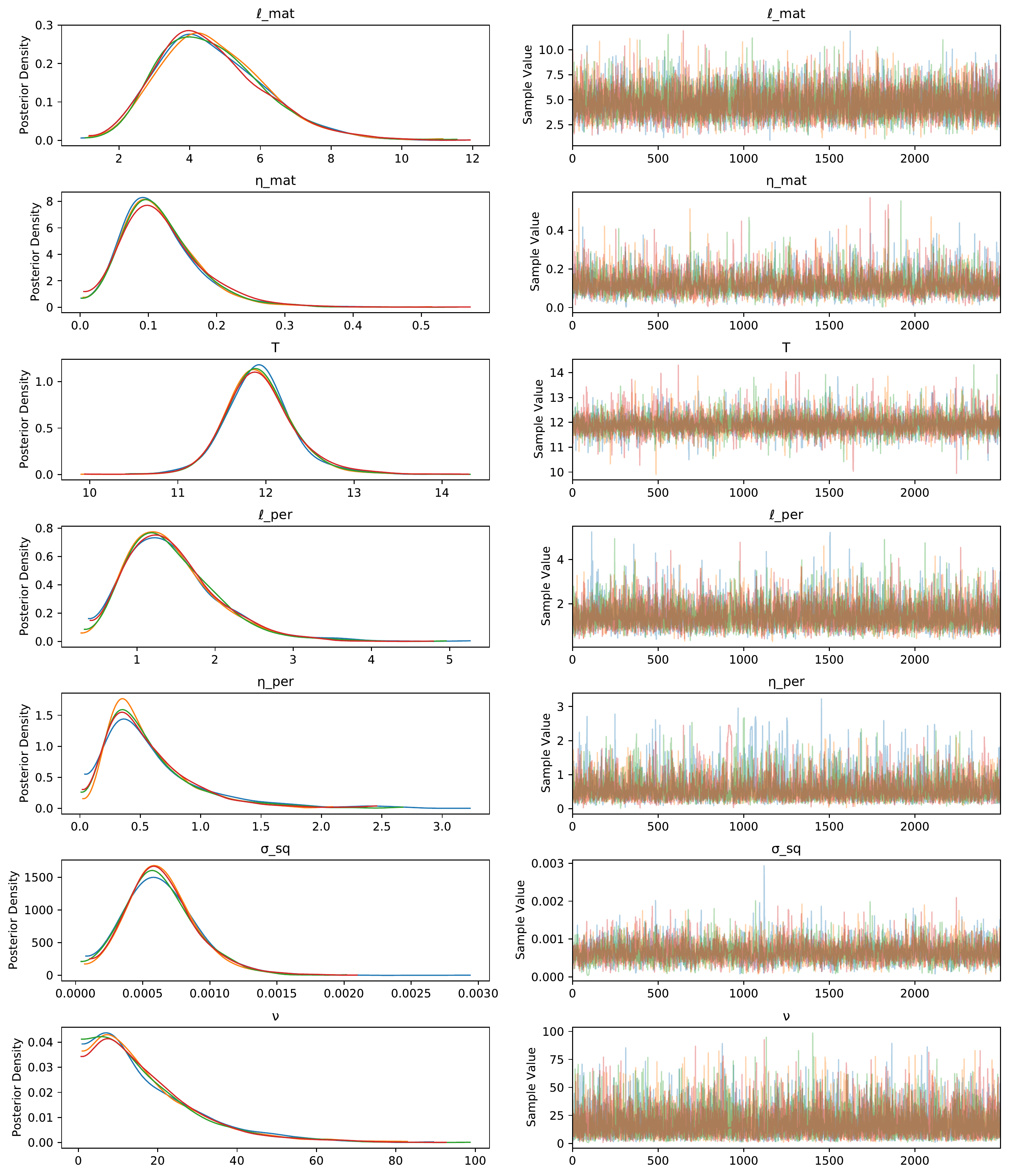}{figures/nd_SF_trace}{\textwidth}{Posterior distributions and trace plots for the North Dakota VIIRS-NDIC scale factor model.}{ Well mixing and convergence have been achieved.}
\csmlongfigure[!htbp]{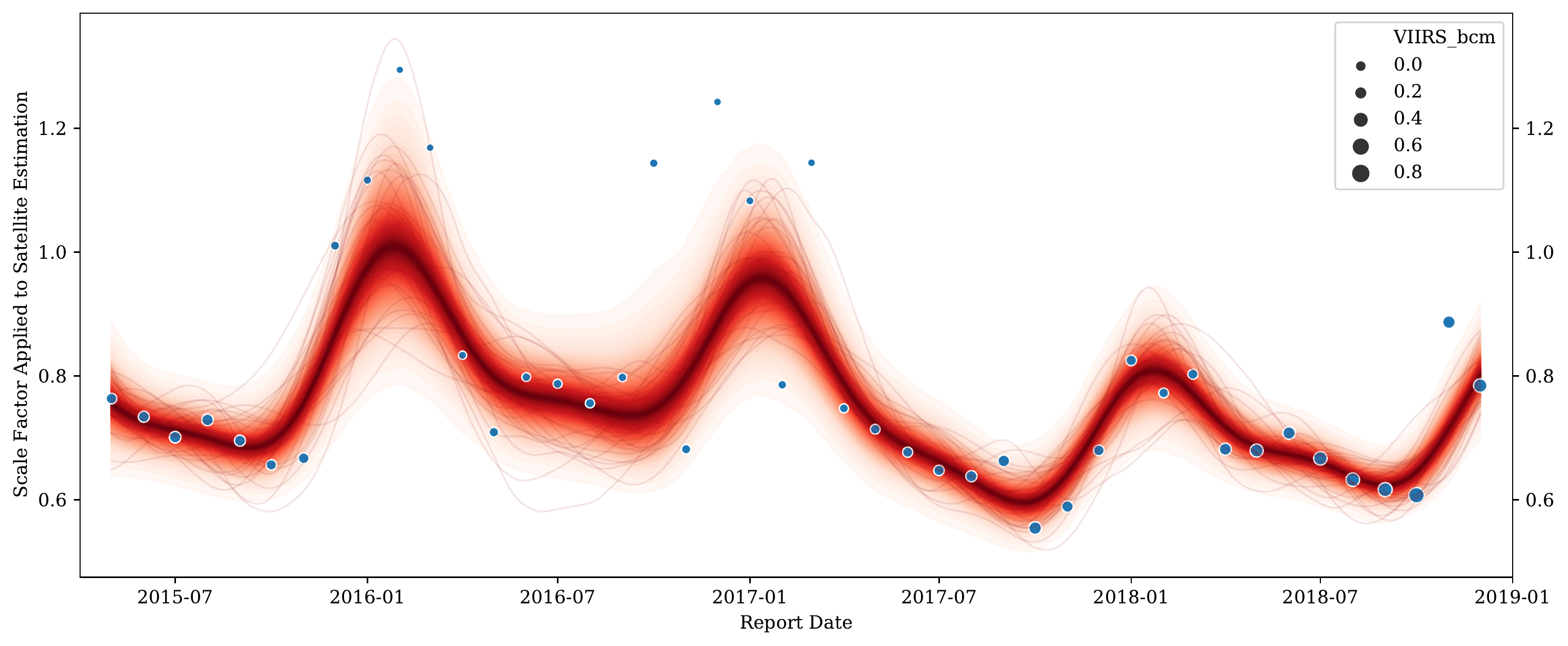}{figures/nd_SF_ppc}{\textwidth}{Posterior predictive samples showing the scale factor variations of North Dakota.}{ Blue points are the observed data while red lines present the posterior samples. Larger points indicate greater flaring magnitude as observed from VIIRS.}

A interesting question arises: is this seasonal behavior universal across all the entities? The answer is unfortunately no, which indicates some operators likely reported their flared volume in an inconsistent manner throughout the entire year. In fact, if the Blue Buttes Oilfield data is used to fit the model, rather consistent behavior is observed. In this case, the posterior distributions and trace plots of the hyperparameters are presented in \ref{fig:oilfield_SF_trace}. The posterior predictive samples for the underlying process of the scale factor variations ($\beta_i$) are demonstrated in \ref{fig:oilfield_SF_ppc}. With the exact same model specification incorporating the periodic kernel, no apparent seasonal behaviors are discerned by the inference process. There are much uncertainties around the time of early 2016, where the point sizes indicate the overall flaring magnitudes were small as observed from VIIRS, and the NDIC reported volumes were actually larger than that of VIIRS. This could be due to the truncation effects instead of the reporting inconsistencies, i.e., when the flares are sporadic and weaker, they are not easily captured by the satellites, resulting in a truncated sample for the VIIRS processing workflow. By applying this model and workflow to the other major producing fields, it will likely pick up the ones who have the ``seesaw'' behaviors in their reporting.
\csmlongfigure[!htbp]{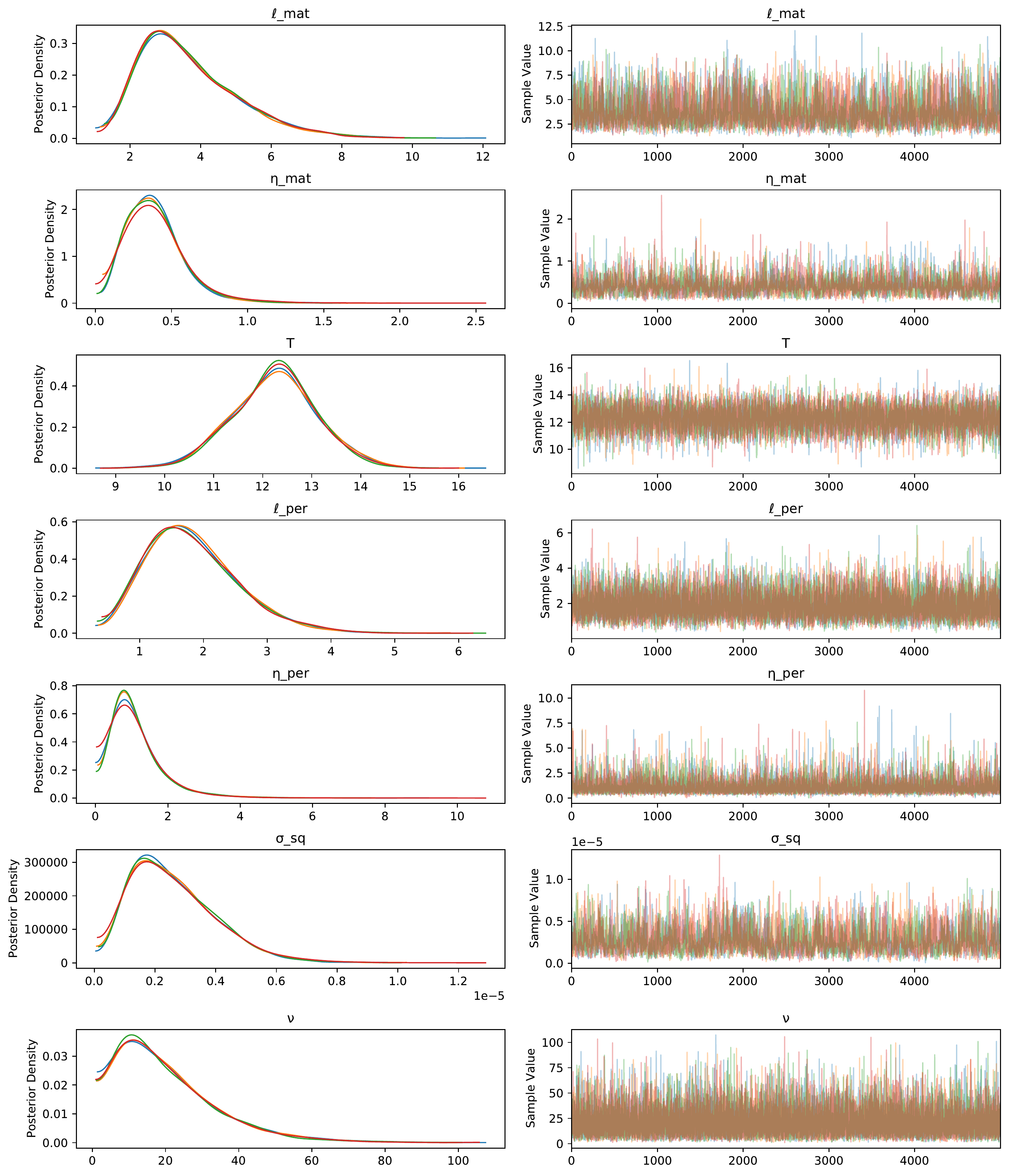}{figures/oilfield_SF_trace}{\textwidth}{Posterior distributions and trace plots for the Blue Buttes Oilfield VIIRS-NDIC scale factor model.}{ Well mixing and convergence have been achieved. Notice the differences between these inference results and those in \ref{fig:nd_SF_trace}, both of which are based on exactly the same priors and likelihood, demonstrating the model specification's wide applicability.}
\csmlongfigure[!htbp]{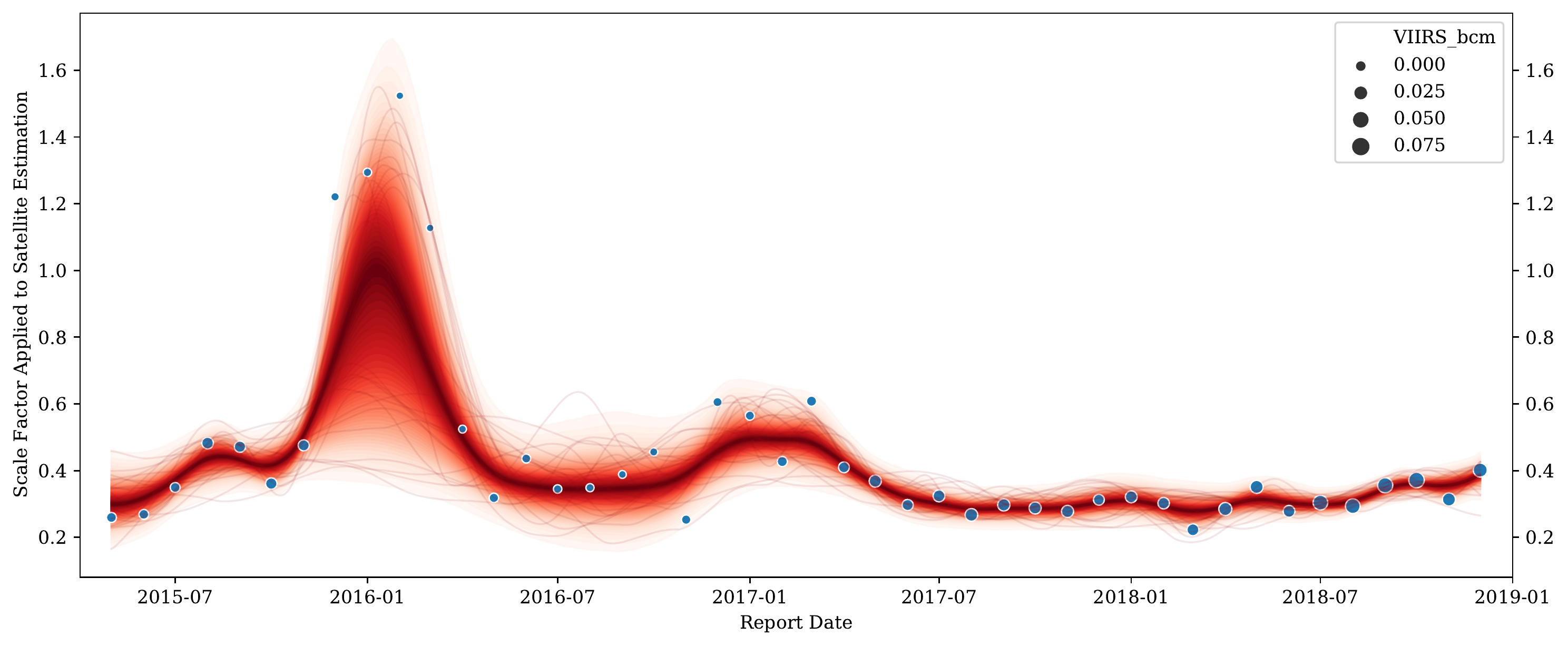}{figures/oilfield_SF_ppc}{\textwidth}{Posterior predictive samples showing the scale factor variations in the Blue Buttes Oilfield.}{ Blue points are the observed data while red lines present the posterior samples. Larger points indicate greater flaring magnitude as observed from VIIRS.}

\subsubsection{Predicting NDIC Flared Volume}
GP is not only fully capable of making predictions once the model hyperparameters are learned, but it can provide rigorously constructed intervals quantifying uncertainties as well through \cref{eq:gp-cond}, for which many of the frequentist machine learning methods fail to do. The author chooses to present one particular prediction case study, that is to predict NDIC reported volume based on the projected scale factor between VIIRS and NDIC. This will be a particular interesting deployment scenario once fast satellite detection/estimation is available, which takes less time than waiting on company reports followed by compiling everything into an analytics-ready format.

The predictions are generated in the form of posterior predictive samples. Along with the historical observations, the predictions of the scale factor for the next six months are presented in \ref{fig:nd_SF_pred}. The very wide percentile bands in the forecasting indicate that the seasonal behaviors will likely take effect again, however with great uncertainties. If point predictions (i.e., without the prediction intervals) are needed, one can always use the posterior mean, mode, etc. to construct that ``best'' function; however this showcases why predicting the future is generally very difficult and uncertainties should always be properly characterized.
\csmlongfigure[!htbp]{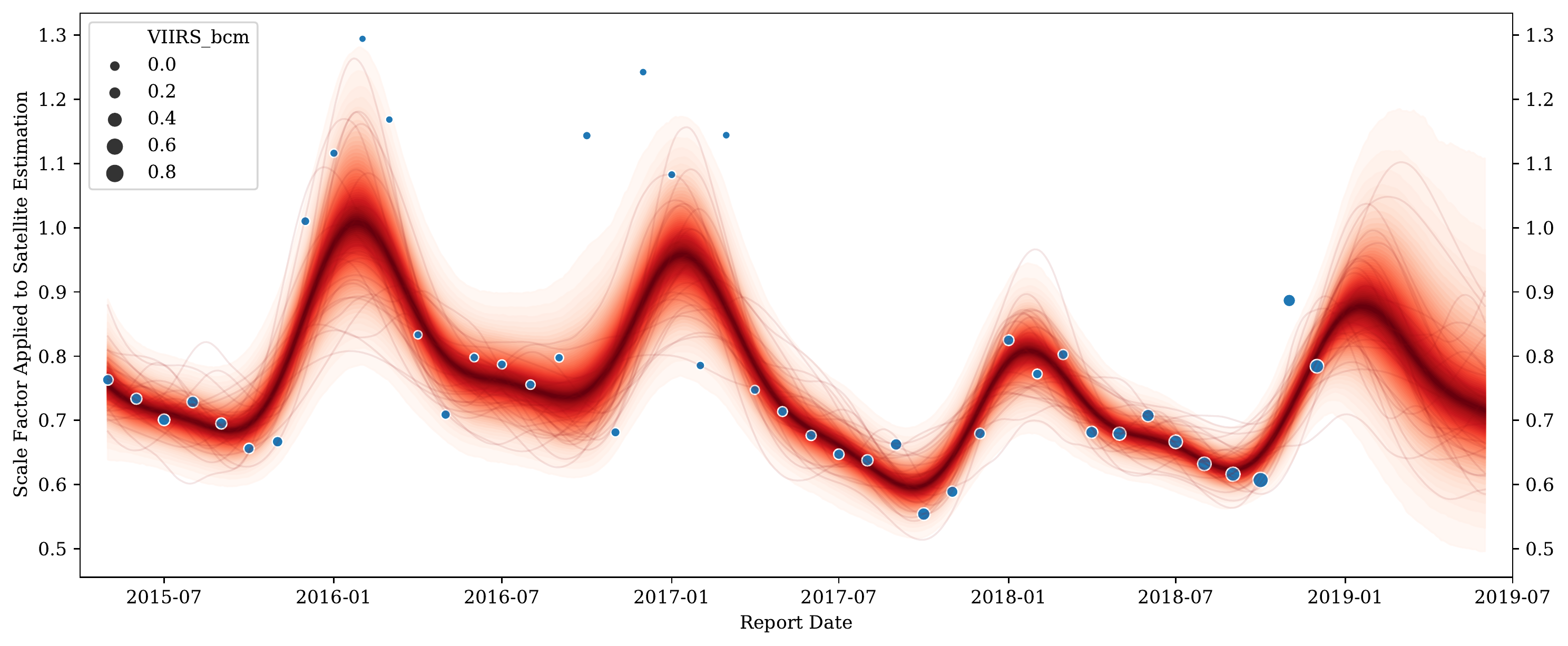}{figures/nd_SF_pred}{\textwidth}{Posterior predictive samples showing predictions of the scale factor for the next six months.}{ Blue points are the observed data while red lines present the posterior samples. Larger points indicate greater flaring magnitude as observed from VIIRS.}

\subsubsection{A Look Back at the Prior Choices}
Looking back at the suite of models developed, the set of priors for the latent functions have been the same (except the scale factor model where a periodic kernel is added). However the posteriors are all updated (i.e., ``learned'') based on each dataset and modeling goal. This means the below set of priors
\begin{subequations}
    \begin{align}
        \ell &\sim \operatorname{Gamma}(2, 1) \\[0.5ex]
        \eta &\sim \operatorname{Half-Cauchy}(5) \\[0.5ex]
        k &= \eta^2 \times k_{\textsc{mat\'ern52}}(x, x'; \ell) \\[0.5ex]
        f &\sim \mathcal{GP}(0, k)
    \end{align}
\end{subequations}
serves as a generic framework and can be recommended for flaring time series analytics in general, in a GP context. Notice this prior choice gives latent function values in the unconstrained space, i.e., $f(x) \in \mathds{R}$. However, in many situations, the domain expertise indicates the quantities of interest live in constrained space, such as:
\begin{itemize}
    \item $\mathds{R}_{>0}$ for Poisson rate parameter when modeling count data, and
    \item $\interval{0}{1}$ for binomial success probability when modeling flaring well proportion.
\end{itemize}
To better reflect the domain expertise, the link functions can be leveraged. For the above scenarios, the log link function and the logit link function can be applied, respectively. Although this prior configuration is the result of several design iterations and tested with real data, there is no reason to think that it is optimal for every entity. Indeed, the model for scale factor between VIIRS and NDIC has bespoke components in its priors. The \textcite{stan_man} also gave some general prior choice recommendations for GP.

The whole suite of models demonstrate full capability of harnessing the temporal structure in flaring time series at different levels for different entities. This provides huge potential for extracting insights from noisy monthly data streams. For the situations where cross-sectional data analytics is desirable, for example when the latest monthly data is available and the state government needs insights from merely that month (before appending it to the whole historical data for a longitudinal study), other types of models can be built. Such is discussed in the next chapter.

\chapter{Unsupervised Learning from Multiple Perspectives}\label{ch:gmm}

\setlength{\epigraphwidth}{0.5\textwidth}
\setlength{\epigraphrule}{0pt}
\epigraph{``Estimation of densities is a universal problem of statistics (knowing the densities one can solve various problems).''}{--- \textcite{Vapnik2000TheTheory}}

\subsection{Learning the Distribution}
In this chapter, the author studies how to describe the flaring related quantities' distribution among the oilfields in North Dakota in a cross-sectional setting. That is, data collected for one point or a period of time (such as a certain month or quarter) is analyzed. In this setting, the data used for learning is unlabeled:
\begin{equation}\label{eq:unsuper_data}
    U = \left\{x_1, x_2, \dots, x_N \right\},
\end{equation}
where $x_i$, $i=1,2,\dots,N$, are the observations for the $i$-th oilfield. Thus unsupervised learning is naturally applied. The model to be learned is in the form of a conditional probability distribution $P_{\bm{\theta}} (x \mid z)$ where $z$ is some latent structure and $\bm{\theta}$ represents the parameters.

This has many application scenarios in practice. When the latest month's or quarter's data is available, the government of North Dakota might need distributional insights of the population (of oilfields), preferably beyond some forms of the order statistics (such as the five-number summary). This cross-sectional study is especially valuable and worth conducting when a direct comparison with previous months/quarters (which can be either the immediately previous one, or the same month/quarter in previous years) is desirable, or deeper understanding of the population is needed, such as looking for potential clusters among the entities.

\subsection{Probability Model Estimation}\label{sec:den_est}
The task of learning distributions is a probability model estimation problem in unsupervised settings~\autocite{lihang_2nd}. It sometimes takes the form of density estimation, which is considered by some statisticians as the most fundamental topic in probabilistic machine learning~\autocite{yujian_ml}. A basic and common technique, the histogram, can be easily misused which leads to biased understanding of the dataset (\ref{fig:hist_non_trivial}). 
\csmlongfigure[!htbp]{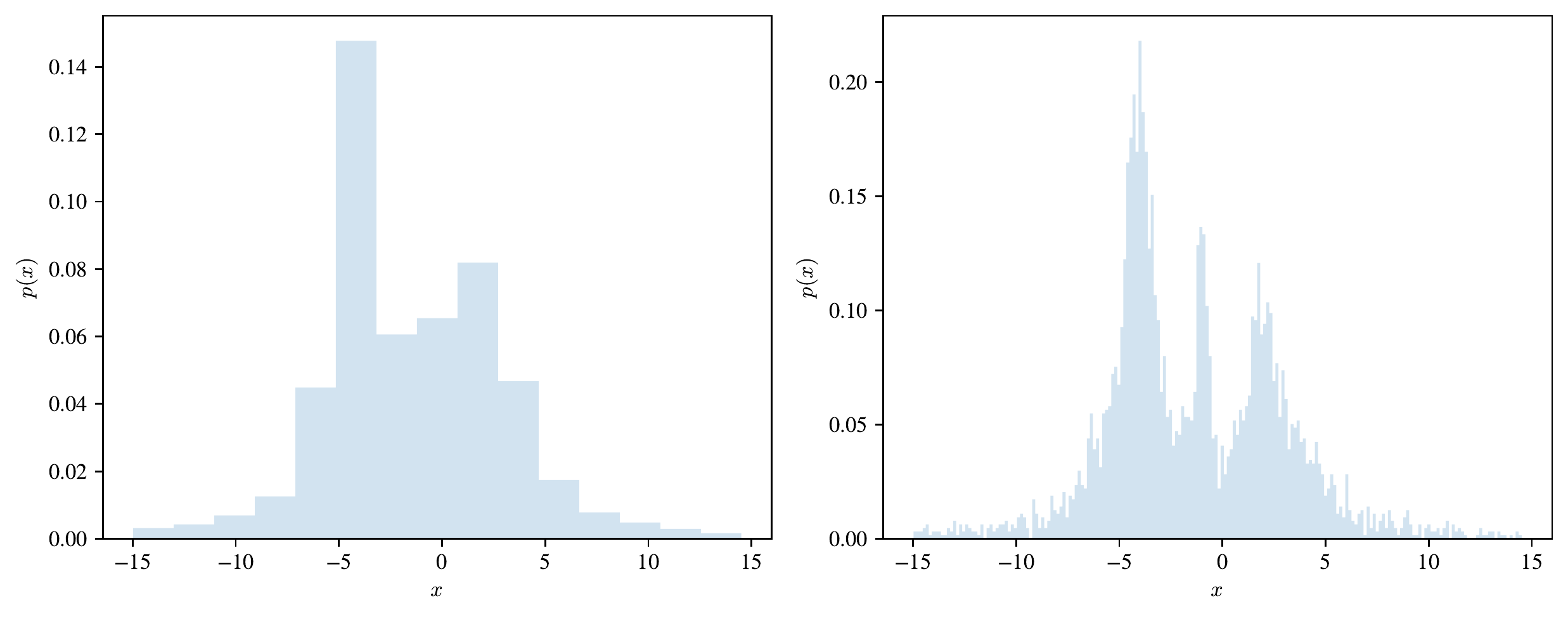}{figures/hist_non_trivial}{\textwidth}{Effective usage of histograms can be surprisingly subtle.}{ With the exact same dataset adapted from~\autocite{astroML}, the two histograms with different bin sizes demonstrate different multimodal features. Accepting some default configuration from some software package yields only one view of the distribution.}

In general, assuming that the data is generated by a probability model, the structure and parameters of that model are learned from the data. The type of the structure, i.e., the set of possible probability models is usually given (assumed), while the specifics of the structure and the parameters have to be learned. The goal is to find the model structure and the parameters which are most likely to have generated the data.

The probability model can be a mixture model or a graphical model. In this dissertation, the mixture model is considered, where the assumption is that data comes from a mixture of distributions. Mathematically, mixture models describe a distribution $p(x)$ by a convex combination of $K$ base distributions:
\begin{subequations}\label{mod:gen_mix_mod}
    \begin{align}
        p(x) &= \sum_{k=1}^{K}\pi_k p_k(x) \label{eq:gen_mix_mod_den} \\[0.5ex]
        \sum_{k=1}^{K}\pi_k &= 1, \; \pi_k \geq 0,
    \end{align}
\end{subequations}
where $p_k$ are the components in the mixture and $\pi_k$ are the mixture weights. Mixture models can be interpreted as the overall population being a combination of distinct subpopulations. Mixture models can be generalized to the continuous cases as well. For example, both the negative binomial distribution and Student's $t$-distribution can be thought of a mixture of some continuous distributions~\autocite{martin2018bayesian}.

In the model representation $P_{\bm{\theta}} (x \mid z)$, $x$ stands for the observations which can be discrete or continuous quantities; $z$ represents the latent structure which is a discrete random variable. The model is parameterized by $\bm{\theta}$. When the model is assumed to be a mixture type, $z$ represents the different components. The knowledge of the model structure and parameters are learned from the data $U = \left\{x_1, x_2, \dots, x_N \right\}$, where in this work $x_{i} \in \mathcal{X} \subseteq \mathds{R}^1$, $i=1,2,\dots,N$, is the observation for the $i$-th oilfield.

\subsection{Modeling VIIRS Detection Count}\label{sec:neg-bin}
In Section~\ref{sec:gp_cox_process}, methods are developed for analyzing the time series of VIIRS detection count for any given oilfield. This section tackles the problem of how to extract insights from any given month's flare detection count in North Dakota's oilfields. Specifically, by learning from each oilfield's detection count, the population of the oilfields is summarized, through which the state government can gain distributional insights.

Following the general form in Section~\ref{sec:den_est}, this problem becomes a special case that the latent structure $z$ does not exist, i.e., satisfying $P_{\bm{\theta}} (x \mid z) = P_{\bm{\theta}} (x)$, where $x$ represents the detection count. It is when estimating conditional probability distributions becomes estimating probability distributions, therefore, only estimating the parameters of $P_{\bm{\theta}} (x)$ is enough. Density estimation in classical statistics, for instance the Gaussian parameters estimation, is an example of such scenarios.

Since the count data is modeled, the author compares the four observation models below with many randomly chosen months' data:
\begin{enumerate}
    \item Poisson likelihood
    \item Negative binomial likelihood
    \item Zero-inflated Poisson (ZIP) likelihood
    \item Zero-inflated negative binomial (ZINB) likelihood
\end{enumerate}

Items 3 and 4 above are experimented with because many of the oilfields in North Dakota did not have detection records from VIIRS for a given month. Therefore, zero-inflated models are tried as well. Through the posterior predictive checks, it is found that the negative binomial observation model fits data in the most compatible manner, which is employed in this work.

The model is specified through \crefrange{nb_begin}{nb_end}:
\begin{subequations}\label{mod:of_ct_negbin}
    \begin{align}
        \mu &\sim \operatorname{Gamma}(2, 1) \label{nb_begin} \\[0.5ex]
        \phi &\sim \operatorname{Exponential}(1) \\[0.5ex]
        C_i &\sim \operatorname{NegBinomial}(\mu, \phi) \label{nb_end}
    \end{align}
\end{subequations}
where $C_i$ denotes the detection count for the $i$-th oilfield. The probability mass function of the negative binomial likelihood is parameterized by a location parameter $\mu \in \mathds{R}_{>0}$, and an overdispersion parameter $\phi \in \mathds{R}_{>0}$, in the following way:
\begin{equation}\label{eq:neg-bin}
P(X=n \mid \mu, \phi) = \frac{\Gamma(\phi+n)}{n! \, \Gamma(\phi)} \left( \frac{\mu}{\mu+\phi} \right)^{\!n} \left(\frac{\phi}{\mu+\phi} \right)^{\!\phi} \quad\text{for }n\in\mathds{N}_{0} \, ,
\end{equation}
where $\Gamma(\cdot)$ is the gamma function. Through this parameterization, the expectation and variance of a random variable $X \sim P$ are:
\begin{equation}
    \mathds{E}[X] = \mu \ \ \
    \text{ and } \ \ \ \mathds{V}[X] = \mu + \frac{\mu^2}{\phi} \, .
\end{equation}

As the negative binomial distribution describes a Poisson random variable whose rate parameter is gamma distributed, and due to the fact that $\operatorname{Poisson}(\mu)$ has variance $\mu$, the learned parameters provide nice interpretations for the state government:
\begin{itemize}
    \item $\mu$ indicates a mean intensity from the detection count's perspective, just like the interpretation of a Poisson's rate parameter. The larger the value of $\mu$, the more flare detections are present on average at an oilfield level.
    \item $\phi$ indicates the heterogeneity among the oilfields in North Dakota. Specifically, $\mu^2 / \phi$ is the additional variance above that of a Poisson with rate $\mu$. The smaller the value of $\phi$, the more oilfields with extreme detection counts (away from $\mu$) are present.
\end{itemize}

To demonstrate this model's compatibility with the observations, the data from October 2018 is used. There are \num{506} oilfields in total. The distribution of the detection count for all the oilfields is illustrated in \ref{fig:oilfield_ct_dist}.
\csmlongfigure[!htbp]{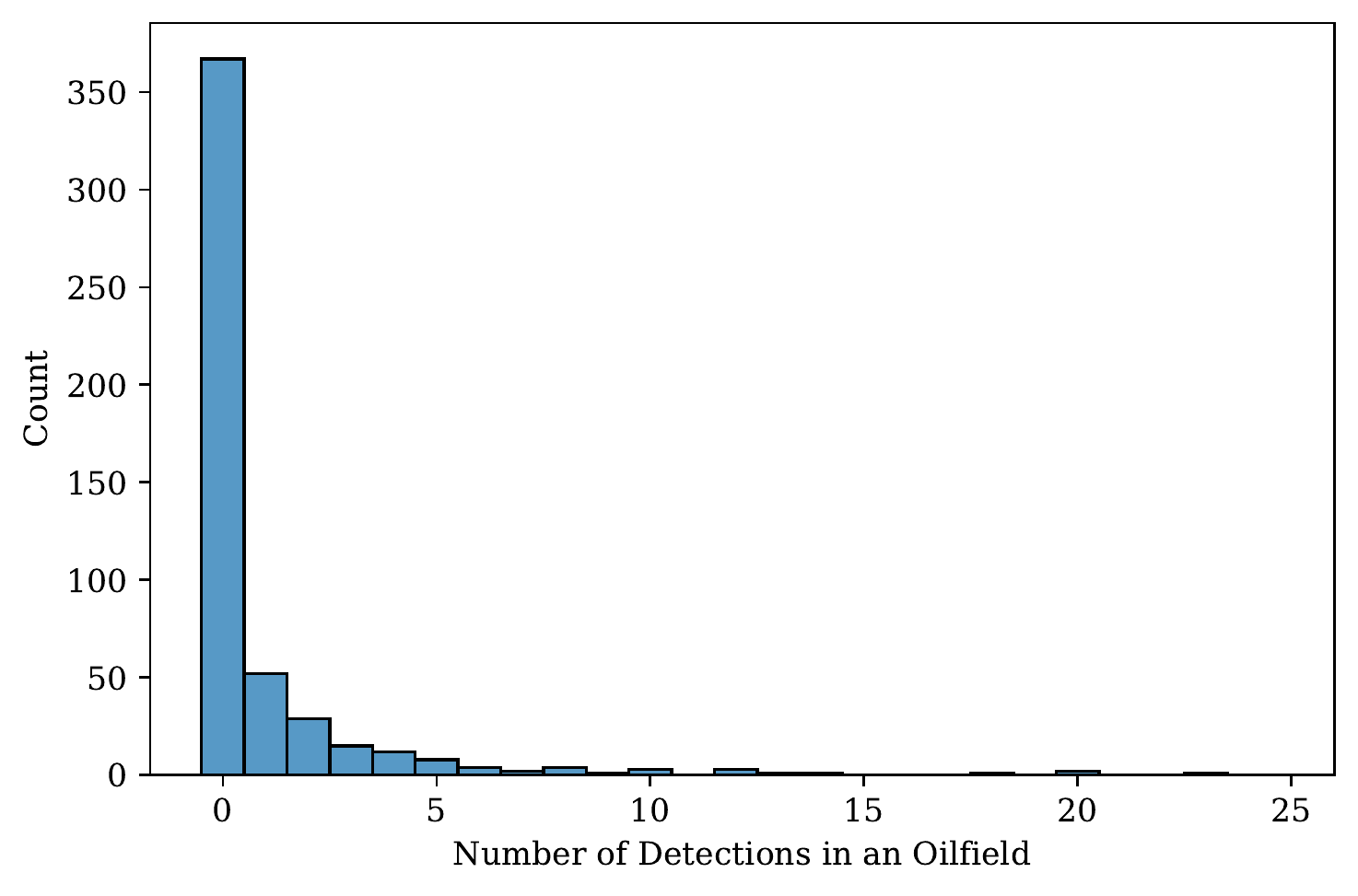}{figures/oilfield_ct_dist}{0.75\textwidth}{A histogram for the distribution of the oilfield detection counts from October 2018.}{ There are lots of zeros (more than \num{350}) and a few oilfields have relatively high detection counts (e.g., greater than or equal to \num{20}).}

After fitting Model~\ref{mod:of_ct_negbin}, the posterior distributions and trace plots of the hyperparameters are presented in \ref{fig:field_countNB_trace}. The parameter estimation results are reported in \ref{tab:of_ct_dist_param}.
\csmlongfigure[!htbp]{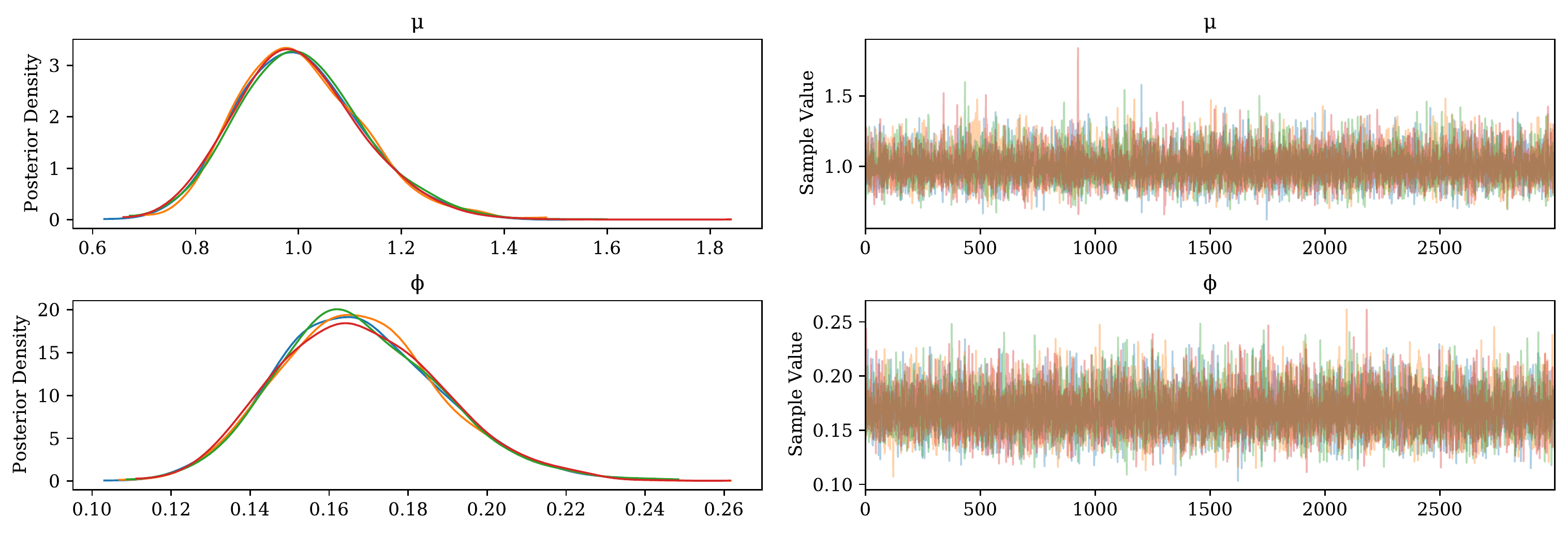}{figures/field_countNB_trace}{\textwidth}{Posterior distributions and trace plots for the oilfield detection counts distribution, fitted with the data from October 2018.}{ Well mixing and convergence have been achieved.}
\begin{table}[!htbp]
\centering
\caption{Parameter Estimates of Oilfield Detection Count Distribution}
\label{tab:of_ct_dist_param}
\begin{tabular}{c l S[table-format=1.3] c}
\toprule
Parameter & Variable & {Point Estimate} & \SI{90}{\percent} CI \\ \midrule
$\mu$ & Intensity & 1.005 & (0.814, 1.200) \\
$\phi$ & Heterogeneity & 0.168 & (0.135, 0.202) \\ \bottomrule
\end{tabular}
\end{table}

The point estimate for the intensity parameter $\mu$ is relatively small ($\hat{\mu}\approx1$), which possibly results from the model being overwhelmed by the large number of zero counts. However, by inspecting the histogram from \ref{fig:oilfield_ct_dist}, the tail of the distribution definitely extends far beyond $\hat{\mu}$. Therefore, posterior predictive checks are performed to scrutinize Model~\ref{mod:of_ct_negbin}'s compatibility with the observations. 

These types of checks substantially harness the information from the samples drawn from the posterior distributions. By combining the uncertainty about the parameters, as described by the posterior, with the uncertainty about the outcomes, as described by the likelihood, the generative model is employed to simulate the implied observations. Subsequently, posterior predictive plots are generated to display the model-based predictions along with the raw data. Such a plot for the detection count distribution model is given in \ref{fig:field_countNB_hist_all}.
\csmlongfigure[!htbp]{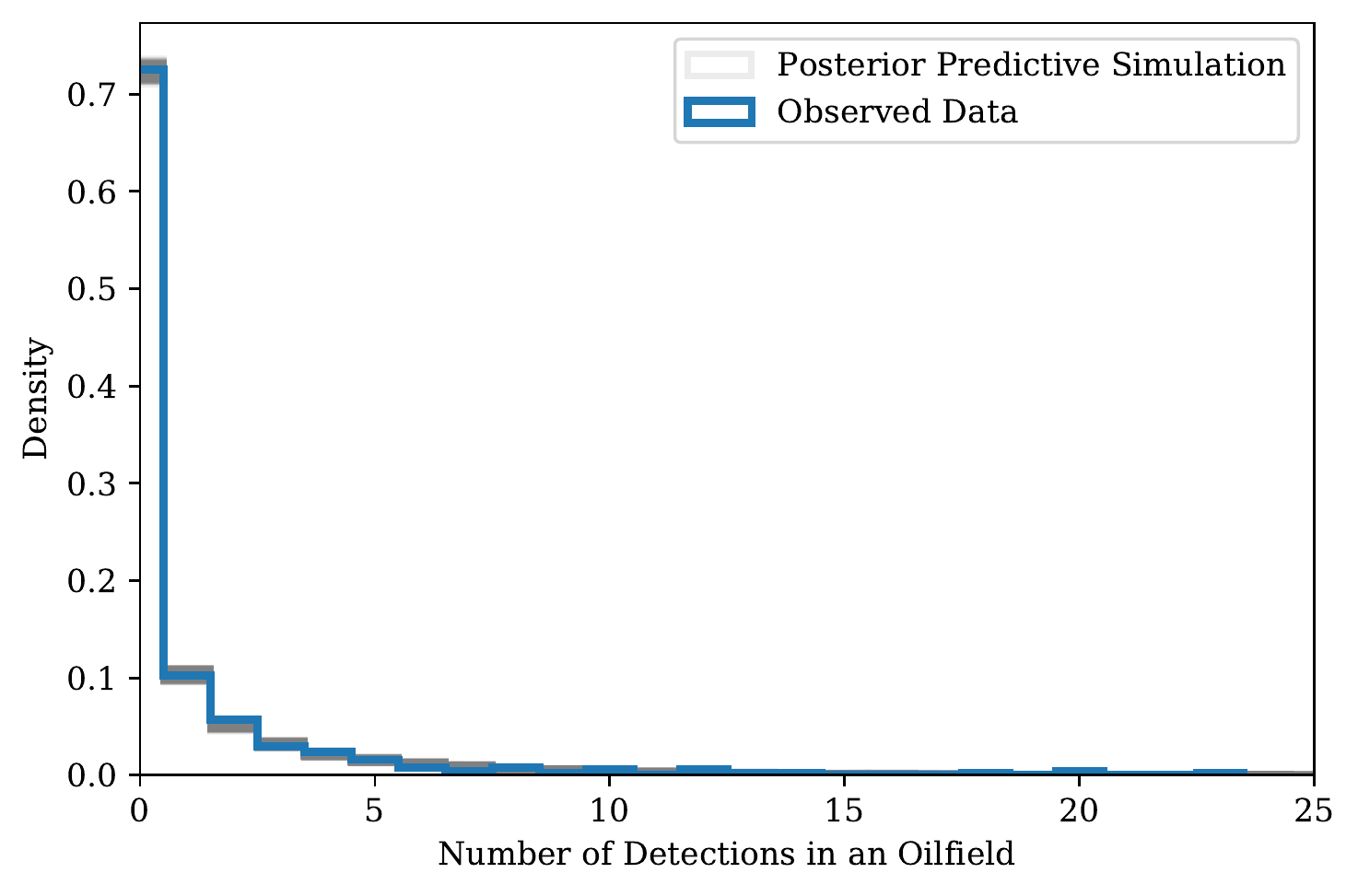}{figures/field_countNB_hist_all}{0.75\textwidth}{Histograms for the distribution of the oilfield detection counts from October 2018.}{ Blue: original data observed from VIIRS. Gray: posterior predictive simulation results obtained from Model~\ref{mod:of_ct_negbin}. Each set of the simulation results is plotted using gray with transparency via alpha blending (setting $\alpha=0.15$), such that the darker gray on the histograms indicates the simulated data which is more aligned with the model's expectation.}

In \ref{fig:field_countNB_hist_all}, the histograms for the original VIIRS observations, as well as all of the posterior predictive simulations are displayed. Each set of the parameter values (of $\mu$ and $\phi$) are used in simulating one synthetic snapshot of the oilfields in North Dakota for October 2018, and there are in total \num{12000} snapshots (constructed by the samples from the four Markov chains, each of which was setup for \num{3000} sampling iterations). Every histogram is visualized through an unfilled line chart, i.e., rendering the ``step'' histogram.

Through \ref{fig:field_countNB_hist_all}, it appears that the model is very compatible with the observations from October 2018, in that there is no obvious and consistent discrepancy between the observed and simulated data. To delve into the tail behaviors, i.e., beyond the zero count, a zoomed-in view is depicted in \ref{fig:field_countNB_hist_clip}. A few discrepancies are observed from this view, for example, when the count $C_i=11$ and $C_i=12$. One thing to note is that, with such a low mean ($\hat{\mu}\approx1$), even with a relatively large overdispersion ($\hat{\phi}\approx0.2$), the model would still be surprised by the high detection count, e.g., when $C_i\geq20$.
\csmlongfigure[!htbp]{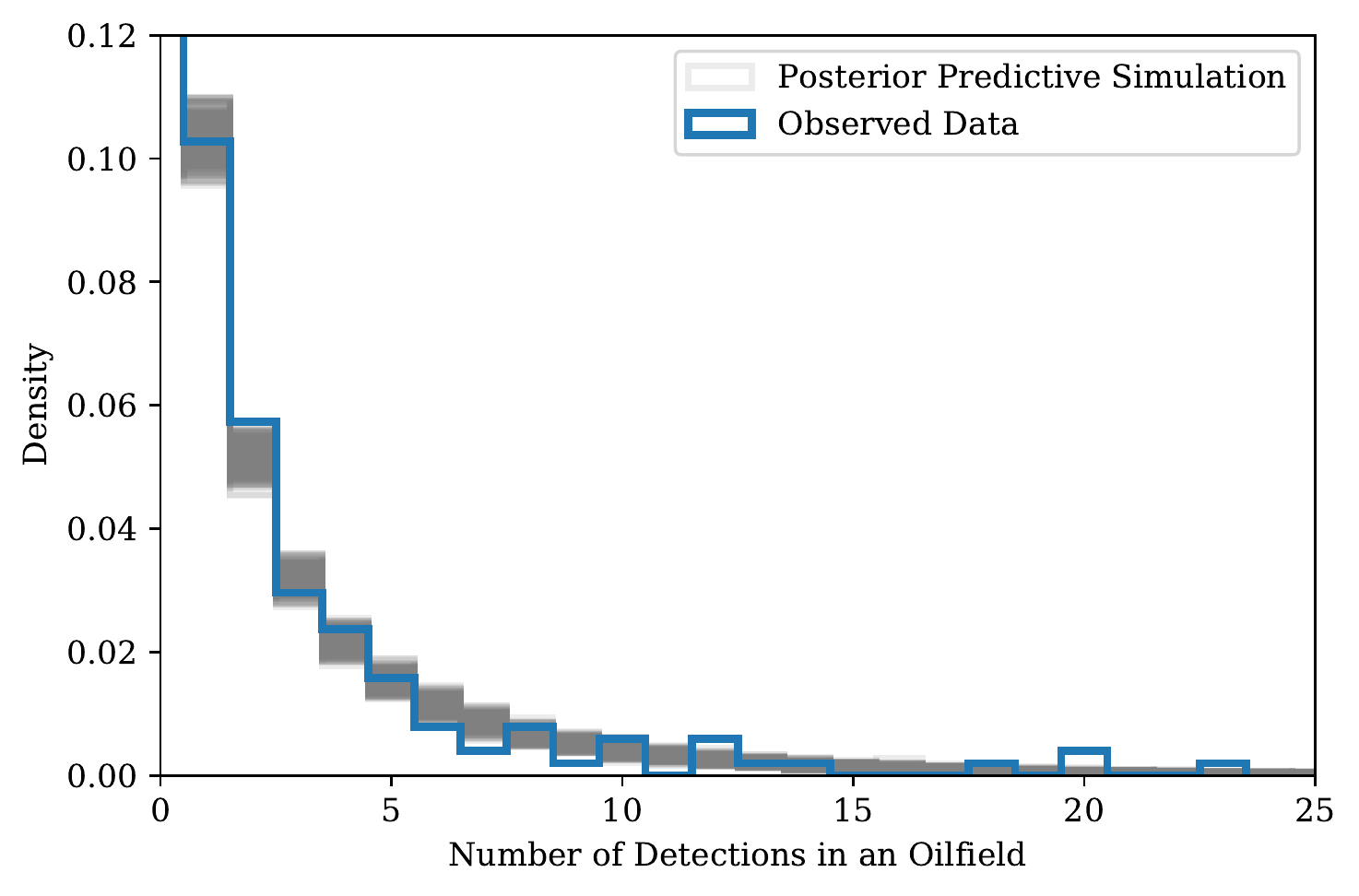}{figures/field_countNB_hist_clip}{0.75\textwidth}{Histograms for the distribution of the oilfield detection counts from October 2018, with the $y$-axis clipped to better present those counts which are greater than zero.}{ The legend with the associated color scheme is the same as in \ref{fig:field_countNB_hist_all}.}

The thorough performance of Model~\ref{mod:of_ct_negbin} that is characterized by a negative binomial likelihood, and the complicatedness of the real data manifest themselves through the posterior predictive checks. As discussed earlier in Section~\ref{sec:neg-bin}, the negative binomial likelihood was compared with three other likelihoods (Poisson, ZIP and ZINB) on many randomly chosen months, and found to outperform them in terms of the compatibility with the data in general. In fact, there are some months' data that are distributed in a ``cleaner'' way, i.e., almost perfectly described by Model~\ref{mod:of_ct_negbin}. The author chooses not to cherry-pick those data, in the hope of not misleading the readers about the performance of the developed model. 

Nevertheless, the simplicity, interpretability, and effectiveness of Model~\ref{mod:of_ct_negbin} proves itself in the mission of modeling detection count distribution. In practice, the state government can benefit from this model in the two use cases below:
\begin{enumerate}
    \item When the latest month's data becomes available, Model~\ref{mod:of_ct_negbin} can be fitted to obtain an estimate for $\mu$ and $\phi$. These parameter estimates along with the credible intervals can be compared with those from the earlier times. In the case of the discussions above, the learned parameters can be compared either with August/September from 2018, or with October from 2016/2017. From the comparison, it provides insights into whether there are more detection counts on average (characterized by a larger $\mu$), or if more oilfields with an atypical number of detections are spotted (characterized by a smaller $\phi$).
    \item After the model is fitted, it is recommended to perform the posterior predictive checks as demonstrated in \ref{fig:field_countNB_hist_all} and \ref{fig:field_countNB_hist_clip}, to identify any issues of the fits. The list of the oilfields which have large deviations from the simulated data, especially those on the far tail (e.g., when $C_i\geq20$), are worth tracking. That is, to investigate whether the ``anomalies'' from each month are random samples from the population or do not change from month to month. This provides further understanding of how the oilfields population behave, from the perspective of the detection count.
\end{enumerate}

A distributional summary of the detection counts exhibits only one facet of the flaring landscape, while the flared volumes distribution provides another crucial one, which is discussed next.

\subsection{Modeling Flared Volume}
In this section, the VIIRS estimated flared volumes for different oilfields are studied from a distributional point of view. The dataset from a three-month period is analyzed for demonstration purposes. Specifically, following the reverse geocoding as discussed in Section~\ref{sec:rev_geocode}, all the oilfields' cumulative flared volumes during Q4 2018 are computed and compiled for analysis.

There are in total \num{152} oilfields that have VIIRS reported volumes in this time span. The data is highly skewed (\ref{fig:bcm_hist_skew}). Therefore, for each oilfield, the order of magnitude of the flared volume (in \si{\bcm}) is computed for the analysis, instead of working with the original absolute volumes.
\csmlongfigure[!htbp]{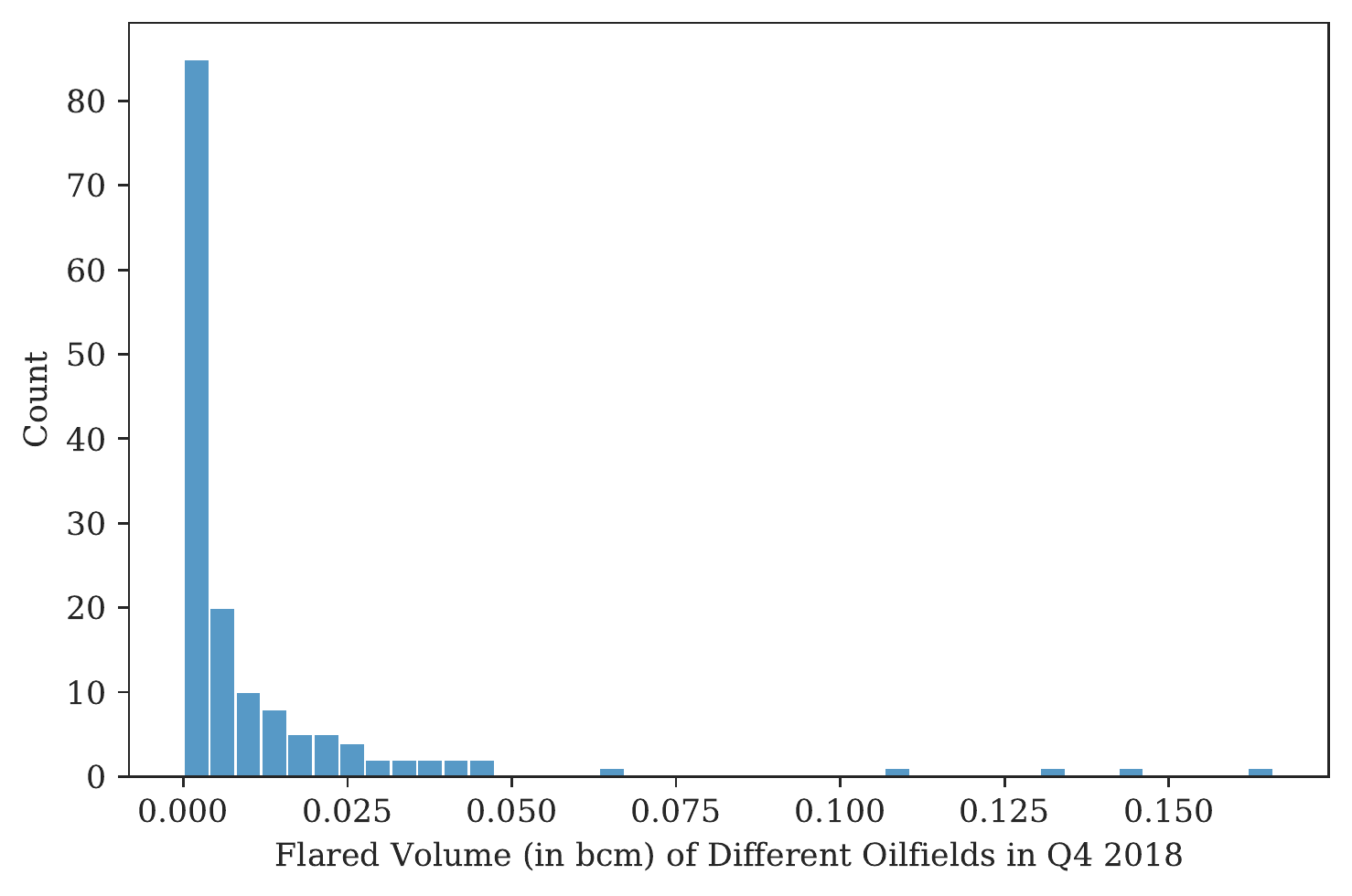}{figures/bcm_hist_skew}{0.75\textwidth}{Histogram for the distribution of the oilfield flared volumes from Q4 2018.}{ Most of the oilfields have relatively small flared volumes (e.g., less than \SI{0.01}{\bcm}), while a few oilfields have volumes that are greater than \SI{0.1}{\bcm}.}

From an applied perspective, taking the log of a measure converts the measure into magnitudes~\autocite{mcelreath2016rethink}, which is applied to each oilfield's flared volume:
\begin{equation}\label{eq:log_vol_def}
    L_i = \log(F_i),
\end{equation}
where $F_i$ is the original flared volume in \si{\bcm}, and $L_i$ is the flared volume magnitude, both of which are for the $i$-th oilfield. In this dissertation, base $e$ is always used for the logarithm (i.e., natural logarithm). A univariate distribution of the magnitudes is visualized in \ref{fig:logbcm_hist}.
\csmlongfigure[!htbp]{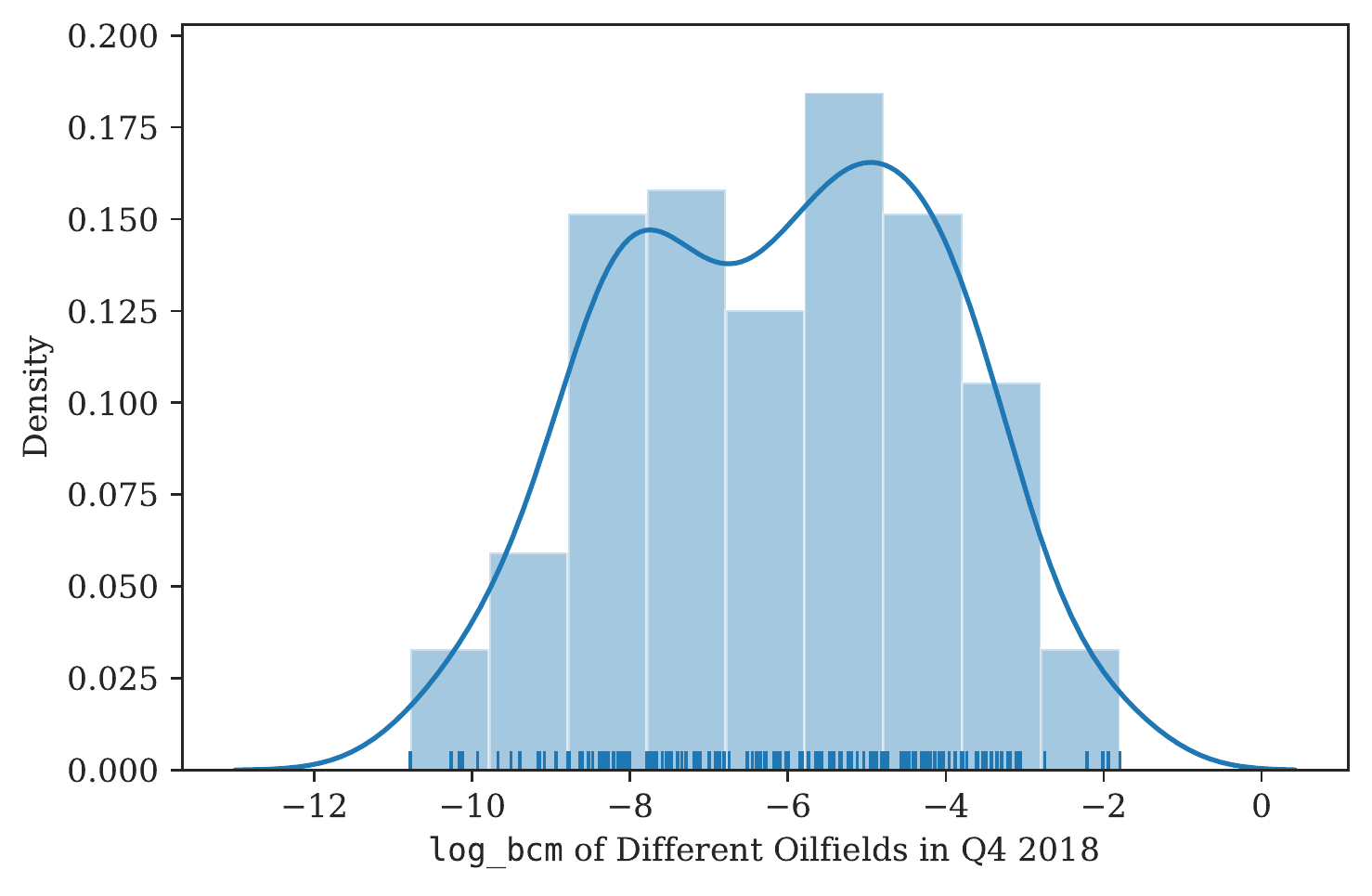}{figures/logbcm_hist}{0.75\textwidth}{Distribution of the oilfield flared volume magnitudes from Q4 2018.}{ The rug plot marks the value for each oilfield. The histogram is generated with nine bins. The curve displays a Gaussian kernel density estimate.}

Among the three approaches used to visualize the distribution, only the rug plot does not lead to subtleties due to the hyperparameters used. However, as a 1D scatter plot, its representation ability is naturally limited. The histogram suffers from the problem as illustrated in \ref{fig:hist_non_trivial}. The curve is generated by kernel density estimation (KDE). For a given dataset as defined in \Cref{eq:unsuper_data}, KDE represents the underlying distribution as:
\begin{equation}
\hat{p}(x) = {\frac {1}{Nh}}\sum_{i=1}^{N} K{\Big (}{\frac{x-x_{i}}{h}}{\Big )},
\end{equation}
where $K(\cdot)$ is a kernel function and $h$ is a bandwidth parameter. To generate \ref{fig:logbcm_hist}, the Gaussian kernel is used, which is given by:
\begin{equation}
    K(z) = \frac{1}{\sqrt{2\uppi}}\exp\biggl( -\frac{z^2}{2} \biggr),
\end{equation}
and $h$ is chosen based on Scott's rule. 

Since the bandwidth plays a similar role as the bin size in histograms, KDE can also lead to the same issue as in histograms.
Nevertheless, all three (the rug plot, histogram and KDE) agree that a single Gaussian approximation of the density which generates this data would be a poor approximation. Therefore, Gaussian mixture model (GMM) is employed to represent the data, i.e. the base distributions in Model~\ref{mod:gen_mix_mod} are chosen to be Gaussians. GMM provides more expressive modeling capabilities and also possibilities for clustering.

\subsubsection{Model Specification}
As discussed earlier, since the flared volume is a continuous quantity, density estimation is applicable and tackled with GMM. At first, the data generating process is considered, which paves the way for potential clustering applications. That is, each data point $L_i$ (defined in \Cref{eq:log_vol_def}) is assumed to be generated by exactly one mixture component. The number of components, $K$, is unknown, and up to seven components are tried to fit the dataset visualized in \ref{fig:logbcm_hist}. A relatively small number of components are experimented, because as the number of components increases, it becomes more difficult to interpret the modeling results. The model is specified through \crefrange{gmm_lat_begin}{gmm_lat_end}, $\forall K \in \{2,\dots,7\}$:
\begin{subequations}\label{mod:gmm_lat_var}
    \begin{alignat}{3}
        \bm{\alpha} &= (\alpha_1, \dots, \alpha_K) = 6 \cdot \mathds{1}_{K} \label{gmm_lat_begin} &&\\[0.5ex]
        \mathbf{p} &\sim \operatorname{Dirichlet}(\bm{\alpha}) &&\\[0.5ex]
        z_i &\sim \operatorname{Categorical}(\mathbf{p}) &&\\[0.5ex]
        l_1 &= \min\{L_1, \dots, L_n\} &&\\[0.5ex]
        l_2 &= \max\{L_1, \dots, L_n\} &&\\[0.5ex]
        \widetilde{\mu}_k &= l_1 + (k-1)\!\left(\frac{l_2-l_1}{K-1}\right), \quad &&k=1,\dots,K \\[0.5ex]
        \mu_k &\sim \mathcal{N}(\widetilde{\mu}_k, 2), \quad &&k=1,\dots,K \\[0.5ex]
        \sigma_k &\sim \operatorname{Half-Normal}(2), \quad &&k=1,\dots,K \\[0.5ex]
        L_i \mid (z_i=j) &\sim \mathcal{N}(\mu_j, \sigma_j) &&j\in\{1,\dots,K\} \label{gmm_lat_end}
    \end{alignat}
\end{subequations}
where:
\begin{description}[noitemsep,itemindent=-8em,leftmargin=6em]
    \item[\bm{\alpha}] is the vector of concentration parameters for the Dirichlet distribution, which is a multivariate generalization of the beta distribution;
    \item[\mathbf{p}] is the simplex of probabilities for the mixture components, which is assigned a Dirichlet prior. This prior with each value inside $\bm{\alpha}$ being \num{6}, is a weakly informative prior, expecting any $p_k$ inside $\mathbf{p}$ could be bigger or smaller than the others. Ten random draws from $\operatorname{Dirichlet}([6,6,6,6,6,6,6])$ are displayed in \ref{fig:dirichlet};
    \item[z_i] is the probable mixture component that the $i$-th oilfield belongs to;
    \item[l_1\text{ and }l_2] are the lower and upper bound for $\left\{ L_i \right\}_{i=1}^{n}$, respectively;
    \item[\widetilde{\mu}_k] is used in ``initializing'' the location of the $k$-th mixture component, and $\left\{\widetilde{\mu}_k\right\}_{k=1}^{K}$ essentially represent the $K$ evenly spaced points between $[l_1,l_2]$;
    \item[\mu_k] is the mean for the $k$-th Gaussian component;
    \item[\sigma_k] is the standard deviation for the $k$-th Gaussian component;
    \item[L_i] is the flared volume magnitude of the $i$-th oilfield, which is generated by the mixture component $z_i$.
\end{description}
\csmlongfigure[!htbp]{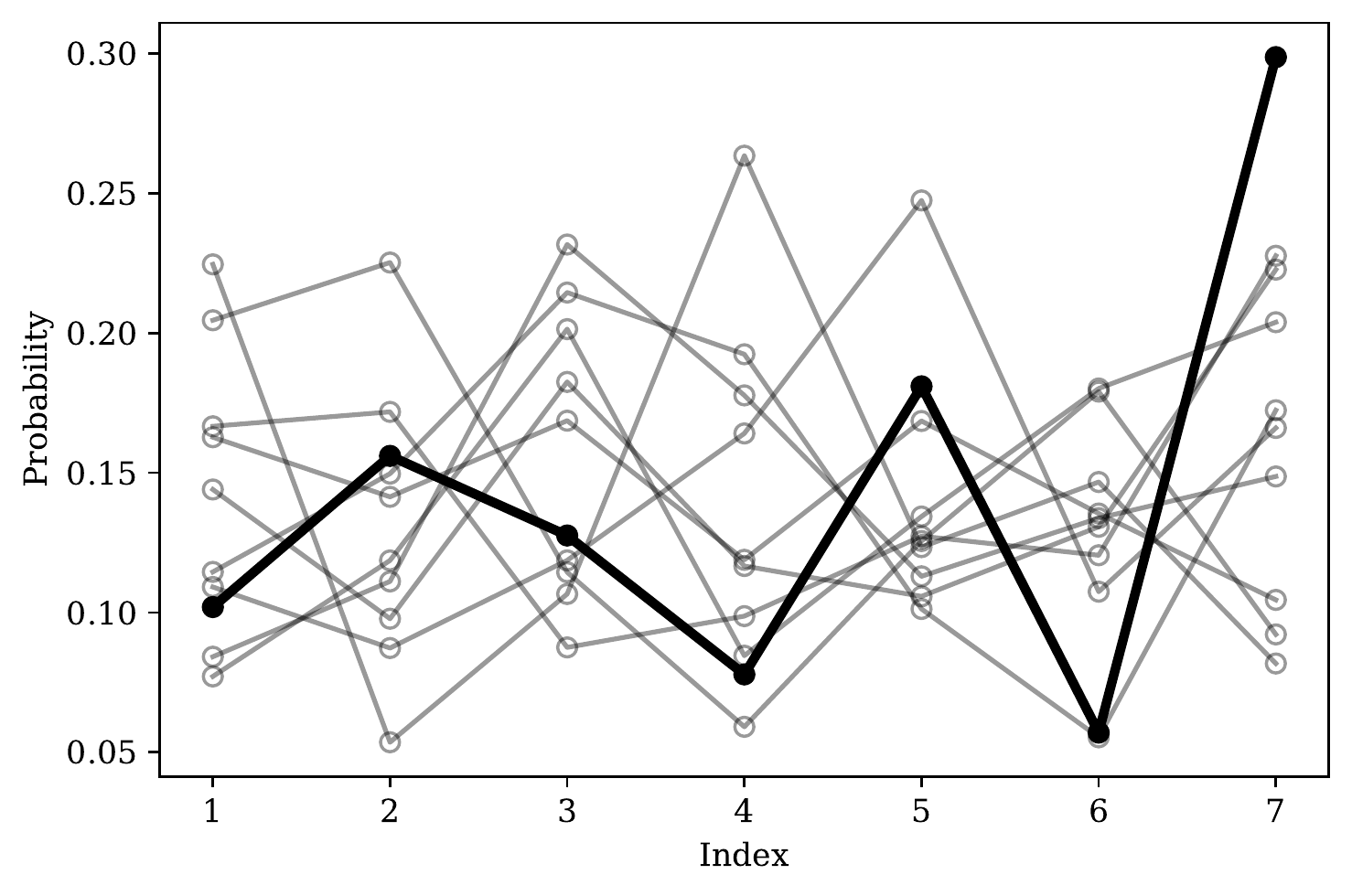}{figures/dirichlet}{0.6\textwidth}{Ten random draws from a Dirichlet prior with $\bm{\alpha}=(6,6,6,6,6,6,6)$.}{ One draw is highlighted to show that this prior is weak, in that it does not force all the probabilities (in any single draw) to be equal.}

Model~\ref{mod:gmm_lat_var}, while unambiguously expressing the assumed generative process, relies on sampling the discrete latent variables $z_n$, which is controlled by a categorical mixing distribution. This reliance causes slow mixing and ineffective exploration of the posterior distribution. An equivalent parameterization which addresses these problems is to marginalize out the $z$ parameter. The marginalized model is specified through \crefrange{gmm_marg_begin}{gmm_marg_end}, $\forall K \in \{2,\dots,7\}$:

\begin{subequations}\label{mod:gmm_marg}
    \begin{alignat}{3}
        \bm{\alpha} &= (\alpha_1, \dots, \alpha_K) = 6 \cdot \mathds{1}_{K} \label{gmm_marg_begin} &&\\[0.5ex]
        \mathbf{w} &\sim \operatorname{Dirichlet}(\bm{\alpha}) &&\\[0.5ex]
        l_1 &= \min\{L_1, \dots, L_n\} &&\\[0.5ex]
        l_2 &= \max\{L_1, \dots, L_n\} &&\\[0.5ex]
        \widetilde{\mu}_k &= l_1 + (k-1)\!\left(\frac{l_2-l_1}{K-1}\right), \quad &&k=1,\dots,K \\[0.5ex]
        \mu_k &\sim \mathcal{N}(\widetilde{\mu}_k, 2), \quad &&k=1,\dots,K \\[0.5ex]
        \sigma_k &\sim \operatorname{Half-Normal}(2), \quad &&k=1,\dots,K \\[0.5ex]
        L_i &\sim \sum_{j = 1}^K w_j \, \mathcal{N}(\mu_j, \sigma_j) &&\label{gmm_marg_end}
    \end{alignat}
\end{subequations}
where $\mathbf{w}$ are the mixture weights (i.e., mixing proportions), and the rest of the symbols have the same meaning as in Model~\ref{mod:gmm_lat_var}. The likelihood function, defined in \cref{gmm_marg_end}, corresponds with the density of a mixture model expressed in its general form (\Cref{eq:gen_mix_mod_den}).

Model~\ref{mod:gmm_marg} is implemented and fitted six times ($\forall K \in \{2,\dots,7\}$) to compare the inference results with different number of components specified. For each $K$, rapid mixing and fast convergence of the Markov chains are obtained. The modeling results are displayed in \ref{fig:logbcm_fit}, where the KDE (same as in \ref{fig:logbcm_hist}) and the Gaussian components inferred are plotted along with the posterior samples.
\csmlongfigure[!htbp]{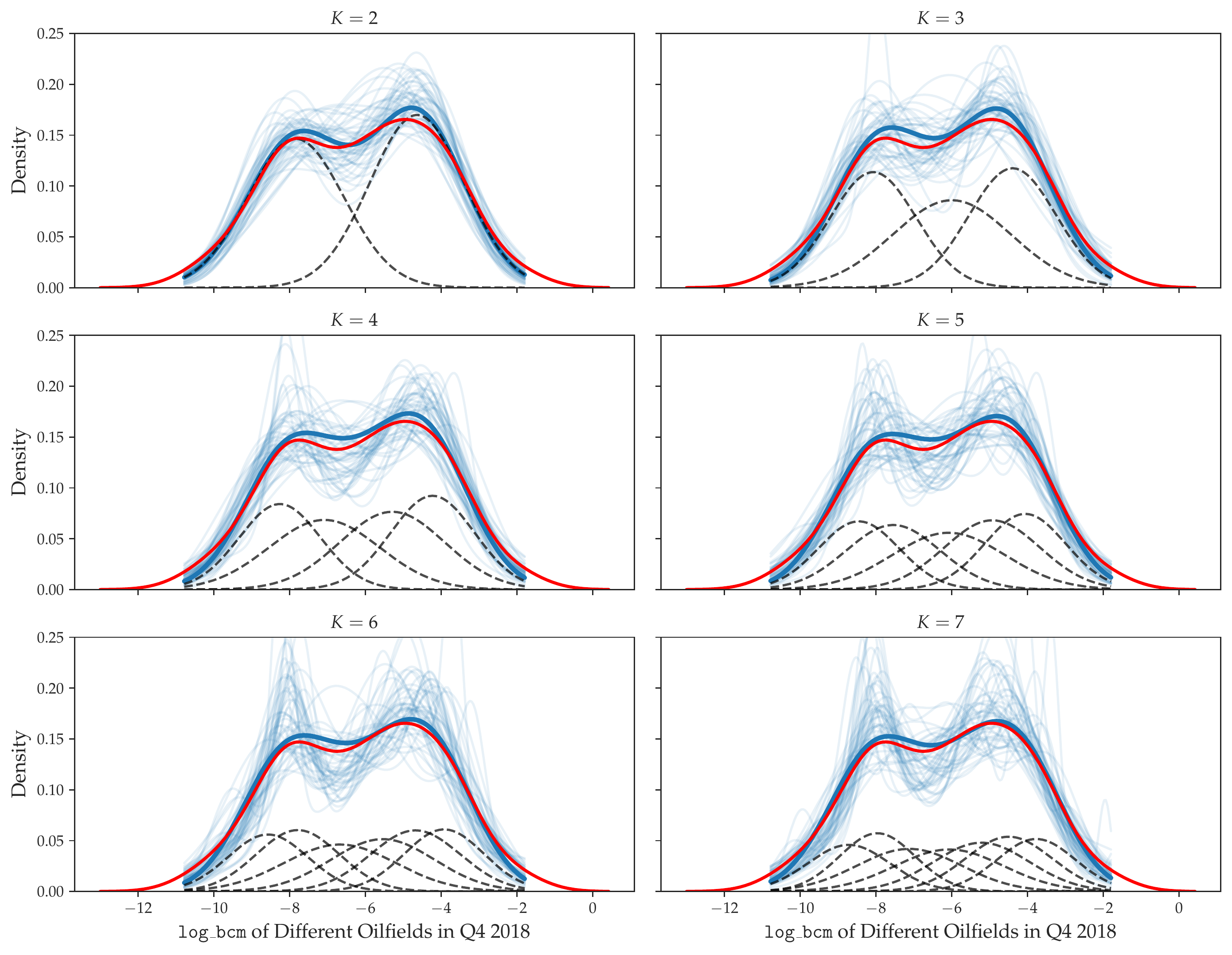}{figures/logbcm_fit}{\textwidth}{GMM inference results with different $K$'s.}{ The thick blue line denotes the posterior mean fit of the underlying density. The light blue lines show \num{50} random samples from the posterior. The dashed lines represent the posterior mean Gaussian components. The red curve shows the fit using KDE.}

It can be observed that, when using a mixture of Gaussians, the multimodal features can be represented in a relative effortlessly way, and all the mean fits are quite close to the one obtained with KDE. As the number of components increases, for example when $K=6$ or $K=7$, the mean density estimation using GMM resembles KDE more closely, but the samples from the posterior show more stochasticity, which is an indicator of potential overfitting. This naturally leads to the question of how to decide the number of components for this dataset.

\subsubsection{Model Comparison}
Choosing the best $K$ is a model comparison problem, for which there does not exist a silver bullet. In this dissertation, the author chooses to take the information criteria approach, specifically leveraging the widely applicable information criterion (WAIC) introduced by \textcite{waic}. Information criteria provide a theoretical estimate of the relative out-of-sample KL divergence~\autocite{mcelreath2020rethink}, and thus a lower value is better. Following \textcite{martin2018bayesian} and \textcite{mcelreath2020rethink}, WAIC is computed by:
\begin{subequations}
    \begin{align}
        \operatorname{WAIC}(\mathbf{y}, \mathbf{\Theta}) &= -2 \times \operatorname{lppd}(\mathbf{y}, \mathbf{\Theta}) + 2 p_{\mathsmaller{\textsc{waic}}} \\[0.5ex]
        &= -2 \sum_{i=1}^{n} \log\Biggl( \frac{1}{S}\sum_{j=1}^{S} p(y_i \mid \Theta_{j}) \Biggr) + 2 \sum_{i=1}^{n} \mathds{V}_{\!\bm{\Theta}}[\log p(y_i \mid \Theta_{j})] \, ,
    \end{align}
\end{subequations}
where:
\begin{description}[noitemsep,itemindent=-8em,leftmargin=6em]
    \item[\mathbf{y}] denotes the observations and $y_i$ is the $i$-th observation;
    \item[\mathbf{\Theta}] is the posterior distribution and $\Theta_{j}$ is the $j$-th set of sampled parameter values;
    \item[S] is the number of posterior samples;
    \item[\operatorname{lppd}(\cdot)] calculates the log pointwise predictive density;
    \item[p_{\mathsmaller{\textsc{waic}}}] is the penalty term given by summing up the variance in the log-likelihood over the $S$ posterior samples, for each observation $i$.
\end{description}

Fundamentally, model comparison is performed by leveraging Occam's razor, i.e., parsimonious models are preferred in light of predictive performance. The models are compared based on their WAIC values, which are summarized using \ref{fig:logbcm_waic}.
\csmlongfigure[!htbp]{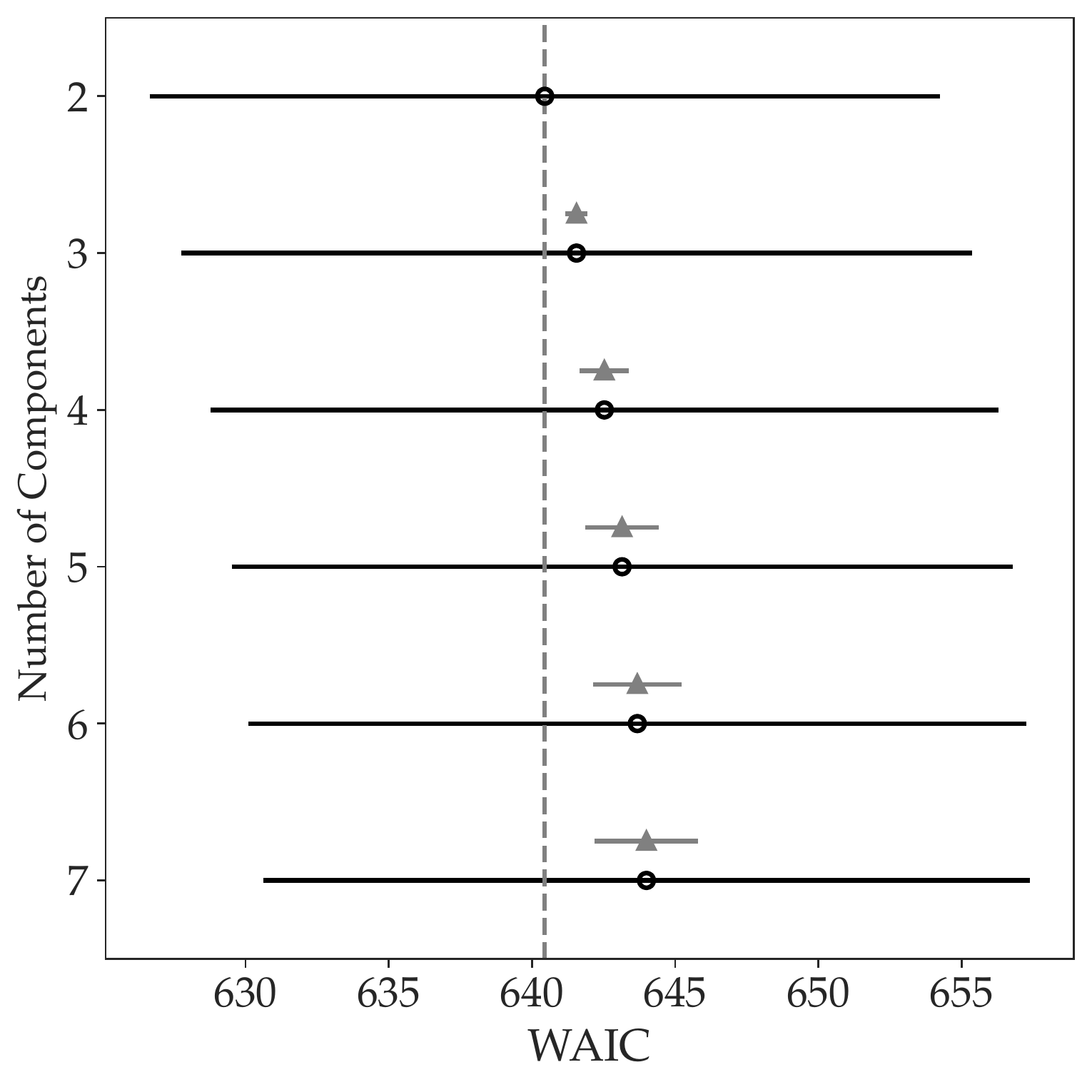}{figures/logbcm_waic}{0.5\textwidth}{WAIC values with different $K$'s.}{ The open points denote the WAIC values. The long horizontal line segments represent the standard error for each WAIC. Standard error of the difference in WAIC (between each model and the top-ranked one) is shown by the lighter line segment with the triangle on it.}

It can be seen that the model with two Gaussian components are the best (smallest WAIC), however, there are considerable overlaps among all of the models when the estimated standard error is taken into consideration. Considering the fact that $K=2$ gives the simplest model, also that there are only \num{152} observations (oilfields) in this dataset, the GMM with two components would be the best choice.

\subsubsection{Clustering}
When looking at the developed model from a latent variable perspective (Model~\ref{mod:gmm_lat_var}), it becomes obvious that the mixture model serves as a natural candidate for solving clustering tasks, in that every observation ($L_i$) can be drawn from one of the $K$ data generating processes, each with its own set of parameters, $\mathcal{N}(L_i \mid \mu_k, \sigma_k)$. Since a probabilistic model is built, for the purpose of clustering, a reasonable choice is to assign a data point to the mixture component (i.e., cluster) with the highest posterior probabilities (which are also interpreted as the responsibilities). In the case of the 2-component GMM trained from the previous sections, for a particular observation $x$, the probability that it belongs to cluster one ($z=1$) can be computed using Bayes' theorem (\Cref{eq:bayes_thm}):
\begin{equation}
    p(z=1 \mid x) = \frac{p(z=1) \, \mathcal{N}(x \mid \mu_1, \sigma_1)}{p(z=1) \, \mathcal{N}(x \mid \mu_1, \sigma_1) + p(z=2) \, \mathcal{N}(x \mid \mu_2, \sigma_2)} \, ,
\end{equation}
where every part in the formula can be obtained from the posterior samples (e.g., using the posterior means).

Clustering, as an unsupervised approach, can be used to reveal the hidden groups in the observations. In the case of the oilfield flaring magnitudes data in this chapter, the two clusters can be directly mapped to concepts such as major and minor flaring fields. However, it is usually the deeper insights into what caused these clusters that the state government is mostly interested in, for the sake of decision- and policy-making for example. If the oilfields belonging to the major flaring cluster seem to be a volatile membership when more months/quarters data are analyzed, the variations in flared volumes are possibly tied more closely to company strategies and movements. On the other hand, if there exists a group of oilfields that are found to join the major flaring cluster on a regular basis, this could provide a perspective in regards to where to construct the next natural gas processing plants, i.e., the locations/capacities of the new gas plants should be optimized based on those oilfields' situations.

In this chapter, the dataset compiled for unsupervised learning is univariate, i.e., $x_{i} \in \mathcal{X} \subseteq \mathds{R}^1$. GMM are also suitable for the density estimation and clustering tasks when the data goes beyond 1D. As an example, for the same oilfields studied for Q4 2018, if their oil production volumes are extracted from NDIC, a scatterplot of gas flaring versus oil production magnitudes is shown in \ref{fig:oil_gas_2d}. It is very possible that the density of the underlying distribution can be modeled by a bivariate normal distribution or a 2D GMM. In such cases, the mixture components become multivariate normal distributions, and the component covariance matrices can be constructed with the help of the LKJ distribution (which is employed in Models~\ref{mod:county_hier_cp} and \ref{mod:county_hier_ncp}). The developed density model can be used, for example, in anomaly detections, looking for any oilfields which have a tendency to creep toward the upper left corner (characterized by very little oil production and a huge flaring magnitude). Similar to all the inferences presented throughout this dissertation, one advantage of doing such is that the decision making can be based on some consistent metrics (such as probability scores), instead of some criteria based on human eyeballing or improvising.
\csmlongfigure[!htbp]{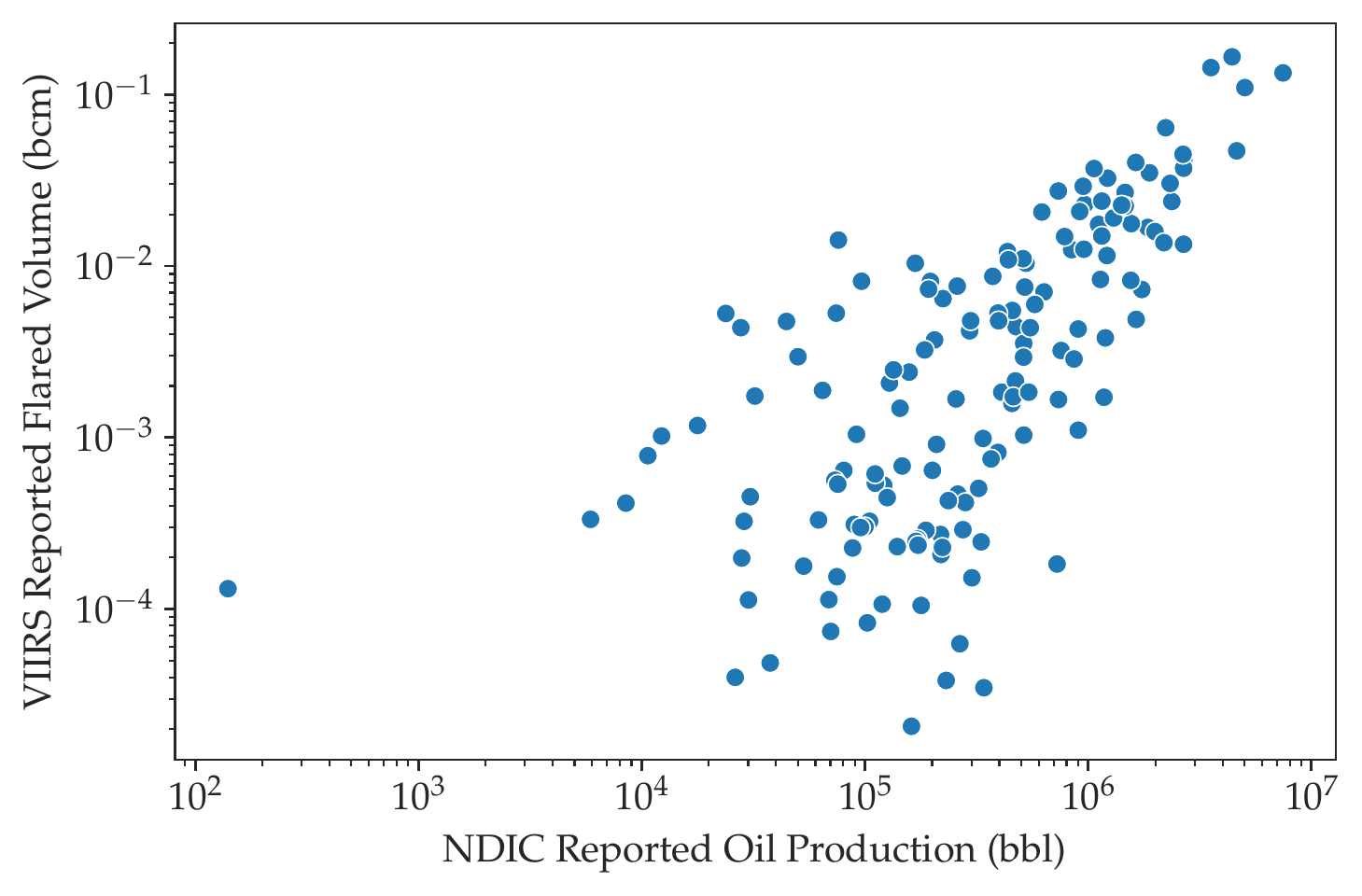}{figures/oil_gas_2d}{0.75\textwidth}{A scatterplot of oil production and flared gas volumes for different oilfields in Q4 2018.}{ Both the $x$- and $y$-axis are in log scale, showing the relationship between the magnitudes.}

This concludes the statistical modeling journey of this dissertation. In the next chapter, discussions are presented on one extension scenario and one bigger picture viewpoint, from applying Bayesian learning to flaring data.

\chapter{Discussion}\label{ch:discuss}
This chapter discusses the possibility of operator level monitoring and analytics, potential result inconsistencies, and relates the endeavors of learning from flaring data to the larger process of applying machine learning in the petroleum engineering domain.

\subsection{Operator Level Monitoring and Analytics}
Up till this point, the satellite-detected flaring statistics have been applied to the state, county, and oilfield levels. This is made possible by the reverse geocoding discussed in Section~\ref{sec:rev_geocode}. An ideal application scenario is operator level monitoring and analytics by leveraging the information from the satellite detections.

Unfortunately, assigning flares to corresponding companies is not a straightforward operation. One possible solution is to make use of the shapefiles of the leases, which are not provided by NDIC. Some data vendors have such files in their database. However, after spending some effort investigating the lease shapefiles from one vendor, the author believes it is possible to create more problems than solving the existing ones, when bringing in such information. In particular, some reasons include:
\begin{itemize}
    \item Multiple companies exist on a single lease.
    \item The company names from the lease shapefiles do not always correspond with those on the NDIC monthly production reports.
    \item Some leases in the vendor's database miss start date or end date data.
    \item It takes time for the vendor to compile and digitize such information, which makes the available lease shapefiles not up to date.
\end{itemize}

Nevertheless, for such an important use case, the author managed to develop a nearest-neighbor-based approach which partly solves the problem (Algorithm~\ref{alg:nn_op}). The essence of this approach is to cautiously assign the closest well's operator to each satellite-detected flare. The closest wells are found based on the corresponding time window. For example, for the flares detected in January 2016, only the active wells reported on the NDIC production report from the same month are looked up. The function \texttt{FindClosestOperator()} returns the closest operator ($\mathrm{OP}_j$) for each VIIRS detection, as well as the calculated distance ($d_j$) between each pair (of flare and well). The distance is calculated based on the haversine metric, i.e., the great-circle distance, thus the Earth radius ($R_E$) is needed. The function is essentially performing the $k$-nearest-neighbors ($k$-NN) search for $k=1$. When the sample is as large as in this case, i.e., there are usually a few hundred VIIRS detections and more than \num{15000} wells for each month, linear scanning each well's location for each VIIRS detection is too slow. Therefore, in this work, the function internally depends on a ball tree implementation from scikit-learn~\autocite{sklearn} for speedup on the $k$-NN search.

Once the $2$-tuple, $(\mathrm{OP}_j, d_j)$, is obtained for each VIIRS detection, some logics are implemented to decide whether to drop or keep the operator assignment. The idea is straightforward: the assignment is immediately kept or discarded, when $d_j$ is very small or very large, respectively. If $d_j$ is mid-range, i.e., $d_{\textsc{secure}} \leq d_j \leq d_{\textsc{cutoff}}$, the assignment will be in effect, only if the flare and the operator are found to be located on the same township/range/section. The township/range/section shapefiles, as part of the input for Algorithm~\ref{alg:nn_op}, are available from the NDIC GIS Map Server. The reverse geocoding follows the exact same procedure as in Section~\ref{sec:rev_geocode}. After the processing is completed, a small portion of the VIIRS detections are not used for operator level analytics, because either they are too far away from the reported well locations, or the townships/ranges/sections fail to match. It should be noted that, the pseudocode for Algorithm~\ref{alg:nn_op} is written in a way that illustrates the precise details in the data processing logics. For the implementation in this work, some of the for-loops are replaced by the vectorized operations for enhanced performance.

\begin{algorithm}[!htbp]
    \DontPrintSemicolon
    \caption{\label{alg:nn_op}Nearest-Neighbor-Based Flare Owner Assignment}
    \KwIn{both VIIRS and NDIC reportings in WGS 84 coordinates, the township/range/section shapefiles for North Dakota, $d_{\textsc{secure}}$, $d_{\textsc{cutoff}}$, $R_E$}
    \KwOut{operators being assigned to most VIIRS detections}
    \SetKwFunction{FindClosestOperator}{FindClosestOperator}
    \BlankLine
    $n \gets$ number of months \;
    \For{$i \gets 1$ \KwTo $n$}{
        $\mathrm{VIIRS}_i \gets$ the $i$-th month's observations from VIIRS \;
        $\mathrm{NDIC}_i \gets$ the $i$-th month's reportings from NDIC \;
        $(\mathrm{OP}, d) \gets$ \FindClosestOperator{$\mathrm{VIIRS}_i$, $\mathrm{NDIC}_i$, $R_E$} \;
        \BlankLine
        $m \gets$ number of records in $\mathrm{OP}$ or $d$ \;
        \For{$j \gets 1$ \KwTo $m$}{
            $\mathrm{OP}_j \gets$ the closest operator found on the $j$-th record \;
            $d_j \gets$ the distance between the flare and the closest well, for the $j$-th record \;
            \BlankLine
            \uIf{$d_j > d_{\textsc{cutoff}}$}{
                drop $\mathrm{OP}_j$ \;
            }
            \uElseIf{$d_j < d_{\textsc{secure}}$}{
                keep $\mathrm{OP}_j$ \;
            }
            \Else{
                \eIf{township/range/section agree}{
                    keep $\mathrm{OP}_j$ \;
                }{
                    drop $\mathrm{OP}_j$ \;
                }
            }
        }
    }
\end{algorithm}

The developed approach is tested with real flaring data from North Dakota. For the demonstrated cases in this section, the values below are chosen for Algorithm~\ref{alg:nn_op}:
\begin{subequations}
    \begin{align}
        d_{\textsc{secure}} &= \SI{300}{\metre} \\[0.5ex]
        d_{\textsc{cutoff}} &= \SI{800}{\metre} \\[0.5ex]
        R_E &= \SI{6371}{\kilo\metre}
    \end{align}
\end{subequations}

\begin{figure}[!htbp]
	\begin{center}
		\subfigure[Operator B]{
			\includegraphics[width=0.47\textwidth]{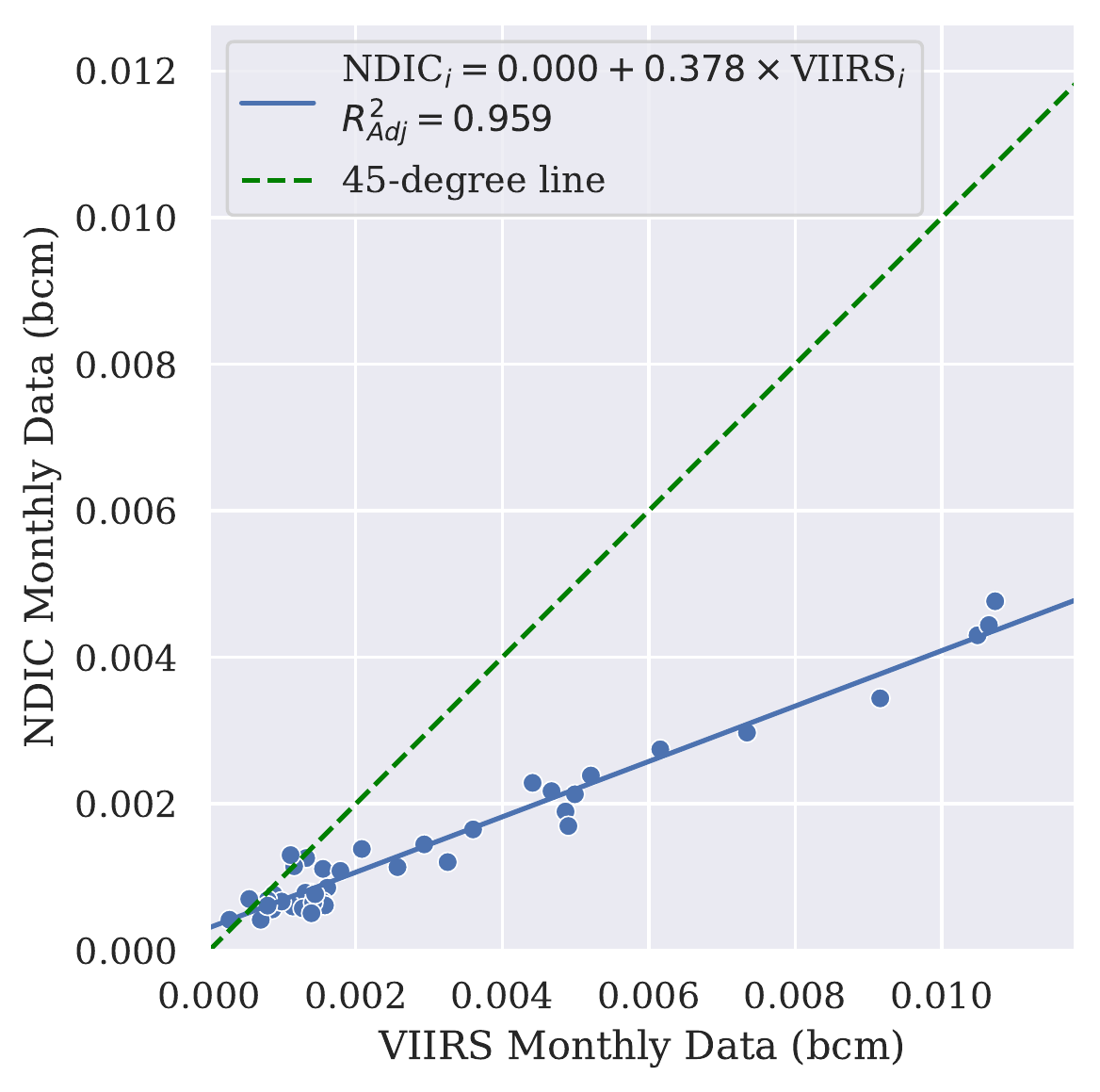}
		}\hfill
		\subfigure[Operator C]{
			\includegraphics[width=0.49\textwidth]{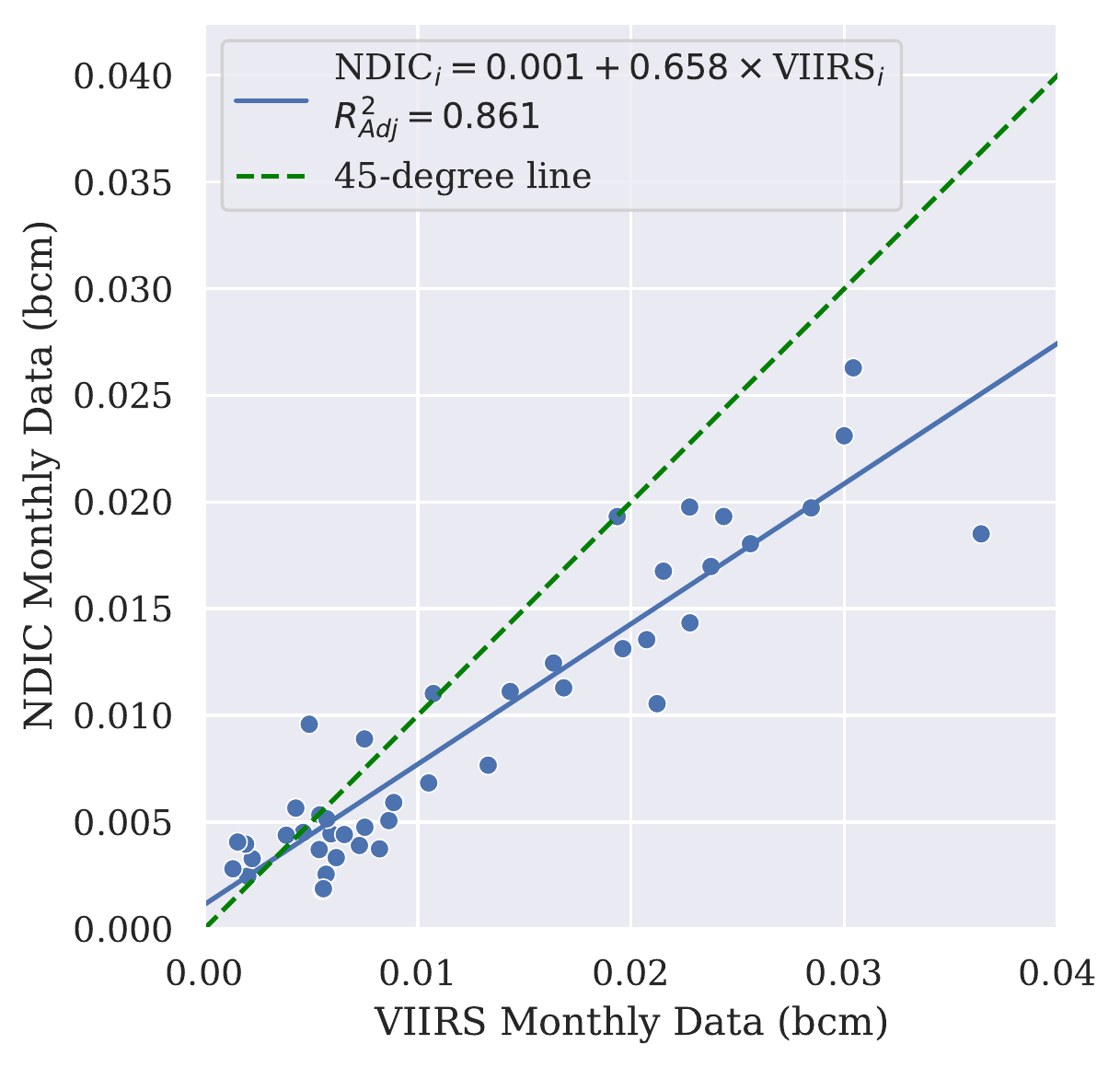}
		}
		\caption{\label{fig:good_match_op}Examples of good fits between the NDIC and VIIRS reported volumes, at the operator level.}
	\end{center}
\end{figure}

Some operators are found to show positive correlations between the NDIC and VIIRS reported volumes. Examples of two operators, denoted by Operator B and Operator C, are shown in \ref{fig:good_match_op}. The axes' meanings are the same as in the right panel of \ref{fig:eda_state_data_vis}. The legend shows the results of fitting \Cref{eq:state_lin_mod} by ordinary least squares (OLS). $R_{\text{adj}}^{2}$ stands for the adjusted $R^2$. Although the differences in $\hat{\beta}_{\mathsmaller{\textsc{operator}}}$ indicate that there is heterogeneity among the different companies, these operators show some consistency in terms of their own reporting and have good matches with the VIIRS data up to a scale factor (as the intercepts are very close to zero).

\begin{figure}[!htbp]
	\begin{center}
		\subfigure[Operator D]{
			\includegraphics[width=0.495\textwidth]{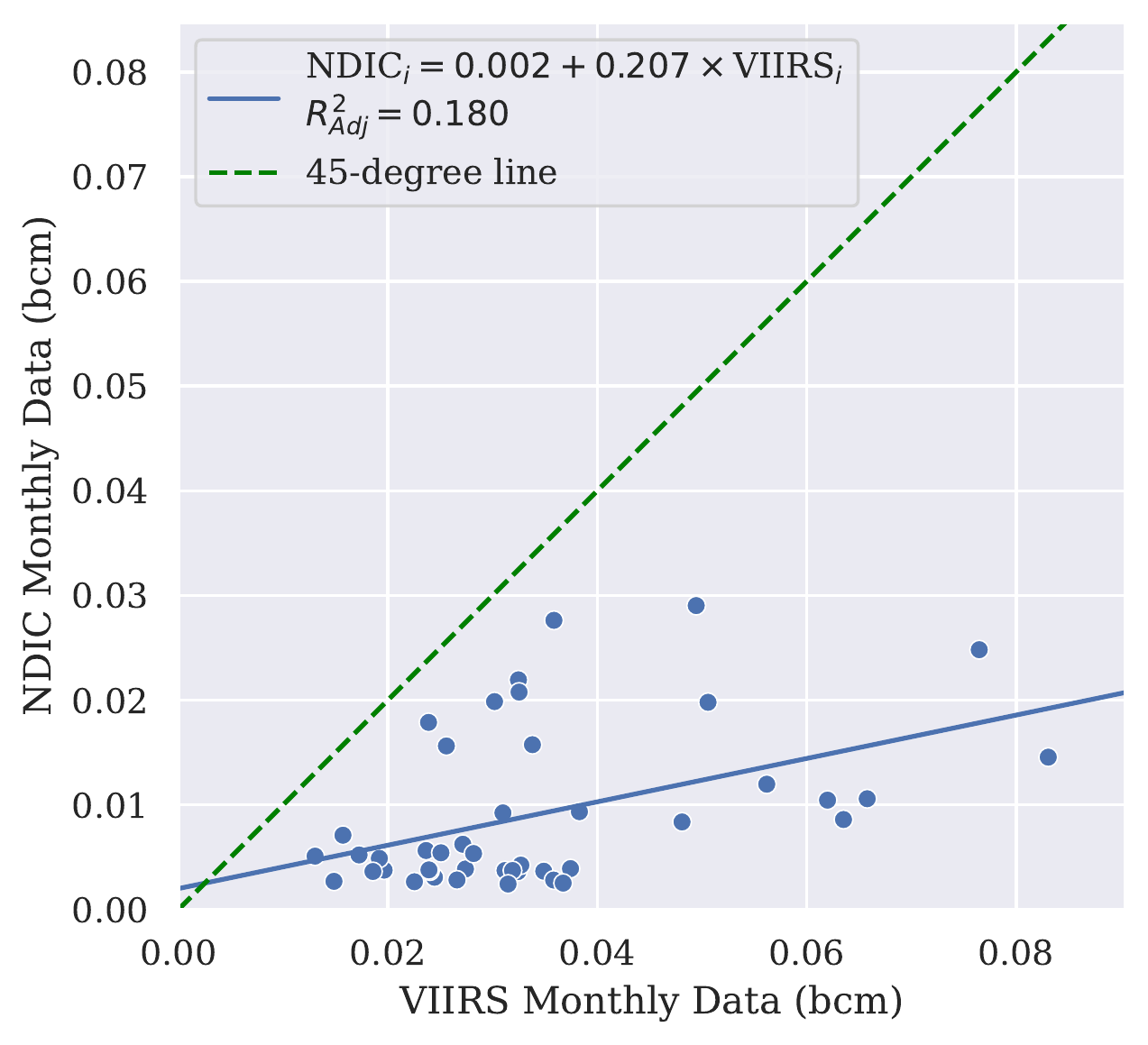}
		}\hfill
		\subfigure[Operator E]{
			\includegraphics[width=0.47\textwidth]{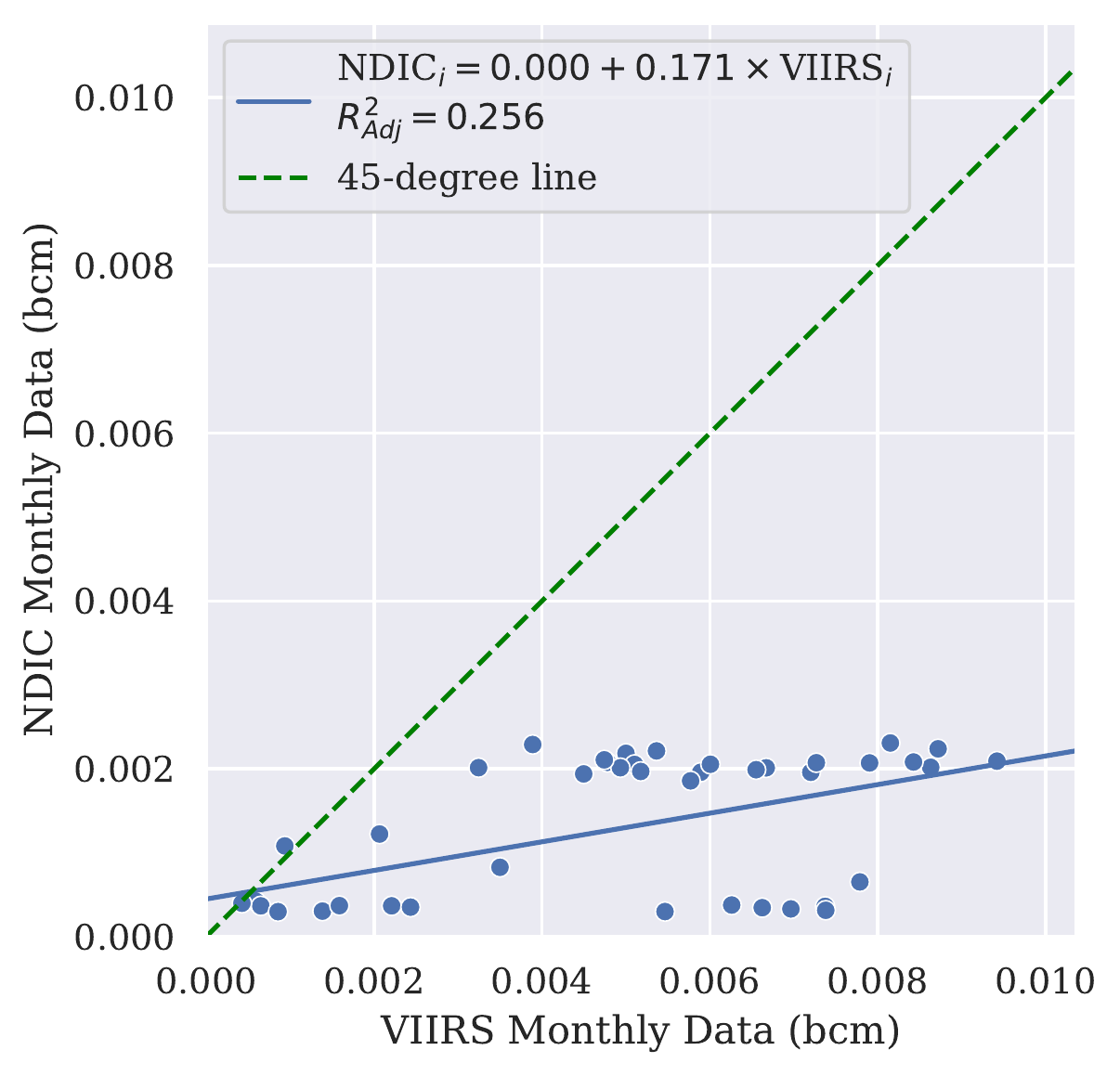}
		}
		\caption{\label{fig:bad_match_op}Examples of poor fits between the NDIC and VIIRS reported volumes, at the operator level.}
	\end{center}
\end{figure}

\begin{figure}[!htbp]
	\begin{center}
		\subfigure[Operator D]{
			\includegraphics[width=0.98\textwidth]{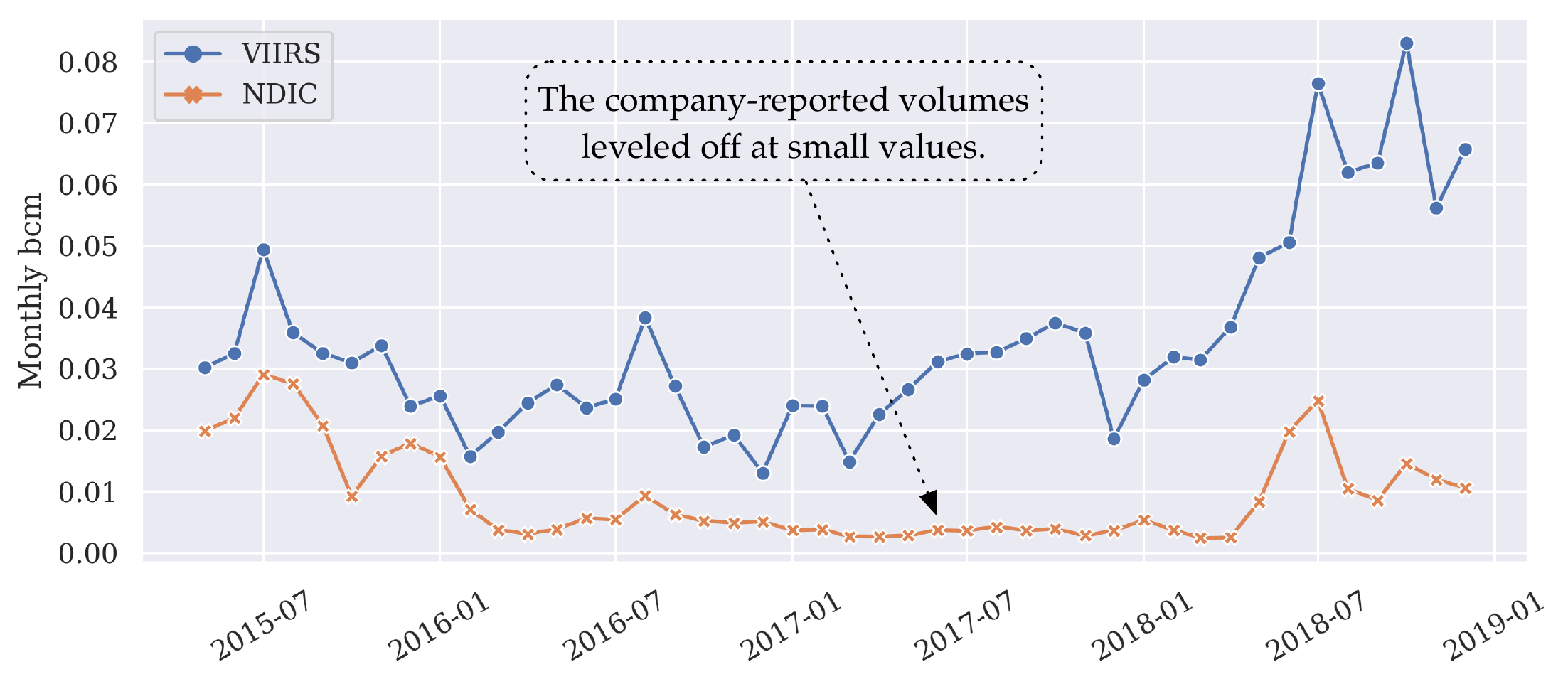}
		}\\
		\subfigure[Operator E]{
			\includegraphics[width=0.98\textwidth]{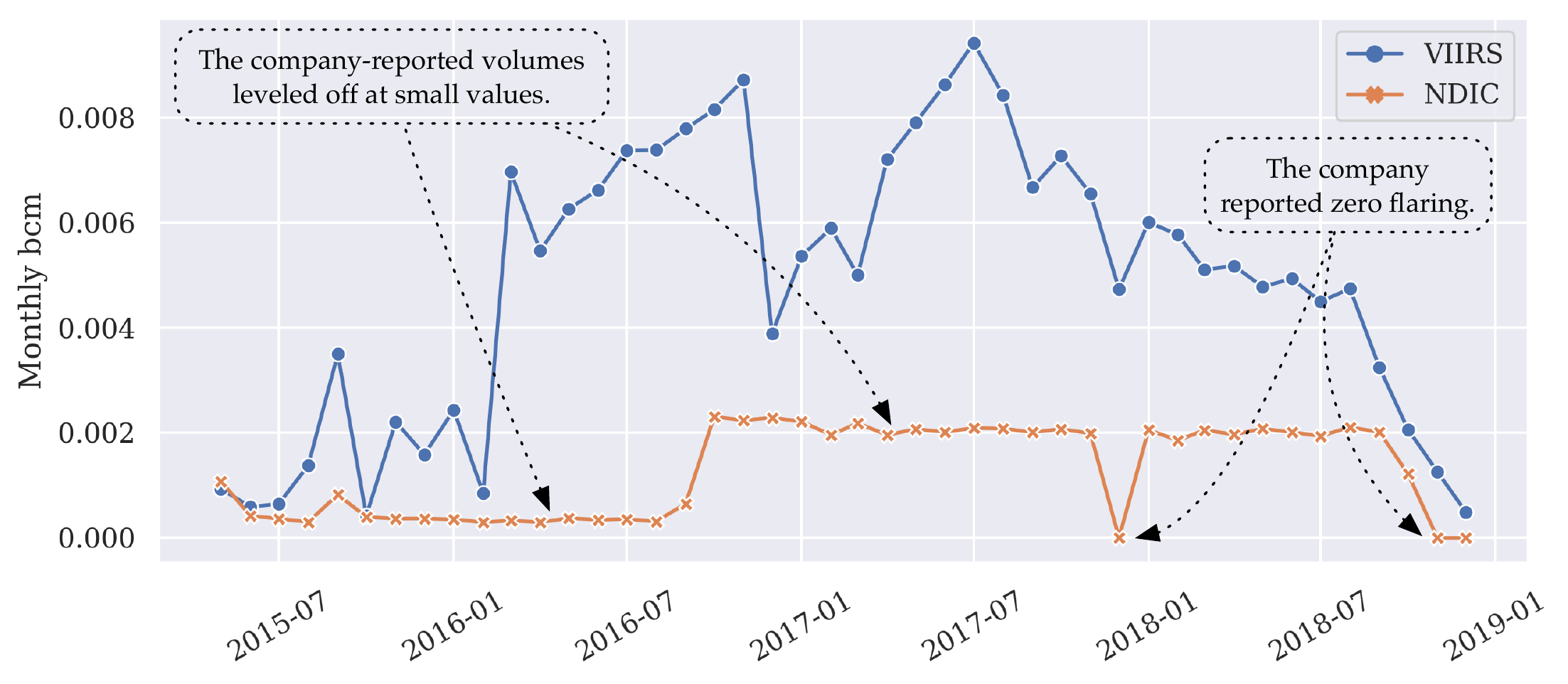}
		}
		\caption{\label{fig:bad_match_op_ts}Time series of the two example operators whose reporting did not quite align with the VIIRS detected trends/patterns. The points or periods in time for which the company-reported data were significantly different from the satellite detections are annotated.}
	\end{center}
\end{figure}

However, some operators (e.g., Operator D and Operator E) show discrepancies between their reportings and the satellite-detected flaring statistics, which are manifested through the poor fits (\ref{fig:bad_match_op}). Certainly, a poor fit with the linear model does not indicate much on its own. Nonetheless, there exists a pattern in both scatterplots that, some points seem to be ``pushed down'' towards the $x$-axis. If the time series of these two operators are drawn, it shows that this behavior is due to company-reported volumes leveling off for a certain period of time (\ref{fig:bad_match_op_ts}). The VIIRS curves in the time series imply that there were flaring intensity variations for those times. This workflow, driven by Algorithm~\ref{alg:nn_op}, is capable of raising a flag when it comes across datasets like these, and can serve as a powerful monitoring and analytics tool, however, strong cautions need to be applied.

The introduced approach, although it looks promising, is by no means a one-stop solution and has the potential for being misapplied. First, there is the possibility of misassigning the satellite-detected flares to the operators. Whenever the concern is raised, further investigations can be conducted by looking into the detection maps as well as the satellite imagery of the operators' production sites. In addition, this method is more effective for the relatively large producing/flaring operators, because when a company conducts very little flaring, the truncation effects discussed for the peak in \ref{fig:oilfield_SF_ppc} are magnified.

\subsection{Warnings Regarding Inconsistencies}
Given the resolution of the satellite imagery, assigning specific flaring volumes to a given operator is fraught with challenges. Although the VIIRS processing workflow is capable of picking up flares with areas around \SI{d0}{\square\metre} (\ref{fig:viirs-detect-lim}), the pixel footprint is much larger (\ref{tab:l8_viirs_resolution}). Since the latitude and longitude of the pixel center is stored for each individual VIIRS observation~\autocite{Elvidge2015MethodsData}, when multiple operators have sub-pixel combustion sources, it makes flare owner assignment extremely challenging. In such situations, conclusions reached by merely benchmarking company reporting against VIIRS reporting would likely be inaccurate. In fact, in the realm of NDIC reporting, warnings must be issued regarding any inconsistencies in those results, with considerations from three aspects. First, the report from the \textcite{U.S.DepartmentofEnergy2019NaturalImpacts} presents data supporting that North Dakota shows closer agreement between the NOAA estimations and state reportings (of flared gas volumes), when compared with Texas and New Mexico. Second, flaring is preferred over venting because methane (the main component of natural gas) is more potent than carbon dioxide which is the main product of flaring~\autocite{EIA2019Natural2018}. Since North Dakota bans venting, the massive flaring magnitude indicates that the direct release of gas into the atmosphere is minimized. Third, estimation of flaring volumes is inherently a difficult task. When it is not practicable to meter the flared gas, the \textcite{CanadianAssociationofPetroleumProducers2002EstimationFacilities} gives guidelines on available volume estimation methods. Every category of methods, no matter using rules of thumb, or experimentally determined correlations, or process simulators, has its own limitations and accuracy issues. Considering the fact that the VIIRS volumes used in this work were largely calibrated using the Cedigaz reported data (Section~\ref{sec:vnf_process_steps}), which has its own error bars~\autocite{Elvidge2015MethodsData}, the difference between company reporting and VIIRS reporting is inconclusive and unsurprising, especially when the standard error of the difference is larger than the difference itself.

By inspecting a more comprehensive profile of time series, both Operator D and Operator~E from the previous section are self-consistent in their reportings to the NDIC. Their time series are displayed in \ref{fig:burlington_ts_set} and \ref{fig:denbury_ts_set}, respectively. The variables and associated labels (shown in the legends) follow the same definitions from Section~\ref{sec:corr_analysis}. The units for all the variables are given in \ref{tab:op_ts_var_units}. Clearly, the reported flared volumes show good correspondence with the gas production and GOR profiles. Some rapid variations in their flared volumes match the fluctuations in the gas prices, i.e., when the gas price drops, the operators tend to flare more, whereas when the gas price reaches peak, there is little flaring. In summary, to nail down the decisions and conclusions with regard to operator reporting quality, better resolution satellite data and a more comprehensive review of the time series profiles are required.

\csmlongfigure[!htbp]{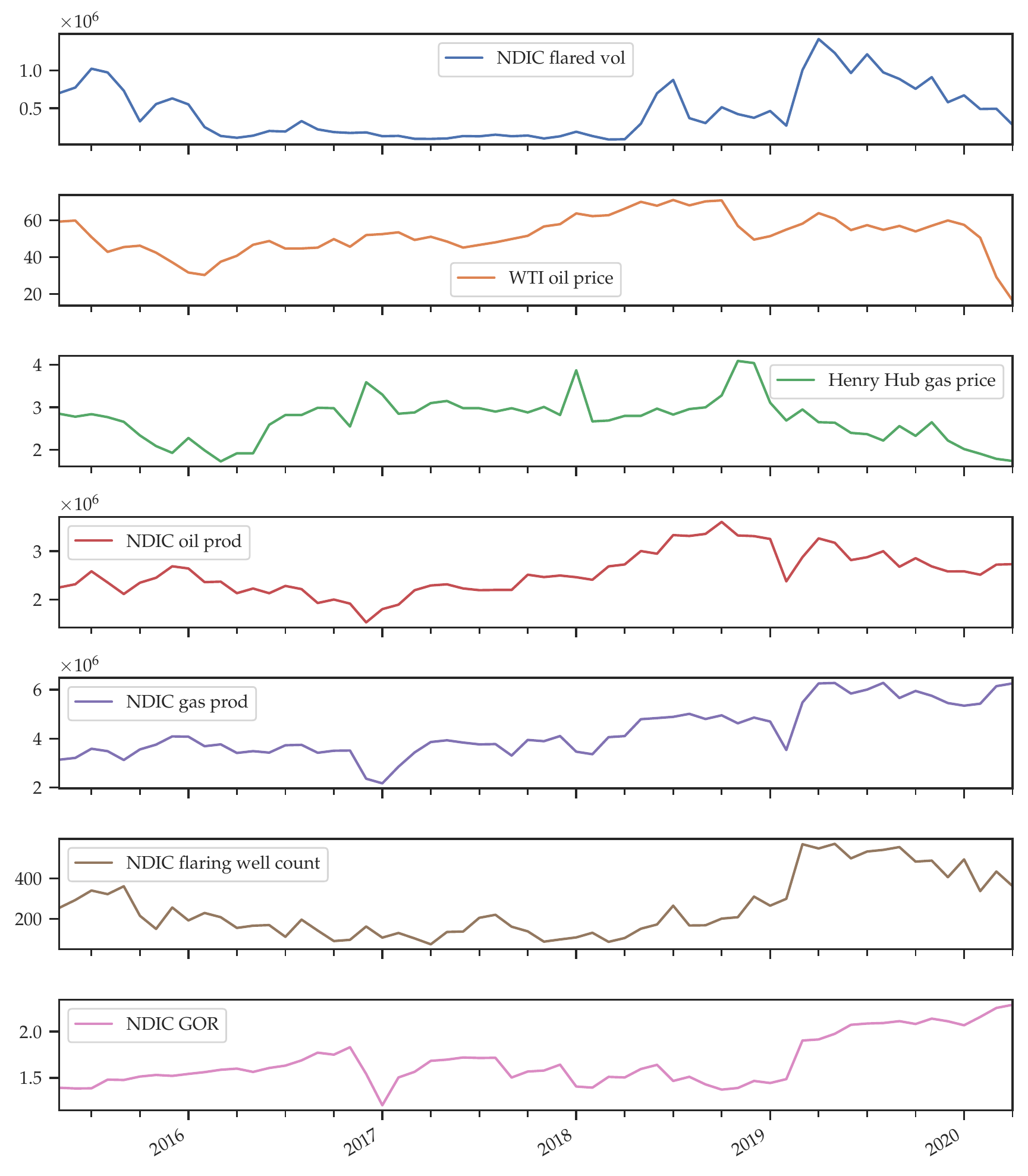}{figures/burlington_ts_set}{\textwidth}{A more comprehensive time series plot for Operator~D\@.}{ The increase in the reported flared volume in early 2019 corresponds to the gas price declining in the same period.}

\csmlongfigure[!htbp]{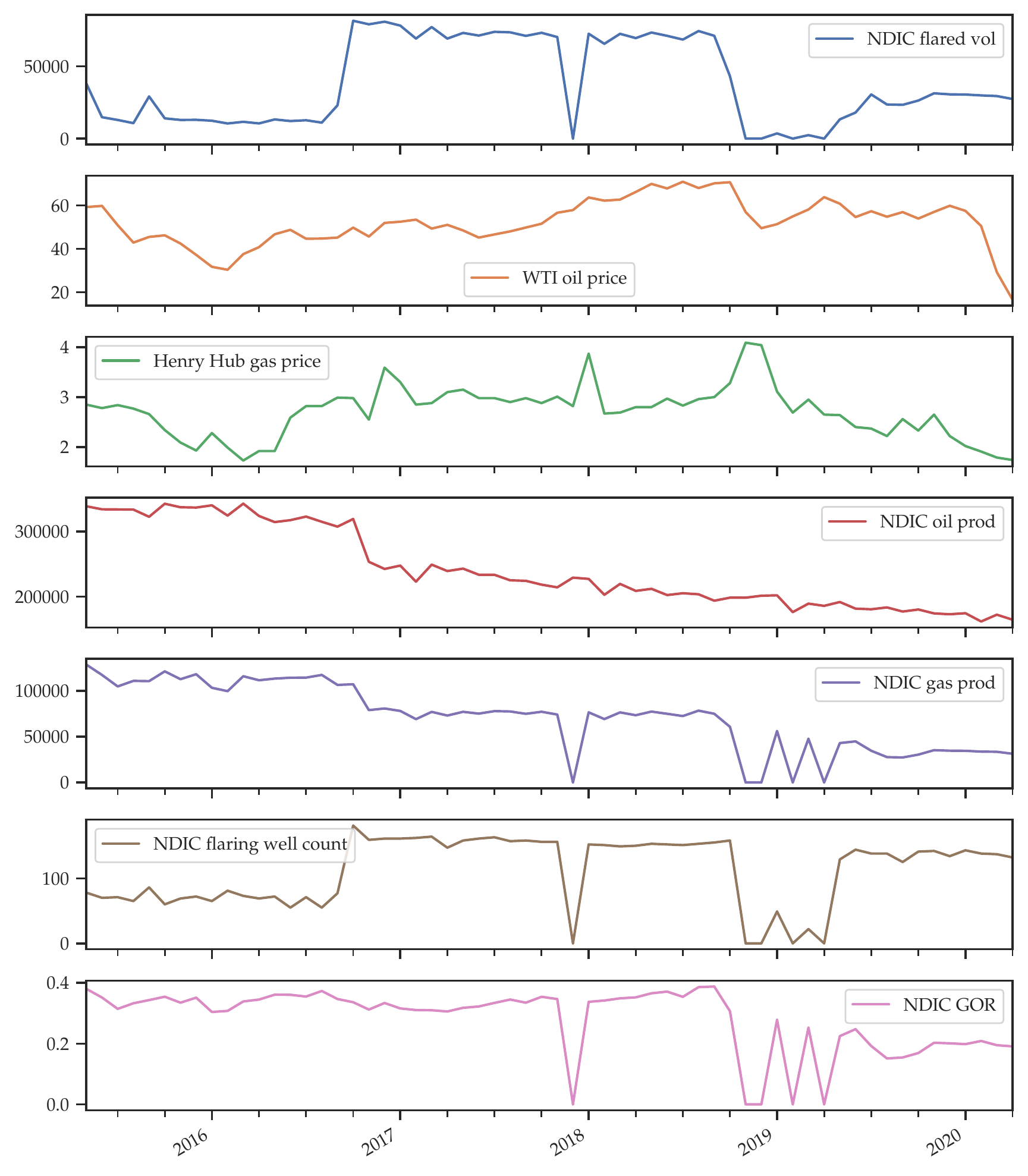}{figures/denbury_ts_set}{\textwidth}{A more comprehensive time series plot for Operator~E\@.}{ The sudden drop in the reported flared volume in late 2018 corresponds to the halted gas production.}

\begin{table}[!htbp]
\centering
\caption{Units for Operator Time Series in Figures 7.4 and 7.5}
\label{tab:op_ts_var_units}
\begin{tabular}{ll}
\toprule
Variable                & Unit                    \\ \midrule
NDIC flared vol         & \si{\mcf}                     \\
WTI oil price           & \si{\dpbbl}      \\
Henry Hub gas price     & \si{\dpmmbtu} \\
NDIC oil prod           & \si{\bbl}                     \\
NDIC gas prod           & \si{\mcf}                     \\
NDIC flaring well count & \SI{1}{}                       \\
NDIC GOR                & \si{\mcfpbbl}                 \\ \bottomrule
\end{tabular}
\end{table}

\FloatBarrier

\subsection{Caveats in Petroleum Data Analytics}
\label{sec:pe_ml_discuss}
As a petroleum engineer, the author is thrilled to witness the oil and gas industry and academia are embracing data-driven mindsets and solutions, while being part of it through writing this dissertation. However, there are certainly areas that could be continuously improved, and this section provides a discussion on one of those. That is, extending a cautious welcome to some black box models.

The pervasive influence of some black box models in the recent years can be seen by performing a rough search on OnePetro (\ref{tab:ml_trend}). One thing to note is that, from an algorithmic point of view, these methods are rather ``glass boxes'' as opposed to ``black boxes'', i.e., everything under the hood in terms of implementation is well understood. For example, backpropagation, which is the core of neural network training, is based on the chain rule. However, for a given task, the learned parameters inside the network provide little or no insights for the problem domain. Therefore, it is considered a black box.

\begin{table}[!htbp]
\centering
\caption{Publication Count Rise on OnePetro}
\label{tab:ml_trend}
\begin{threeparttable}
\begin{tabular}{l c S[table-format=3] S[table-format=4]}
\toprule
\multirow{2}{*}{\shortstack[c]{Exact Phrase \\Searched}} & \multirow{2}{*}{\shortstack[c]{Year Method \\Introduced}} & \multicolumn{2}{c}{Publication Count} \\ \cmidrule(lr){3-4}
                  &      & {2010--2014} & {2015--2019} \\ \midrule
neural network    & \cite*{Rosenblatt1958TheBrain.}\tnote{\dag} & 843         & 2044        \\
gradient boosting & \cite*{Friedman2001GreedyMachine}\tnote{\ddag} & 1         & 110        \\
random forest     & \cite*{Breiman2001RandomForests}\tnote{\S} & 9         & 245       \\ \bottomrule
\end{tabular}
\begin{tablenotes}
    \item[\dag] Based on~\autocite{Rosenblatt1958TheBrain.}
    \item[\ddag] Based on~\autocite{Friedman2001GreedyMachine}
    \item[\S] Based on~\autocite{Breiman2001RandomForests}
\end{tablenotes}
\end{threeparttable}
\end{table}

The wide adoption of such models is largely due to the availability of the open source libraries, for example in the Python ecosystem, construction and training of neural networks become much simpler thanks to TensorFlow and PyTorch, and gradient boosting models can be built within a few lines of code with the help of XGBoost, LightGBM, or CatBoost. In other words, with the mathematical details of those statistical routines abstracted away, for a practitioner, implementing those models is almost as easy as pushing a \textit{Learning} button on a GUI.

Unfortunately, easiness in the implementation does not imply appropriateness for the problem. In particular, those black box models face the challenges below:
\begin{enumerate}
    \item How to incorporate domain expertise.
    
    A lot of the black box models in the frequentist framework make the assumption that the observations are conditionally i.i.d. The hope is that by feeding a huge number of i.i.d.\ samples to a universal approximator, such as a neural network, some function for prediction can be optimized with a certain accuracy. For some applications, the domain expertise is often encoded in the feature selection process. For example, to train a model to predict oil production, the analyst might choose some completion parameters other than the API well number or well name, as input features.
    
    However, in the author's opinion, this way of incorporating domain expertise is still a shallow one, which is far from what the oil and gas industry have accumulated in many decades. For example, the phenomena of well interference through fracture hits leave the assumption of some neighboring wells being i.i.d.\ in an unfavorable position. Another example would be, when looking at a populations of wells from one basin that are completed by $N$ oilfield service companies, domain expertise might indicate that, each company deserves its own model while each company is not completely independent from others in terms of the completion technologies, etc. In this situation, the hierarchical model employed in Chapter~\ref{ch:hier} might be a better choice, in which case a lot of the prior knowledge about the different service companies can be incorporated into the population model.

    \item How to interpret the results.
    
    As discussed earlier, the black box models suffer from the interpretability issues. Using the shale gas wells example from Item 1 above, if a black box model is trained, it is impossible (at this point) to attribute the failure in capturing the well interference effects to a certain part of the neural network, or to a certain portion of the decision trees (in the case of gradient boosted trees or random forest). \textcite{Rudin2019StopInstead} asserted that people should ``stop explaining black box machine learning models'' and use interpretable models for high-stakes decisions. In the petroleum industry, there are a number of high-stakes decision scenarios, such as real-time well integrity anomaly detection and production forecasting in a high well cost context. Blindly applying black box models to those scenarios might involve serious losses. In terms of providing interpretability, the Bayesian approach employed throughout this dissertation is much more effective. Each and every assumption is expressed in the generative model through either the priors or the likelihood.

    \item How to quantify the uncertainties, especially in the context of risk management and decision making.
    
    Along the lines of Item 2 above, error bars are vital, especially in high-stakes prediction applications. In the case of predicting oil production using a trained data-driven model, point prediction results such as \SI{1000}{\bpd} are not really insightful. In fact, if the \SI{95}{\percent} prediction interval (PI) is \SI{1000 \pm 50}{\bpd}, that point prediction becomes more informative. However, if the \SI{95}{\percent} PI is \SI{1000 \pm 1500}{\bpd}, that same point prediction is unhelpful or misleading. What shall be reported instead is either the considered model yields much uncertainty in this given task, or there is possibility that the entity will not produce anything at all. 
    
    It should be noted that, the `\num{95}' in the CI/PI is not a ``magic number''. A state government or an oil company might want to make decisions based on \SI{73}{\percent} or \SI{99.6}{\percent} confidence, or any other arbitrary choices. What really matters is the necessity of a principled way to quantify the uncertainties in machine learning-based estimations/predictions, such that any intervals can be computed. As presented throughout this dissertation, the Bayesian approach provides full capacity and flexibility is this regard. In fact, for parameter estimates, the author chooses to give \SI{90}{\percent} CI instead of the ``conventional'' \SI{95}{\percent}, to emphasize that this should be a domain's consideration rather than a statistical one.
    
    A lot of the black box models in the frequentist framework, however, fall short of this requirement. Maximum likelihood estimation (MLE), which is fundamentally relied upon by some frequentist learning methods, enjoys really nice properties and is capable of quantifying uncertainties, but only when a massive amount of data is at hand such that the asymptotic properties could take effect. Unfortunately, that is not the case in many scenarios for the petroleum engineering domain, which is discussed next.
    
    \item How to mitigate overfitting when the data is not ``big''.
    
    Two aspects are worth discussing here. For one thing, the big data is not everywhere. Indeed, the author believes that the claim of \textcite{Gelman2015NLarge} that, ``sample sizes are never large'', applies to a lot of problems in the petroleum industry. The reason is that, if the data were large, the analyst would already be on to the next problem for which more data is needed. For example, a sample of \num{500} producing wells in the Bakken Formation could make some general study possible. When the analyst has access to a dataset of more than \num{15000} wells, some granular insights are desirable. Especially, if partial pooling is needed among the different service companies/operators, different members of the formation, or different completion technologies, data for some units of the population could be very small (which happens for the analysis in Chapter~\ref{ch:hier}).
    
    On the other hand, the sample size should be inspected in the light of model complexity. The number of parameters provides one measure of such. For example, consider a hypothetical classification problem, whose goal is to determine if a given well will deliver good or average or poor production performance. Ten completion parameters (features) are available to train the multilayer perceptron illustrated in \ref{fig:nn}.
    \csmlongfigure[!htbp]{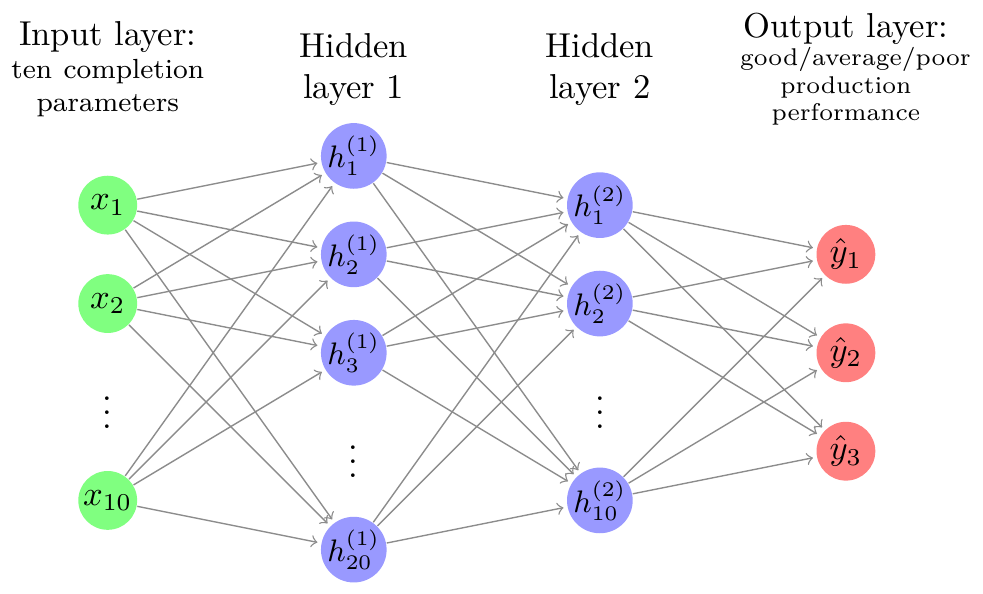}{figures/nn}{0.7\textwidth}{A neural network designed for the hypothetical well performance classification problem.}{ The input layer has \num{10} neurons for the completion parameters. The first and second hidden layer has \num{20} and \num{10} neurons, respectively. The output layer has three neurons for multiclass classification.}
    
    In this (small) neural network, the number of parameters $n_p$ is given by:
    \begin{equation}
        n_p = 11\times20 + 21\times10 + 11\times3 = 463,
    \end{equation}
    when considering a single bias node for every layer except the last one. To train this model, a dataset of \num{500} wells would definitely be a small sample. There is still possibility to train such a model with a small sample, however, great efforts in regularization have to be made, in the hope that the neural network will learn something that can be generalized, instead of merely memorizing the observed samples (i.e., overfitting).
    
    By utilizing the regularizing priors, the Bayesian approach's built-in Occam's razor greatly mitigate the risk of overfitting. In particular, Bayesian nonparametric models, such as the Gaussian processes employed in Chapter~\ref{ch:gp}, are very attractive in a sense that the sizes of models are allowed to grow with the size of data~\autocite{Orbanz2010}. This makes the developed model flexible while being robust to overfitting.
\end{enumerate}

Although the Bayesian learning models (such as the ones developed in this work) have outstanding merits and deserve wider utilization in petroleum data analytics, they are not cure-alls. Recently researchers have started to stress the necessity of \textit{bespoke} statistical models~\autocite{Andorra2020ABetancourt,mcelreath2020rethink}. The argument is that, off-the-shelf models, no matter neural networks or generalized linear models, interrupt the incorporation of domain expertise. This is especially relevant in the field of petroleum engineering. For instance, when conducting data-driven analysis for hydraulic fracturing performance, it makes sense to bring in the fracture propagation models to the machine learning workflow. That way, statistical models are motivated by the physically informed models. The Bayesian framework, as employed throughout this dissertation, readily embraces this strategy, in that the domain knowledge, which is represented by differential equations for example, can be inserted into the generative model. One advantage is that a lot of the parameters will have direct scientific meanings, and more informative priors can be placed based on scientific constraints, field experience, etc. The final outcome should be better inferences and predictions.

\chapter{Conclusions and Recommendations}\label{ch:conclusions}
In this dissertation, the effectiveness of a full Bayesian approach has been observed in learning models from natural gas flaring data. The author hopes this work contributes to the understanding of the options and considerations when applying data-driven approaches to gas flaring. In closing, this chapter presents the major conclusions and recommendations for future work.
\subsection{Conclusions}
The major conclusions are:
\begin{enumerate}
    \item Bayesian learning implemented using Hamiltonian Monte Carlo can be effectively applied to real problems in gas flaring analytics, in both supervised and unsupervised settings. The advantages of the Bayesian approach indicate it deserves wider usage in the petroleum engineering domain in general; these advantages are listed below:
        \begin{enumerate}
            \item Petrotechnical domain expertise can be incorporated in a principled way.
            \item Model interpretability is drastically improved, facilitating communications with petroleum engineers. 
            \item Quantification of uncertainty leads to more robust decision making, which is important for oil exploration and production companies.
            \item The built-in Occam's razor makes the model less prone to overfitting, in the context of noisy field measurements.
        \end{enumerate}
    \item The development of a suite of models (\ref{tab:list_models}), with both parametric and nonparametric techniques, provides guidance on how insights can be extracted from various angles. The presented models are designed and tested to be able to generalize to different entities at various levels.
    \item To investigate the heterogeneity among the different entities (such as counties or oilfields), partial pooling is recommended, because some entities have very little data.
    \item Gaussian processes demonstrate very attractive traits in revealing the patterns and trends from flaring time series. A set of priors with the Mat\'ern $5/2$ kernel works very well across different modeling goals, observation models, and data sources.
    \item From a distributional point of view, the negative binomial and Gaussian mixture models are good representations of the oilfield flare counts and flared volumes, respectively. The learned parameters and structures are very interpretable. Hidden clusters are found by fitting Gaussian mixture models.
    \item A nearest-neighbor-based approach for operator level monitoring and analytics is introduced. Its performance is tested on real data and defendable results are obtained. However, better resolution satellite data is needed for the scenario of multiple operators' wells being very close to each other.
    \item All the dissertation objectives (Section~\ref{sec:thesis_obj}) have been achieved. In particular, the flared volumes missed from VIIRS for the state and each county are estimated via fitting the intercept parameter and reported in \ref{tab:eda_st_param} and \ref{tab:county_param}. The nighttime combustion source detection limits of Landsat 8, without being corrected for artifacts due to glow, are determined and reported in \ref{fig:L8-detect-lim}. Correlations between financial factors, production performance, and flared volumes at a state level are computed using Spearman's $\rho$ and reported in \ref{fig:oil_pr_corr} and \ref{fig:oil_pr_corr_lag1} for the original data and lag-$1$ differences, respectively. Most pairs of the variables do not show strong correlations on the lag-$1$ differences. Robust Gaussian process modeling serves as a generic framework for addressing the rest of the objectives, including demonstrating operator approaches, evaluating if the goals of the North Dakota regulatory policy (Order 24665) have been achieved, and predicting NDIC flared volumes.
\end{enumerate}
\begin{table}[!htbp]
\centering
\caption{Models Developed in this Dissertation}
\label{tab:list_models}
\resizebox{\textwidth}{!}{%
\begin{tabular}{l l S[table-format=2]}
\toprule
Numbering                             & Target of Modeling                                                         & {Page}                                \\ \midrule
Model~\ref{mod:state_lin_mod}         & Associations between VIIRS and NDIC at a state level                       & \pageref{mod:state_lin_mod}         \\
Model~\ref{mod:county_hier_cp}        & Associations between VIIRS and NDIC at a county level (centered)           & \pageref{mod:county_hier_cp}        \\
Model~\ref{mod:county_hier_ncp}       & Associations between VIIRS and NDIC at a county level (noncentered)        & \pageref{mod:county_hier_ncp}       \\
Model~\ref{eq:gp_gas_p}               & Proportion of gas production being flared as time series                   & \pageref{eq:gp_gas_p}               \\
Model~\ref{mod:well_proportion_flare} & Proportion of wells that conduct flaring as time series                    & \pageref{mod:well_proportion_flare} \\
Model~\ref{mod:cox_flare_ct}          & VIIRS detection count as time series                                       & \pageref{mod:cox_flare_ct}          \\
Model~\ref{mod:oil_proportion_flare}  & Proportion of oil being flared as time series                              & \pageref{mod:oil_proportion_flare}  \\
Model~\ref{mod:sf_viirs_ndic}         & Scale factor between VIIRS and NDIC as time series                         & \pageref{mod:sf_viirs_ndic}         \\
Model~\ref{mod:of_ct_negbin}          & VIIRS detection count distribution for oilfields                           & \pageref{mod:of_ct_negbin}          \\
Model~\ref{mod:gmm_lat_var}           & VIIRS volume distribution for oilfields (latent discrete parameterization) & \pageref{mod:gmm_lat_var}           \\
Model~\ref{mod:gmm_marg}              & VIIRS volume distribution for oilfields (marginalized)                     & \pageref{mod:gmm_marg}              \\ \bottomrule
\end{tabular}%
}
\end{table}

\subsection{Future Work}
A number of areas for future research include:
\begin{enumerate}
    \item L8 processing workflow.
    
    The studies of Section~\ref{sec:l8_glow} indicate that the inclusion of L8 information (using the existing VIIRS workflow) faces the challenges of the processing artifacts due to glow. It would be interesting to tailor the processing algorithm for L8, which opens the door for data fusion of VIIRS and L8, providing much better resolution interpretations.

    \item Fast detection of flares on a monthly basis.
    
    The development of a rapid flare detection and volume estimation method (based on satellite imagery) will lead to continuous monthly data streams. Since NDIC needs about two months' turnaround time to compile and digitize the company reports, many of the machine learning workflows proposed in this dissertation will be able to provide predictive insights with rapid detection data.

    \item Hierarchical Gaussian processes.
    
    The models in Chapter~\ref{ch:gp} are learned from each entity's own data. It would be interesting to see how far the scheme of partial pooling (Chapter~\ref{ch:hier}) can be taken. Can pooling across different entities via hierarchical Gaussian processes improve the inferences?

    \item Spatial-temporal analysis.
    
    One step further from Item 3 above, the efficacy of spatial-temporal models (which allow for pooling information across time and space) are worth investigating. Are neighboring entities exhibiting close resemblance in flaring behaviors?

    \item Unify everything under Bayesian nonparametrics.
    
    The model comparison for GMMs in Chapter~\ref{ch:gmm} depends on specifying the potential numbers of clusters a priori. In fact, Dirichlet process, as an infinite-dimensional generalization of the Dirichlet distribution, is nonparametric and allows for automatically choosing the number of necessary clusters. Considering the effectiveness of GP (Chapter~\ref{ch:gp}), it would be interesting to see how far the nonparametric models can be taken in flaring data analytics. Can all of the gas flaring analytics problems be addressed in an unified framework of Bayesian nonparametrics?
    
\end{enumerate}

\backmatter


\setcounter{biburllcpenalty}{7000} 
\setcounter{biburlucpenalty}{8000} 
\setlength\bibitemsep{1.5\itemsep} 
\begingroup 
    \setstretch{1.3}
    
    \xpatchbibmacro{date+extradate}{%
      \printtext[parens]%
    }{%
      \setunit{\addperiod\space}%
      \printtext%
    }{}{}
    \printbibliography
\endgroup





\end{document}